\documentclass{article}

\usepackage{microtype}
\usepackage{graphicx}
\usepackage{booktabs} 
\usepackage{amsthm}
\usepackage{amsmath}
\usepackage{amssymb}
\usepackage{subfig}
\usepackage{multirow}
\usepackage{wrapfig}
\usepackage{enumerate}
\usepackage{mathtools}
\usepackage[makeroom]{cancel}
\usepackage[title]{appendix}
\usepackage{moresize}
\usepackage{todonotes}

\usepackage[utf8]{inputenc}
\usepackage{pgfplots}
\pgfplotsset{compat=newest}
\usepgfplotslibrary{groupplots}
\usepgfplotslibrary{dateplot}

\graphicspath{{figs/}}

\usepackage{tikz}
\usetikzlibrary{external}
\usepackage{hyperref}
\usepackage{cleveref}

\usepackage[accepted]{icml2020}

\usepackage{algpseudocode}
\renewcommand{\algorithmicrequire}{\textbf{Input:}}
\renewcommand{\algorithmicensure}{\textbf{Output:}}

\usepackage{amsmath,amsfonts,bm}

\def\1{\bm{1}}

\def\vmu{{\boldsymbol{\mu}}}
\def\vtheta{{\boldsymbol{\theta}}}

\def\vb{{\mathbf{b}}}
\def\vc{{\mathbf{c}}}
\def\vd{{\mathbf{d}}}
\def\ve{{\mathbf{e}}}

\def\vu{{\mathbf{u}}}

\def\vw{{\mathbf{w}}}
\def\vx{{\mathbf{x}}}
\def\vy{{\mathbf{y}}}
\def\vz{{\mathbf{z}}}

\def\mA{{\mathbf{A}}}
\def\mB{{\mathbf{B}}}

\def\mE{{\mathbf{E}}}
\def\mF{{\mathbf{F}}}

\def\mH{{\mathbf{H}}}
\def\mI{{\mathbf{I}}}
\def\mJ{{\mathbf{J}}}

\def\mS{{\mathbf{S}}}
\def\mT{{\mathbf{T}}}
\def\mU{{\mathbf{U}}}
\def\mV{{\mathbf{V}}}
\def\mW{{\mathbf{W}}}

\def\mSigma{{\boldsymbol{\Sigma}}}

\DeclareMathAlphabet{\mathsfit}{\encodingdefault}{\sfdefault}{m}{sl}
\SetMathAlphabet{\mathsfit}{bold}{\encodingdefault}{\sfdefault}{bx}{n}

\newcommand{\E}{\mathbb{E}}

\newcommand{\R}{\mathbb{R}}

\newcommand{\softmax}{\mathrm{softmax}}

\DeclareMathOperator*{\argmin}{arg\,min}

\newcommand{\vphi}{\boldsymbol{\phi}}

\renewcommand{\vx}{\mathbf{x}}
\renewcommand{\vy}{\mathbf{y}}
\renewcommand{\vz}{\mathbf{z}}

\renewcommand{\vw}{\mathbf{w}}
\renewcommand{\vb}{\mathbf{b}}

\renewcommand{\b}{\mathbf}
\renewcommand{\vec}{\mathrm{vec}}

\newcommand{\MN}{\mathcal{MN}}
\newcommand{\N}{\mathcal{N}}
\newcommand{\diag}[1]{\mathrm{diag}(#1)}

\renewcommand{\R}{\mathbb{R}}
\renewcommand{\E}{\mathbb{E}}
\newcommand{\D}{\mathcal{D}}
\newcommand{\norm}[1]{\Vert #1 \Vert}
\newcommand{\abs}[1]{\vert #1 \vert}
\newcommand{\inv}{{-1}}
\newcommand{\lambdamin}{\lambda_\text{min}}
\newcommand{\lambdamax}{\lambda_\text{max}}

\usepackage{thmtools}
\usepackage{thm-restate}

\newtheorem{proposition}{Proposition}[section]
\newtheorem{lemma}[proposition]{Lemma}
\newtheorem{theorem}[proposition]{Theorem}

\theoremstyle{definition}

\Crefname{appsec}{Appendix}{Appendices}

\icmltitlerunning{Being Bayesian, Even Just a Bit, Fixes Overconfidence in ReLU Networks}

\begin{document}

\twocolumn[
    \icmltitle{Being Bayesian, Even Just a Bit,\\ Fixes Overconfidence in ReLU Networks}



\icmlsetsymbol{equal}{*}

\begin{icmlauthorlist}
\icmlauthor{Agustinus Kristiadi}{tue}
\icmlauthor{Matthias Hein}{tue}
\icmlauthor{Philipp Hennig}{tue,mpi}
\end{icmlauthorlist}

\icmlaffiliation{tue}{University of T\"{u}bingen}
\icmlaffiliation{mpi}{MPI for Intelligent Systems, T\"{u}bingen}

\icmlcorrespondingauthor{Agustinus Kristiadi}{agustinus.kristiadi@uni-tuebingen.de}

\icmlkeywords{Bayesian deep learning, uncertainty quantification, robustness}

\vskip 0.3in
]



\printAffiliationsAndNotice{}  

\begin{abstract}
The point estimates of ReLU classification networks---arguably the most widely used neural network architecture---have been shown to yield arbitrarily high confidence far away from the training data. This architecture, in conjunction with a maximum a posteriori estimation scheme, is thus not calibrated nor robust. Approximate Bayesian inference has been empirically demonstrated to improve predictive uncertainty in neural networks, although the theoretical analysis of such Bayesian approximations is limited. We theoretically analyze approximate Gaussian distributions on the weights of ReLU networks and show that they fix the overconfidence problem. Furthermore, we show that even a simplistic, thus cheap, Bayesian approximation, also fixes these issues. This indicates that a sufficient condition for a calibrated uncertainty on a ReLU network is ``to be a bit Bayesian''. These theoretical results validate the usage of last-layer Bayesian approximation and motivate a range of a fidelity-cost trade-off. We further validate these findings empirically via various standard experiments using common deep ReLU networks and Laplace approximations.
\end{abstract}

\section{Introduction}
\label{sec:intro}

As neural networks have been successfully applied in ever more domains, including safety-critical ones, the robustness and uncertainty quantification of their predictions have moved into focus, subsumed under the notion of AI safety \citep{amodei2016concrete}. A principal goal of uncertainty quantification is that learning machines and neural networks in particular, should assign low confidence to test cases not explained well by the training data or prior information \citep{gal2016uncertainty}. The most obvious cases are test points that lie ``far away'' from the training data. Many methods to achieve this goal have been proposed, both Bayesian \citep[e.g.][]{blundell_weight_2015, louizos2017multiplicative, DBLP:journals/corr/abs-1712-02390} and non-Bayesian \citep[e.g.][]{lakshminarayanan2017simple,liang2018enhancing,hein2019relu}.

ReLU networks are currently among the most widely used neural architectures. This class comprises any network that can be written as a composition of linear layers (including fully-connected, convolutional, and residual layers) and a ReLU activation function. But, while ReLU networks often achieve high accuracy, the \emph{uncertainty} of their predictions has been shown to be miscalibrated \citep{guo17calibration}. Indeed, \citet{hein2019relu} demonstrated that ReLU networks are \emph{always} overconfident ``far away from the data'': scaling a training point $\vx$ (a vector in a Euclidean input space) with a scalar $\delta$ yields predictions of arbitrarily high confidence in the limit $\delta\to\infty$. This means ReLU networks are susceptible to out-of-distribution (OOD) examples. Meanwhile, probabilistic methods (in particular Bayesian methods) have long been known empirically to improve predictive uncertainty estimates. \citet{mackay1992evidence} demonstrated experimentally that the predictive uncertainty of Bayesian neural networks will naturally be high in regions not covered by training data. Although the theoretical analysis is still lacking, results like this raise the hope that the overconfidence problem of ReLU networks, too, might be mitigated by the use of probabilistic and Bayesian methods.

\begin{figure*}[t]
  \centering

  \subfloat{\includegraphics[width=0.22\textwidth, height=0.15\textwidth]{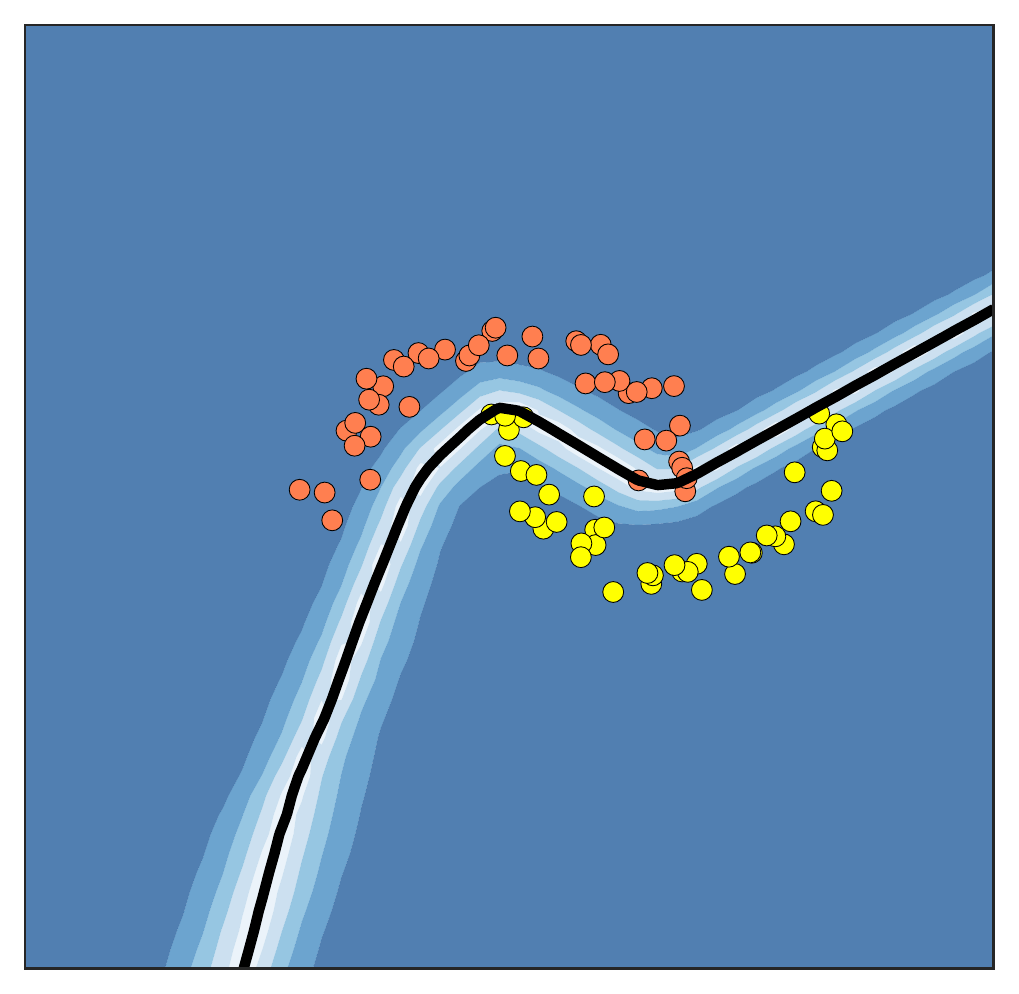}}
  \quad
  \subfloat{\includegraphics[width=0.22\textwidth, height=0.15\textwidth]{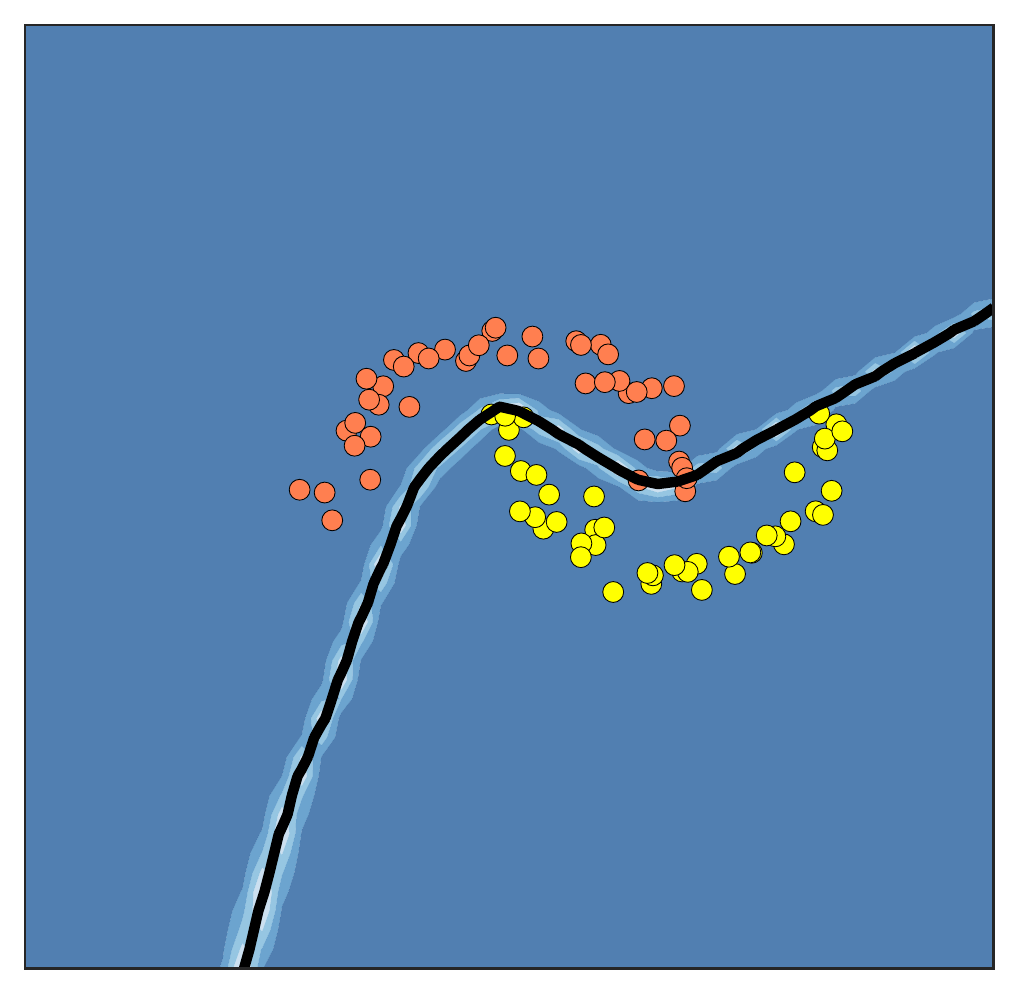}}
  \quad
  \subfloat{\includegraphics[width=0.22\textwidth, height=0.15\textwidth]{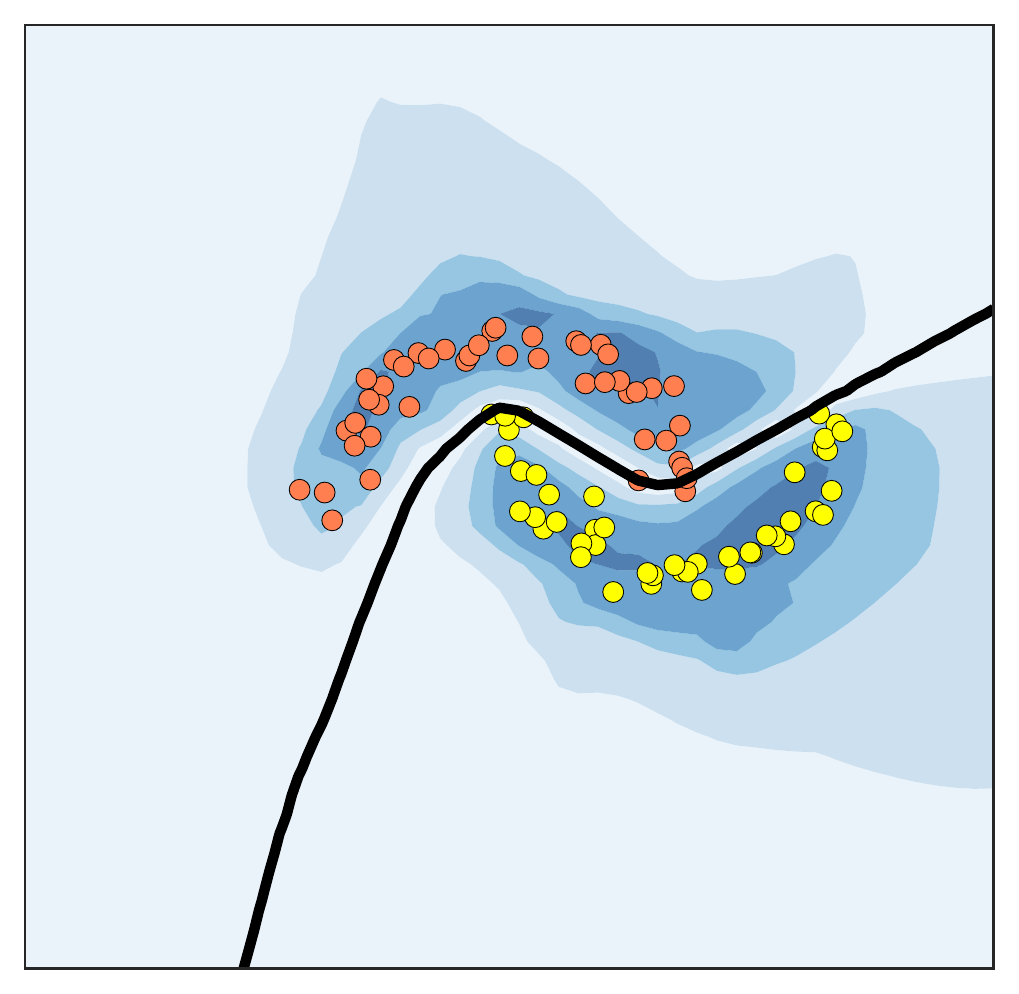}}
  \quad
  \subfloat{\includegraphics[width=0.22\textwidth, height=0.15\textwidth]{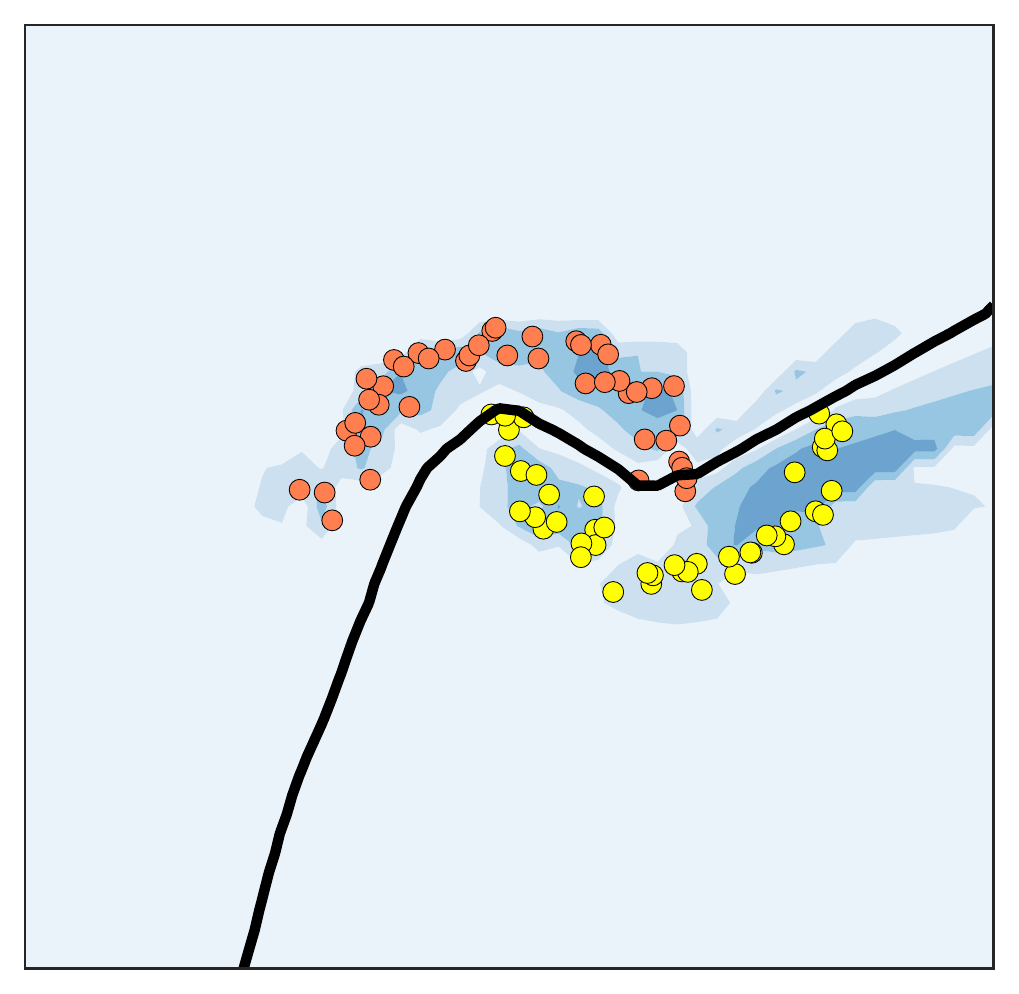}}
  \quad
  \subfloat{\includegraphics[width=0.03\textwidth, height=0.15\textwidth]{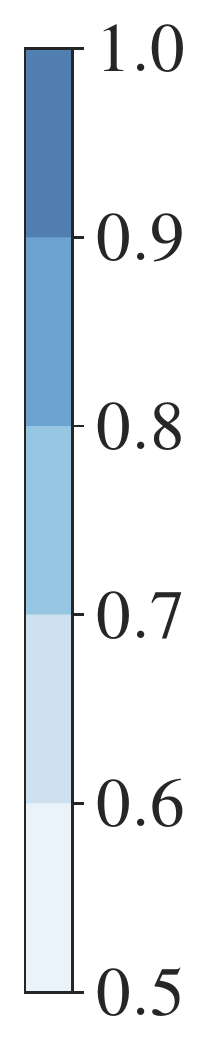}}

  \setcounter{subfigure}{0}

  \subfloat[MAP]{\includegraphics[width=0.22\textwidth, height=0.15\textwidth]{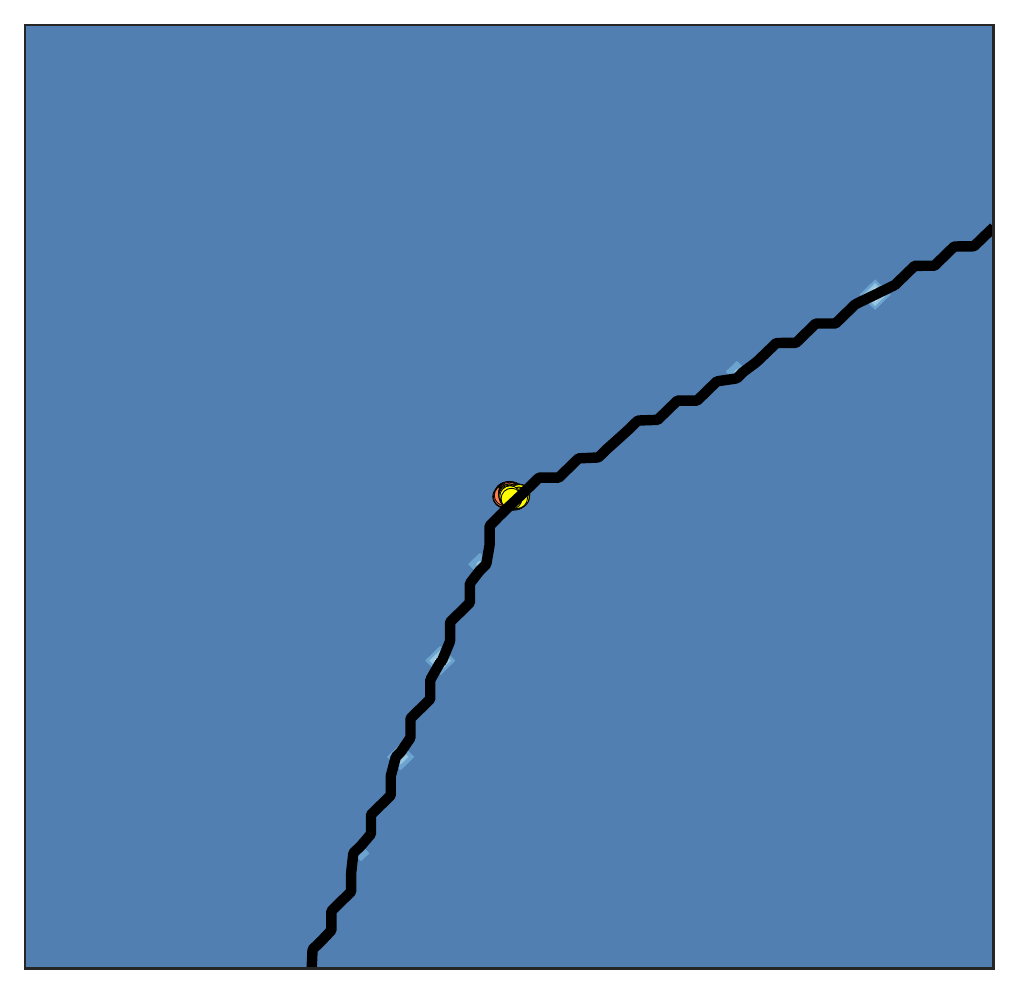}}
  \quad
  \subfloat[Temp. scaling]{\includegraphics[width=0.22\textwidth, height=0.15\textwidth]{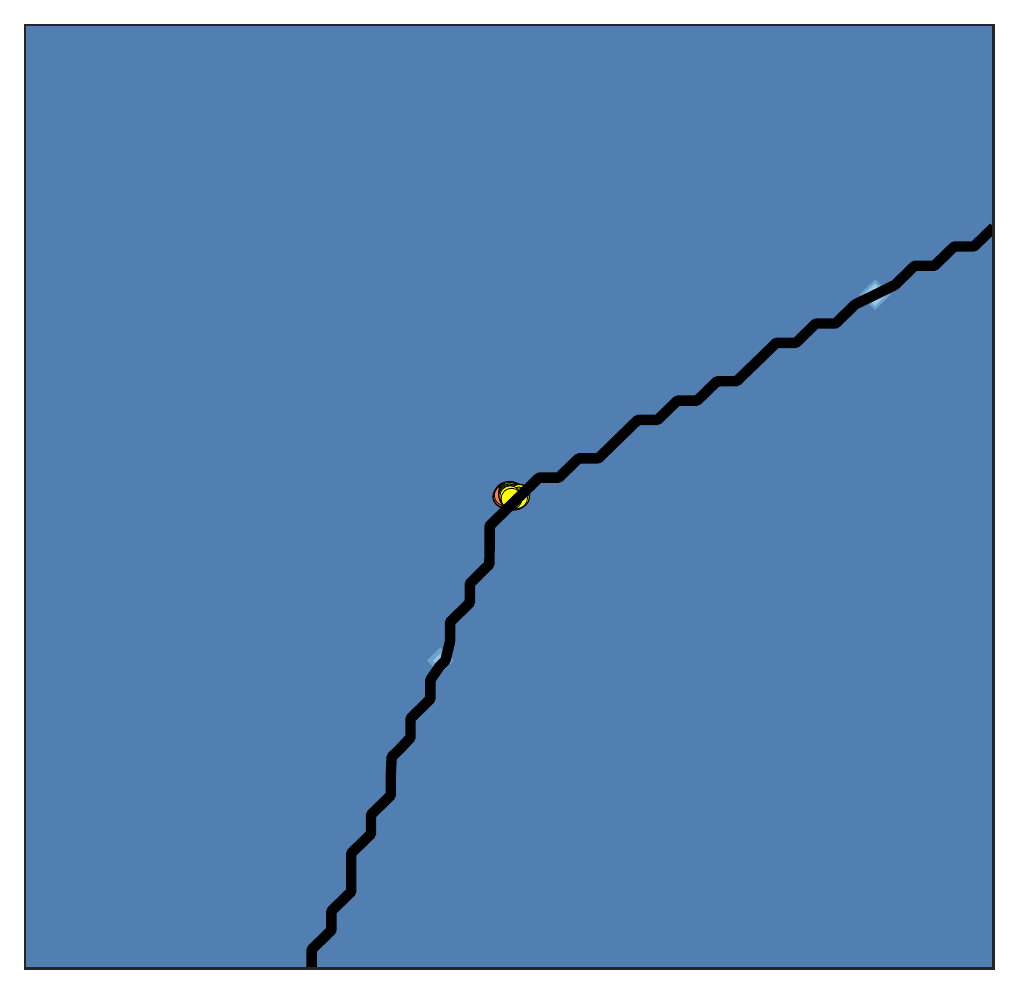}}
  \quad
  \subfloat[Bayesian (last-layer)]{\includegraphics[width=0.22\textwidth, height=0.15\textwidth]{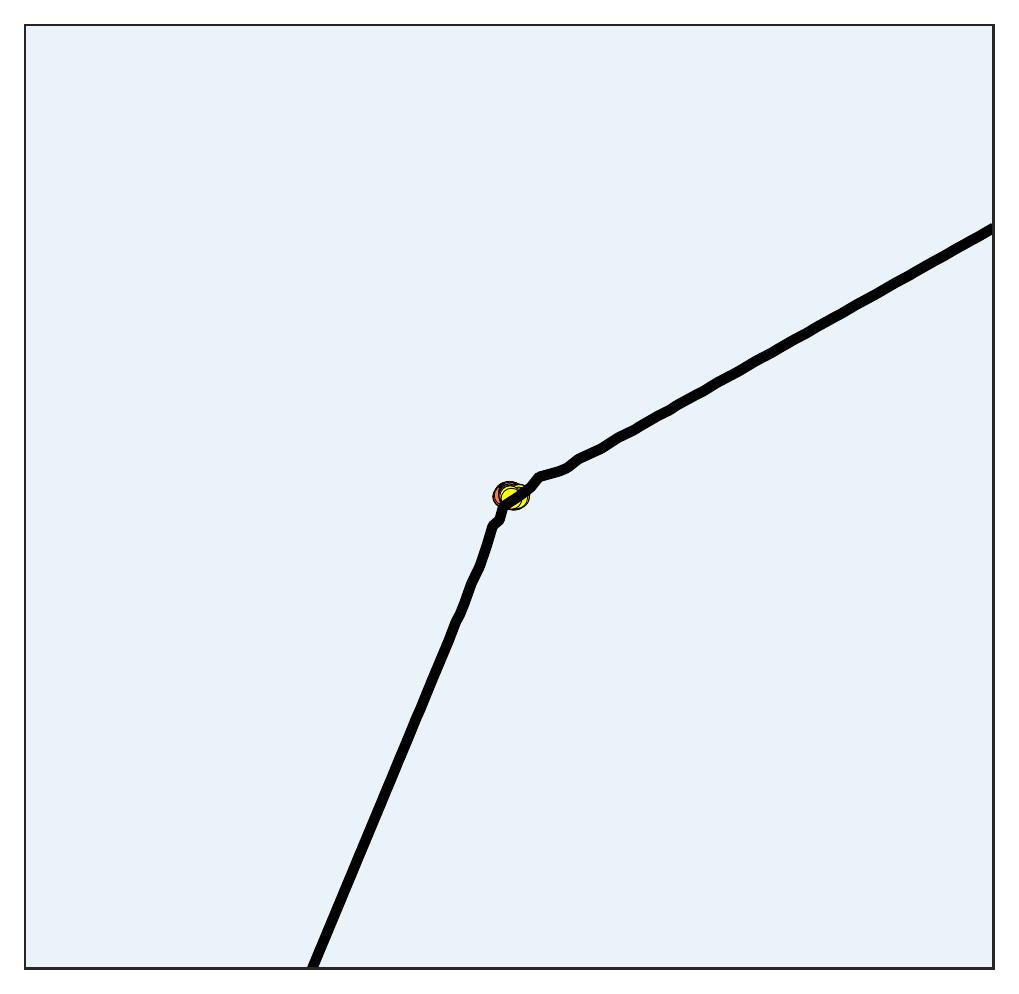}}
  \quad
  \subfloat[Bayesian (all-layer)]{\includegraphics[width=0.22\textwidth, height=0.15\textwidth]{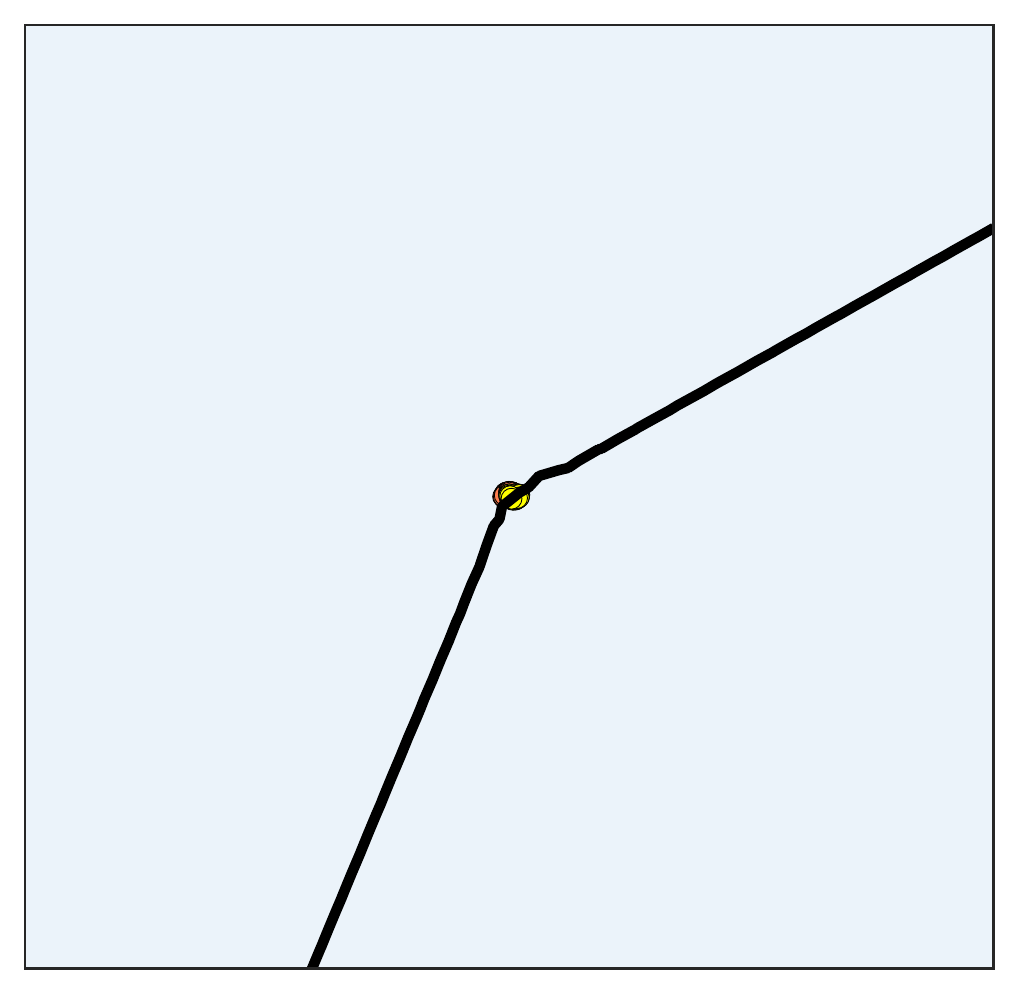}}
  \quad
  \subfloat{\includegraphics[width=0.03\textwidth, height=0.15\textwidth]{toy_2d_bnn_colorbar}}

  \caption{Binary classification on a toy dataset using a MAP estimate, temperature scaling, and both last-layer and all-layer Gaussian approximations over the weights which are obtained via Laplace approximations. Background color and black line represent confidence and decision boundary, respectively. Bottom row shows a zoomed-out view of the top row. The Bayesian approximations---even in the last-layer case---give desirable uncertainty estimates: confident close to the training data and uncertain otherwise. MAP and temperature scaling yield overconfident predictions. The optimal temperature is picked as in \citet{guo17calibration}.}

  \label{fig:laplaces}
\end{figure*}

This paper offers a theoretical analysis of the binary classification case of ReLU networks with a logistic output layer. We show that equipping such networks with a Gaussian approximate distribution over the weights mitigates the aforementioned theoretical problem, in the sense that the predictive confidence far away from the training data approaches a known limit, bounded away from one, whose value is controlled by the covariance. In the case of Laplace approximations \citep{mackay1992practical,ritter_scalable_2018}, this treatment in conjunction with the probit approximation \citep{spiegelhalter1990sequential,mackay1992evidence} does not change the decision boundary of the trained network, so it has no negative effect on the predictive performance (cf. \Cref{fig:laplaces}). Furthermore, we show that a sufficient condition for this desirable property to hold is to apply a Gaussian approximation \emph{only} to the last layer of a ReLU network. This motivates the commonly used approximation scheme where an $L$-layer network is decomposed into a fixed feature map composed by the first $L-1$ layers and a Bayesian linear classifier \citep[etc.]{gelman2008weakly,wilson2016deep,riquelme2018deep,ober2019benchmarking,brosse2020last}. This particular result implies that just being ``a bit'' Bayesian---at low cost overhead---already gives desirable benefits.

We empirically validate our results through various Laplace approximations on common deep ReLU networks. Furthermore, while our theoretical analysis is focused on the binary classification case, we also experimentally show that these Bayesian approaches yield good performance in the multi-class classification setting, suggesting that our analysis may carry over to this case.

To summarize, our contributions are three-fold:

\begin{enumerate}[(i)]
    \item we provide theoretical analysis on why ReLU networks equipped with Gaussian distributions over the weights mitigate the overconfidence problem in the binary classification setting,

    \item we show that a sufficient condition for having this property is to be ``a bit'' Bayesian: employing a last-layer Gaussian approximation---in particular a Bayesian one, and

    \item we validate our theoretical findings via a series of comprehensive experiments involving commonly-used deep ReLU networks and Laplace approximations in both binary and multi-class cases.
\end{enumerate}

\Cref{sec:theory} begins with definitions, assumptions, and the problem statement, then develops the main theoretical results. Proofs are available in \Cref{appendix:proofs}. We discuss related work in \Cref{sec:related}, while empirical results are shown in \Cref{sec:experiments}.

\section{Analysis}
\label{sec:theory}

\subsection{Preliminaries}

\paragraph{Definitions} We call a function $f: \R^n \to \R^k$ piecewise affine if there exists a finite set of polytopes $\{ Q_r \}_{r=1}^R$, referred to as the \emph{linear regions} of $f$, such that $\cup_{r=1}^R Q_r = \R^n$ and $f \vert_{Q_r}$ is an affine function for every $Q_r$. ReLU networks are networks that result in piecewise affine classifier functions \citep{arora2018understanding}, which include networks with fully-connected, convolutional, and residual layers where just ReLU or leaky-ReLU are used as activation functions and max or average pooling are used in convolution layers. Let $\D := \{ \vx_i \in \R^n, t_i \}_{i=1}^m$ be a dataset, where the targets are $t_i \in \{ 0, 1 \}$ or $t_i \in \{ 1, \dots, k \}$ for the binary and multi-class case, respectively. Let $\phi: \R^n \to \R^d$ be an arbitrary fixed feature map and write $\vphi := \phi(\vx)$ for a given $\vx$. We define the logistic (\emph{sigmoid}) function as $\sigma(z) := 1/(1 + \exp(-z))$ for $z \in \R$ and the softmax function as $\mathrm{softmax}(\vz, i) := \exp(z_i)/\sum_{j} \exp(z_j)$ for $\vz \in \R^k$ and $i \in \{ 1, \dots, k \}$.
Given a neural network $f_\vtheta$, we consider the distribution $p(\vtheta \vert \D)$ over its parameters. Note that even though we use the notation $p(\vtheta \vert \D)$, we do \emph{not} require this distribution to be a posterior distribution in the Bayesian sense.
The \emph{predictive} distribution for the binary case is
\begin{align} \label{eq:two_class_pred}
  p(y = 1 \vert \vx, \D) &= \int \sigma(f_\vtheta(\vx)) \, p(\vtheta \vert \D) \, d\vtheta \, , \\
\shortintertext{and for the multi-class case}
  \label{eq:multi_class_pred}
  p(y = i \vert \vx, \D) &= \int \mathrm{softmax}(f_\vtheta(\vx), i) \, p(\vtheta \vert \D) \, d\vtheta \, .
\end{align}
The functions $\lambda_i(\cdot), \lambda_\text{max}(\cdot)$, and $\lambda_\text{min}(\cdot)$ return the $i$th, maximum, and minimum eigenvalue (which are assumed to exist) of their matrix argument, respectively.\footnote{We assume they are sorted in a descending order.} Similarly for the function $s_i(\cdot)$, $s_\text{max}(\cdot)$, and $s_\text{min}(\cdot)$ which return singular values instead. Finally, we assume that $\Vert \cdot \Vert$ is the $\ell^2$ norm.

\vspace{-0.5em}

\paragraph{Problem statement} The following theorem from \citet{hein2019relu} shows that ReLU networks exhibit arbitrarily high confidence far away from the training data: If a point $\vx \in \R^n$ is scaled by a sufficiently large scalar $\delta > 0$, the input $\delta \vx$ attains arbitrarily high confidence.

\vspace{0.25em}

\begin{theorem}[\citeauthor{hein2019relu}, \citeyear{hein2019relu}] \label{thm:hein_relu}
  Let $\R^d = \cup^R_{r=1} Q_r$ and $f \vert_{Q_r} (\vx) = \mU_r \vx + \vc_r$ be the piecewise affine representation of the output of a ReLU network on $Q_r$. Suppose that $\mU_r$ does not contain identical rows for all $r = 1, \dots, R$, then for almost any $\vx \in \R^n$ and any $\epsilon > 0$, there exists a $\delta > 0$ and a class $i \in \{ 1, \dots, k \}$ such that it holds $\softmax(f(\delta \vx), i) \geq 1 - \epsilon$. Moreover, $\lim_{\delta \to \infty} \softmax(f(\delta \vx), i) = 1$. \qed
\end{theorem}


It is standard to treat neural networks as probabilistic models of the conditional distribution $p(y \vert \vx, \vtheta)$ over the prediction $y$. In this case, we define the confidence of any input point $\vx$ as the maximum predictive probability, which in the case of a binary problem, can be written as
$\max_{i \in \{ 0, 1 \}} \, p(y = i | \vx, \vtheta) = \sigma(\abs{f_\vtheta(\vx)})$. Standard training involves assigning a maximum a posteriori (MAP) estimate $\vtheta_\text{MAP}$ to the weights, ignoring potential uncertainty on $\vtheta$. We will show that this lack of uncertainty is the primary cause of the overconfidence discussed by \citet{hein2019relu} and argue that it can be mitigated by considering the marginalized prediction in \eqref{eq:two_class_pred} instead.

Even for a linear classifier parametrized by a single weight matrix $\vtheta = \vw$, there is generally no analytic solution for \eqref{eq:two_class_pred}. But, good approximations exist when the distribution over the weights is a Gaussian $p(\vw \vert \D) \approx \N(\vw | \vmu, \b{\Sigma})$ with mean $\vmu$ and covariance $\b{\Sigma}$. One such approximation \citep{spiegelhalter1990sequential,mackay1992evidence} is constructed by scaling the input of the \emph{probit} function\footnote{The probit function $\Phi$ is another sigmoid, the distribution function (CDF) of the standard Gaussian.} $\Phi$ by a constant $\lambda=\sqrt{\pi/8}$
. Using this approximation and the Gaussian assumption, if we let $a := \vw^\top \vphi$, we get
\begin{align}
    \label{eq:pred_approx}
  p(y = 1 \vert \vx, \D) &\approx \int \Phi(\sqrt{\pi/8} \, a) \, \N(a \vert \vmu^\top \vphi, \vphi^\top \b{\Sigma} \vphi) \, da \nonumber \\
      &= \Phi \left( \frac{\vmu^\top \vphi}{\sqrt{8/\pi + \vphi^\top \b{\Sigma} \vphi}} \right) \approx \sigma \left(z(\vx) \right) \, ,
\end{align}
where the last step uses the approximation $\Phi(\sqrt{\pi/8} \, x) \approx \sigma(x)$ a second time, with
\begin{equation}
    \label{eq:z}
    z(\vx) := \frac{\vmu^\top \vphi}{\sqrt{1 + \pi/8 \, \vphi^\top \b{\Sigma} \vphi}} \, .
\end{equation}
In the case of $\vmu = \vw_\text{MAP}$, \Cref{eq:pred_approx} can be seen as the ``softened'' version of the MAP prediction of the classifier, using the covariance of the Gaussian. The confidence in this case is $\max_{i \in \{ 0, 1 \}} \, p(y = i | \vx, \D) = \sigma(\abs{z(\vx)})$.

We can generalize the previous insight to the case where the parameters of the feature map $\phi$ are also approximated by a Gaussian. Let $\vtheta \in \R^p$ be the parameter vector of a NN $f_\vtheta: \R^n \to \R$ with a given Gaussian approximation $p(\vtheta \vert \D) \approx \N(\vtheta \vert \vmu, \mSigma)$. Let $\vx \in \R^n$ be an arbitrary input point. Letting $\vd := \nabla f_\vtheta(\vx) \vert_\vmu$, we do a first-order Taylor expansion of $f_\vtheta$ at $\vmu$ \citep{mackay1995probable}: $f_\vtheta(\vx) \approx f_\vmu(\vx) + \vd^\top (\vtheta - \vmu)$.
This implies that the distribution over $f_\vtheta(\vx)$ is given by $p(f_\vtheta(\vx) \vert \vx, \D) \approx \N(f_\vtheta(\vx) \vert f_\vmu(\vx), \vd^\top \mSigma \vd)$. Therefore, we have
\begin{align} \label{eq:z_full}
  z(\vx) := \frac{f_\vmu(\vx)}{\sqrt{1 + \pi/8 \, \vd^\top \mSigma \vd}} \, .
\end{align}
It is easy to see that \eqref{eq:z} is indeed a special case of \eqref{eq:z_full}.

As the first notable property of this approximation, we show that, in contrast to some other methods for uncertainty quantification (e.g.~Monte Carlo dropout, \citeauthor{gal2016dropout}, \citeyear{gal2016dropout}) it preserves the decision boundary induced by the MAP estimate.

\vspace{0.25em}

\begin{restatable}[Invariance property]{proposition}{proptwotwo}
  \label{prop:same_dec_bdry}
  Let $f_\vtheta: \R^n \to \R$ be a binary classifier network parametrized by $\vtheta$ and let $\N(\vtheta \vert \vmu, \b{\Sigma})$ be the distribution over $\vtheta$. Then for any $\vx \in \R^n$, we have $\sigma(f_{\vmu}(\vx)) = 0.5$ if and only if $\sigma(z(\vx)) = 0.5$.
\end{restatable}

This property is useful in practice, particularly whenever $\vmu = \vtheta_\text{MAP}$, since it guarantees that employing a Gaussian approximation on top of a MAP-trained network will not reduce the original classification accuracy. Virtually all state-of-the-art models in deep learning are trained via MAP estimation and sacrificing the classification performance that makes them attractive in the first place would be a waste.

\subsection{Main Results}
\label{subsec:main_results}

\begin{figure}[t]
  \centering
  \input{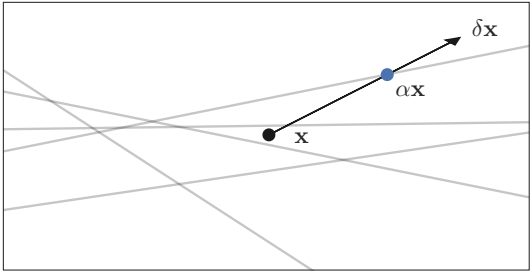}

  \caption{The situation in \Cref{thm:full_bounded_confidence_affine_relu,thm:bounded_confidence_affine_relu}. The intersections of gray lines are the linear regions.}
  \label{fig:nonasymp_situation}
\end{figure}

As our central theoretical contribution, we show that for any $\vx \in \R^n$, as $\delta \to \infty$, the value of $\abs{z(\delta \vx)}$ in \eqref{eq:z_full} goes to a quantity that only depends on the mean and covariance of the Gaussian over the weights. Moreover, this property also holds in the finite asymptotic regime, far enough from the training data. This result implies that one can drive the confidence closer to the uniform (one-half) far away from the training points by shifting $\abs{z(\vx)}$ closer to zero by controlling the Gaussian. We formalize this result in the following theorem. The situation is illustrated in \Cref{fig:nonasymp_situation}.

\vspace{0.25em}

\begin{restatable}[All-layer approximation]{theorem}{thmtwothree}
\label{thm:full_bounded_confidence_affine_relu}
  Let $f_\vtheta: \R^n \to \R$ be a binary ReLU classification network parametrized by $\vtheta \in \R^p$ with $p \geq n$, and let $\N(\vtheta \vert \vmu, \b{\Sigma})$ be the Gaussian approximation over the parameters. Then for any input $\vx \in \R^n$,
  \begin{equation}\label{eq:bound_relu_asymp_full}
    \lim_{\delta \to \infty} \sigma(\abs{z(\delta \vx)}) \leq \sigma \left( \frac{\norm{\vu}}{s_\text{\emph{min}} \left( \mJ \right)\sqrt{\pi/8 \, \lambda_\text{\emph{min}}(\b{\Sigma})}} \right) \, ,
  \end{equation}
  where $\vu \in \R^n$ is a vector depending only on $\vmu$ and the $n \times p$ matrix $\mJ := \left. \frac{\partial \vu}{\partial \vtheta} \right\rvert_\vmu$ is the Jacobian of $\vu$ w.r.t. $\vtheta$ at $\vmu$. Moreover, if $f_\vtheta$ has no bias parameters, then there exists $\alpha > 0$ such that for any $\delta \geq \alpha$, we have that
  \begin{equation*}
    \sigma(\abs{z(\delta \vx)}) \leq \lim_{\delta \to \infty} \sigma(\abs{z(\delta \vx)}) \, .
  \end{equation*}
\end{restatable}

The following question is practically interesting: Do we have to construct a probabilistic (Gaussian) uncertainty to the whole ReLU network for the previous property (\Cref{thm:full_bounded_confidence_affine_relu}) to hold? Surprisingly, the answer is no. The following theorem establishes that a guarantee similar to \Cref{thm:full_bounded_confidence_affine_relu}, is feasible even if \emph{only the last layer's weights} are assigned a Gaussian distribution. This amounts to a form of Bayesian logistic regression, where the features are provided by the ReLU network.

\vspace{0.25em}

\begin{restatable}[Last-layer approximation]{theorem}{thmtwofour}
\label{thm:bounded_confidence_affine_relu}
  Let $g: \R^d \to \R$ be a binary linear classifier defined by $g(\phi(\vx)) := \vw^\top \phi(\vx)$ where $\phi: \R^n \to \R^d$ is a fixed ReLU network and let $\N(\vw \vert \vmu, \b{\Sigma})$ be the Gaussian approximation over the last-layer's weights. Then for any input $\vx \in \R^n$,
  \begin{equation}\label{eq:bound_relu_asymp}
    \lim_{\delta \to \infty} \sigma(\abs{z(\delta \vx)}) \leq \sigma \left( \frac{\norm{\vmu}}{\sqrt{\pi/8 \, \lambda_\text{\emph{min}}(\b{\Sigma})}} \right) \, .
  \end{equation}
  Moreover, if $\phi$ has no bias parameters, then there exists $\alpha > 0$ such that for any $\delta \geq \alpha$, we have that
  \begin{equation*}
    \sigma(\abs{z(\delta \vx)}) \leq \lim_{\delta \to \infty} \sigma(\abs{z(\delta \vx)}) \, .
  \end{equation*}
  %
\end{restatable}

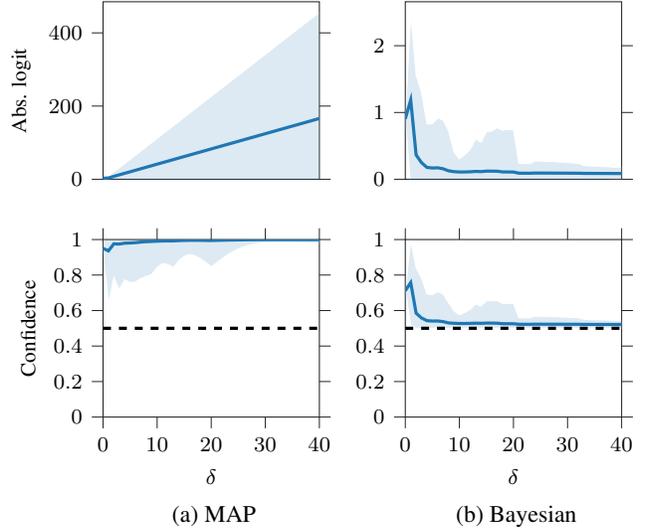
\begin{figure}[t]
  \centering

  \hspace{2.5em}
  \subfloat{
\begin{tikzpicture}[trim axis left, trim axis right]

\tikzstyle{every node}=[font=\fontsize{8}{9}\selectfont]

\definecolor{color0}{rgb}{0.12156862745098,0.466666666666667,0.705882352941177}

\begin{axis}[
width=0.26\textwidth,
height=0.23\textwidth,
axis line style={white!15.0!black},
tick align=outside,
x grid style={white!80.0!black},
xmajorticks=false,
xmin=0, xmax=40,
xtick style={color=white!15.0!black},
y grid style={white!80.0!black},
ylabel={Abs. logit},
ymin=0, ymax=484.985790252686,
ytick style={color=white!15.0!black}
]
\path [draw=white, fill=color0, opacity=0.15]
(axis cs:0,2.94754600524902)
--(axis cs:0,2.94754314422607)
--(axis cs:1,-0.767542362213135)
--(axis cs:2,-3.69651985168457)
--(axis cs:3,-7.02665328979492)
--(axis cs:4,-10.2227935791016)
--(axis cs:5,-13.4786949157715)
--(axis cs:6,-16.6781349182129)
--(axis cs:7,-19.8060913085938)
--(axis cs:8,-22.9405212402344)
--(axis cs:9,-26.0917167663574)
--(axis cs:10,-29.2240219116211)
--(axis cs:11,-32.371696472168)
--(axis cs:12,-35.5400886535645)
--(axis cs:13,-38.7061195373535)
--(axis cs:14,-41.8264007568359)
--(axis cs:15,-44.9658317565918)
--(axis cs:16,-48.1151123046875)
--(axis cs:17,-51.2697830200195)
--(axis cs:18,-54.4349517822266)
--(axis cs:19,-57.6080703735352)
--(axis cs:20,-60.7841873168945)
--(axis cs:21,-63.9270324707031)
--(axis cs:22,-67.0752029418945)
--(axis cs:23,-70.2225799560547)
--(axis cs:24,-73.3713760375977)
--(axis cs:25,-76.5224609375)
--(axis cs:26,-79.6749572753906)
--(axis cs:27,-82.8280258178711)
--(axis cs:28,-85.9840469360352)
--(axis cs:29,-89.1404724121094)
--(axis cs:30,-92.2972183227539)
--(axis cs:31,-95.454345703125)
--(axis cs:32,-98.6116790771484)
--(axis cs:33,-101.769714355469)
--(axis cs:34,-104.927963256836)
--(axis cs:35,-108.086456298828)
--(axis cs:36,-111.244979858398)
--(axis cs:37,-114.402542114258)
--(axis cs:38,-117.560028076172)
--(axis cs:39,-120.717407226562)
--(axis cs:40,-123.874740600586)
--(axis cs:40,455.992431640625)
--(axis cs:40,455.992431640625)
--(axis cs:39,444.515869140625)
--(axis cs:38,433.039337158203)
--(axis cs:37,421.562744140625)
--(axis cs:36,410.086181640625)
--(axis cs:35,398.609100341797)
--(axis cs:34,387.132080078125)
--(axis cs:33,375.6552734375)
--(axis cs:32,364.1787109375)
--(axis cs:31,352.702697753906)
--(axis cs:30,341.226959228516)
--(axis cs:29,329.751495361328)
--(axis cs:28,318.276397705078)
--(axis cs:27,306.801727294922)
--(axis cs:26,295.329132080078)
--(axis cs:25,283.857055664062)
--(axis cs:24,272.386047363281)
--(axis cs:23,260.916320800781)
--(axis cs:22,249.447204589844)
--(axis cs:21,237.9765625)
--(axis cs:20,226.509765625)
--(axis cs:19,215.020263671875)
--(axis cs:18,203.530349731445)
--(axis cs:17,192.043731689453)
--(axis cs:16,180.562515258789)
--(axis cs:15,169.084609985352)
--(axis cs:14,157.613174438477)
--(axis cs:13,146.155059814453)
--(axis cs:12,134.666763305664)
--(axis cs:11,123.176445007324)
--(axis cs:10,111.698219299316)
--(axis cs:9,100.227890014648)
--(axis cs:8,88.7461547851562)
--(axis cs:7,77.2740631103516)
--(axis cs:6,65.8065795898438)
--(axis cs:5,54.291187286377)
--(axis cs:4,42.7452621459961)
--(axis cs:3,31.2342872619629)
--(axis cs:2,19.6021251678467)
--(axis cs:1,7.87259721755981)
--(axis cs:0,2.94754600524902)
--cycle;

\addplot [very thick, color0]
table {%
0 2.94754457473755
1 3.55252742767334
2 7.95280265808105
3 12.103816986084
4 16.2612342834473
5 20.4062461853027
6 24.5642242431641
7 28.7339859008789
8 32.9028167724609
9 37.0680885314941
10 41.2370986938477
11 45.4023742675781
12 49.5633354187012
13 53.7244682312012
14 57.8933868408203
15 62.0593910217285
16 66.2237014770508
17 70.3869705200195
18 74.5476989746094
19 78.7060928344727
20 82.8627853393555
21 87.0247650146484
22 91.1859970092773
23 95.3468780517578
24 99.5073318481445
25 103.667297363281
26 107.827087402344
27 111.986854553223
28 116.146171569824
29 120.305511474609
30 124.464866638184
31 128.624176025391
32 132.783508300781
33 136.942779541016
34 141.102066040039
35 145.261322021484
36 149.420608520508
37 153.580093383789
38 157.739654541016
39 161.899230957031
40 166.058853149414
};
\end{axis}

\end{tikzpicture}}
  \qquad  \quad
  \subfloat{
\begin{tikzpicture}[trim axis left, trim axis right]

\tikzstyle{every node}=[font=\fontsize{8}{9}\selectfont]

\definecolor{color0}{rgb}{0.12156862745098,0.466666666666667,0.705882352941177}

\begin{axis}[
width=0.26\textwidth,
height=0.23\textwidth,
axis line style={white!15.0!black},
tick align=outside,
x grid style={white!80.0!black},
xmajorticks=false,
xmin=0, xmax=40,
xtick style={color=white!15.0!black},
y grid style={white!80.0!black},
ymin=0, ymax=2.65813687443733,
ytick style={color=white!15.0!black}
]
\path [draw=white, fill=color0, opacity=0.15]
(axis cs:0,0.90503865480423)
--(axis cs:0,0.905038297176361)
--(axis cs:1,-0.120920419692993)
--(axis cs:2,-0.830900549888611)
--(axis cs:3,-0.806322395801544)
--(axis cs:4,-0.473646849393845)
--(axis cs:5,-0.494460582733154)
--(axis cs:6,-0.576881766319275)
--(axis cs:7,-0.572706282138824)
--(axis cs:8,-0.471340209245682)
--(axis cs:9,-0.181344792246819)
--(axis cs:10,-0.0889140740036964)
--(axis cs:11,-0.148112386465073)
--(axis cs:12,-0.222506776452065)
--(axis cs:13,-0.370525062084198)
--(axis cs:14,-0.3226038813591)
--(axis cs:15,-0.469069272279739)
--(axis cs:16,-0.478437870740891)
--(axis cs:17,-0.532117664813995)
--(axis cs:18,-0.515885174274445)
--(axis cs:19,-0.527222573757172)
--(axis cs:20,-0.525423049926758)
--(axis cs:21,-0.0572653412818909)
--(axis cs:22,-0.0561226010322571)
--(axis cs:23,-0.0543845817446709)
--(axis cs:24,-0.0890093892812729)
--(axis cs:25,-0.086185559630394)
--(axis cs:26,-0.0827085748314857)
--(axis cs:27,-0.082223929464817)
--(axis cs:28,-0.0780405551195145)
--(axis cs:29,-0.0734783634543419)
--(axis cs:30,-0.0686720237135887)
--(axis cs:31,-0.062778502702713)
--(axis cs:32,-0.058055967092514)
--(axis cs:33,-0.0289075076580048)
--(axis cs:34,-0.0248431488871574)
--(axis cs:35,-0.0207791030406952)
--(axis cs:36,-0.0173951387405396)
--(axis cs:37,-0.0142026916146278)
--(axis cs:38,-0.0115715935826302)
--(axis cs:39,-0.00343369692564011)
--(axis cs:40,-0.00890735536813736)
--(axis cs:40,0.181839793920517)
--(axis cs:40,0.181839793920517)
--(axis cs:39,0.175533026456833)
--(axis cs:38,0.185174286365509)
--(axis cs:37,0.188311547040939)
--(axis cs:36,0.19216912984848)
--(axis cs:35,0.196110039949417)
--(axis cs:34,0.200826644897461)
--(axis cs:33,0.205504238605499)
--(axis cs:32,0.238931864500046)
--(axis cs:31,0.244415730237961)
--(axis cs:30,0.251204371452332)
--(axis cs:29,0.256787419319153)
--(axis cs:28,0.262108862400055)
--(axis cs:27,0.267051041126251)
--(axis cs:26,0.267480194568634)
--(axis cs:25,0.271640002727509)
--(axis cs:24,0.275189161300659)
--(axis cs:23,0.234663903713226)
--(axis cs:22,0.237543761730194)
--(axis cs:21,0.239486753940582)
--(axis cs:20,0.741776704788208)
--(axis cs:19,0.74538642168045)
--(axis cs:18,0.734944522380829)
--(axis cs:17,0.77313095331192)
--(axis cs:16,0.721214294433594)
--(axis cs:15,0.712766230106354)
--(axis cs:14,0.551881194114685)
--(axis cs:13,0.60555511713028)
--(axis cs:12,0.445897996425629)
--(axis cs:11,0.366587668657303)
--(axis cs:10,0.306411653757095)
--(axis cs:9,0.408978044986725)
--(axis cs:8,0.723684310913086)
--(axis cs:7,0.886318266391754)
--(axis cs:6,0.919819355010986)
--(axis cs:5,0.830975532531738)
--(axis cs:4,0.828846454620361)
--(axis cs:3,1.30324387550354)
--(axis cs:2,1.5594300031662)
--(axis cs:1,2.49199223518372)
--(axis cs:0,0.90503865480423)
--cycle;

\addplot [very thick, color0]
table {%
0 0.905038475990295
1 1.18553590774536
2 0.364264726638794
3 0.248460724949837
4 0.17759981751442
5 0.168257459998131
6 0.171468794345856
7 0.156805977225304
8 0.126172035932541
9 0.113816633820534
10 0.108748786151409
11 0.109237633645535
12 0.111695602536201
13 0.11751501262188
14 0.114638656377792
15 0.121848486363888
16 0.121388226747513
17 0.120506636798382
18 0.109529696404934
19 0.109081946313381
20 0.108176797628403
21 0.0911107063293457
22 0.0907105803489685
23 0.090139664709568
24 0.0930898785591125
25 0.0927272140979767
26 0.0923858061432838
27 0.0924135521054268
28 0.0920341461896896
29 0.0916545316576958
30 0.0912661775946617
31 0.0908186137676239
32 0.0904379487037659
33 0.0882983654737473
34 0.087991751730442
35 0.087665468454361
36 0.0873869955539703
37 0.0870544239878654
38 0.0868013501167297
39 0.0860496684908867
40 0.0864662155508995
};
\end{axis}

\end{tikzpicture}}

  \setcounter{subfigure}{0}

  \hspace{2.5em}
  \subfloat[MAP]{
\begin{tikzpicture}[trim axis left, trim axis right]

\tikzstyle{every node}=[font=\fontsize{8}{9}\selectfont]

\definecolor{color0}{rgb}{0.12156862745098,0.466666666666667,0.705882352941177}

\begin{axis}[
width=0.26\textwidth,
height=0.23\textwidth,
axis line style={white!15.0!black},
tick align=outside,
x grid style={white!80.0!black},
xlabel={\(\displaystyle \delta\)},
xmin=0, xmax=40,
xtick style={color=white!15.0!black},
y grid style={white!80.0!black},
ylabel={Confidence},
ymin=0, ymax=1,
ytick style={color=white!15.0!black}
]
\path [draw=white, fill=color0, opacity=0.15]
(axis cs:0,0.950148344039917)
--(axis cs:0,0.950146913528442)
--(axis cs:1,0.635052442550659)
--(axis cs:2,0.787433624267578)
--(axis cs:3,0.71764862537384)
--(axis cs:4,0.773244738578796)
--(axis cs:5,0.756262302398682)
--(axis cs:6,0.760955929756165)
--(axis cs:7,0.78092497587204)
--(axis cs:8,0.792203783988953)
--(axis cs:9,0.803522348403931)
--(axis cs:10,0.84212589263916)
--(axis cs:11,0.865508437156677)
--(axis cs:12,0.858246743679047)
--(axis cs:13,0.842653870582581)
--(axis cs:14,0.880234599113464)
--(axis cs:15,0.905740022659302)
--(axis cs:16,0.914998114109039)
--(axis cs:17,0.908735990524292)
--(axis cs:18,0.892238974571228)
--(axis cs:19,0.870440006256104)
--(axis cs:20,0.846281409263611)
--(axis cs:21,0.868497550487518)
--(axis cs:22,0.892198979854584)
--(axis cs:23,0.91359007358551)
--(axis cs:24,0.932008862495422)
--(axis cs:25,0.947182416915894)
--(axis cs:26,0.959185242652893)
--(axis cs:27,0.96827358007431)
--(axis cs:28,0.974794924259186)
--(axis cs:29,0.979130268096924)
--(axis cs:30,0.981694042682648)
--(axis cs:31,0.9829341173172)
--(axis cs:32,0.983279049396515)
--(axis cs:33,0.983069241046906)
--(axis cs:34,0.982530355453491)
--(axis cs:35,0.981794536113739)
--(axis cs:36,0.980933129787445)
--(axis cs:37,0.979982912540436)
--(axis cs:38,0.978961825370789)
--(axis cs:39,0.978045701980591)
--(axis cs:40,0.977132499217987)
--(axis cs:40,1.0213371515274)
--(axis cs:40,1.0213371515274)
--(axis cs:39,1.02047157287598)
--(axis cs:38,1.01959908008575)
--(axis cs:37,1.01862323284149)
--(axis cs:36,1.01770639419556)
--(axis cs:35,1.01686370372772)
--(axis cs:34,1.01612746715546)
--(axis cs:33,1.0155633687973)
--(axis cs:32,1.01529550552368)
--(axis cs:31,1.0155394077301)
--(axis cs:30,1.01662111282349)
--(axis cs:29,1.0189516544342)
--(axis cs:28,1.02295660972595)
--(axis cs:27,1.02902412414551)
--(axis cs:26,1.03750574588776)
--(axis cs:25,1.04871904850006)
--(axis cs:24,1.06289517879486)
--(axis cs:23,1.08008623123169)
--(axis cs:22,1.10002505779266)
--(axis cs:21,1.12207913398743)
--(axis cs:20,1.14260792732239)
--(axis cs:19,1.11965799331665)
--(axis cs:18,1.0987673997879)
--(axis cs:17,1.08267056941986)
--(axis cs:16,1.07610261440277)
--(axis cs:15,1.08423435688019)
--(axis cs:14,1.10768139362335)
--(axis cs:13,1.14200031757355)
--(axis cs:12,1.12633430957794)
--(axis cs:11,1.11855721473694)
--(axis cs:10,1.13974571228027)
--(axis cs:9,1.17493152618408)
--(axis cs:8,1.18399810791016)
--(axis cs:7,1.19073712825775)
--(axis cs:6,1.20385122299194)
--(axis cs:5,1.20497155189514)
--(axis cs:4,1.1860545873642)
--(axis cs:3,1.2317978143692)
--(axis cs:2,1.16438853740692)
--(axis cs:1,1.23746192455292)
--(axis cs:0,0.950148344039917)
--cycle;

\addplot [very thick, color0]
table {%
0 0.95014762878418
1 0.936257183551788
2 0.97591108083725
3 0.974723219871521
4 0.979649662971497
5 0.980616927146912
6 0.982403576374054
7 0.985831022262573
8 0.988100945949554
9 0.989226937294006
10 0.990935802459717
11 0.992032825946808
12 0.992290556430817
13 0.992327094078064
14 0.993957996368408
15 0.994987189769745
16 0.995550394058228
17 0.995703279972076
18 0.995503187179565
19 0.995048999786377
20 0.994444668292999
21 0.995288372039795
22 0.996112048625946
23 0.9968381524086
24 0.997452020645142
25 0.997950732707977
26 0.998345494270325
27 0.998648822307587
28 0.998875737190247
29 0.999040961265564
30 0.999157547950745
31 0.999236762523651
32 0.999287247657776
33 0.999316275119781
34 0.999328911304474
35 0.999329090118408
36 0.999319791793823
37 0.999303042888641
38 0.999280452728271
39 0.999258637428284
40 0.999234795570374
};
\addplot [very thick, black, dashed]
table {%
0 0.5
40 0.5
};
\end{axis}

\end{tikzpicture}}
  \qquad \quad
  \subfloat[Bayesian]{
\begin{tikzpicture}[trim axis left, trim axis right]

\tikzstyle{every node}=[font=\fontsize{8}{9}\selectfont]

\definecolor{color0}{rgb}{0.12156862745098,0.466666666666667,0.705882352941177}

\begin{axis}[
width=0.26\textwidth,
height=0.23\textwidth,
axis line style={white!15.0!black},
tick align=outside,
x grid style={white!80.0!black},
xlabel={\(\displaystyle \delta\)},
xmin=0, xmax=40,
xtick style={color=white!15.0!black},
y grid style={white!80.0!black},
ymin=0, ymax=1,
ytick style={color=white!15.0!black}
]
\path [draw=white, fill=color0, opacity=0.15]
(axis cs:0,0.711984097957611)
--(axis cs:0,0.711983382701874)
--(axis cs:1,0.518743634223938)
--(axis cs:2,0.5)
--(axis cs:3,0.5)
--(axis cs:4,0.5)
--(axis cs:5,0.5)
--(axis cs:6,0.5)
--(axis cs:7,0.5)
--(axis cs:8,0.5)
--(axis cs:9,0.5)
--(axis cs:10,0.5)
--(axis cs:11,0.5)
--(axis cs:12,0.5)
--(axis cs:13,0.5)
--(axis cs:14,0.5)
--(axis cs:15,0.5)
--(axis cs:16,0.5)
--(axis cs:17,0.5)
--(axis cs:18,0.5)
--(axis cs:19,0.5)
--(axis cs:20,0.5)
--(axis cs:21,0.5)
--(axis cs:22,0.5)
--(axis cs:23,0.5)
--(axis cs:24,0.5)
--(axis cs:25,0.5)
--(axis cs:26,0.5)
--(axis cs:27,0.5)
--(axis cs:28,0.5)
--(axis cs:29,0.5)
--(axis cs:30,0.5)
--(axis cs:31,0.5)
--(axis cs:32,0.5)
--(axis cs:33,0.5)
--(axis cs:34,0.5)
--(axis cs:35,0.5)
--(axis cs:36,0.5)
--(axis cs:37,0.5)
--(axis cs:38,0.5)
--(axis cs:39,0.5)
--(axis cs:40,0.5)
--(axis cs:40,0.545360326766968)
--(axis cs:40,0.545360326766968)
--(axis cs:39,0.543805301189423)
--(axis cs:38,0.546179294586182)
--(axis cs:37,0.546947717666626)
--(axis cs:36,0.547891557216644)
--(axis cs:35,0.548851191997528)
--(axis cs:34,0.549997508525848)
--(axis cs:33,0.551128506660461)
--(axis cs:32,0.55924791097641)
--(axis cs:31,0.560564935207367)
--(axis cs:30,0.562188446521759)
--(axis cs:29,0.563519477844238)
--(axis cs:28,0.564784407615662)
--(axis cs:27,0.565957307815552)
--(axis cs:26,0.566066205501556)
--(axis cs:25,0.567052066326141)
--(axis cs:24,0.56789356470108)
--(axis cs:23,0.558016240596771)
--(axis cs:22,0.558706402778625)
--(axis cs:21,0.559177994728088)
--(axis cs:20,0.640162825584412)
--(axis cs:19,0.640627801418304)
--(axis cs:18,0.639881193637848)
--(axis cs:17,0.656255066394806)
--(axis cs:16,0.654749751091003)
--(axis cs:15,0.654612600803375)
--(axis cs:14,0.624900698661804)
--(axis cs:13,0.637109637260437)
--(axis cs:12,0.606755375862122)
--(axis cs:11,0.589744329452515)
--(axis cs:10,0.575875103473663)
--(axis cs:9,0.598283708095551)
--(axis cs:8,0.64487612247467)
--(axis cs:7,0.693272531032562)
--(axis cs:6,0.705569326877594)
--(axis cs:5,0.689133584499359)
--(axis cs:4,0.692926526069641)
--(axis cs:3,0.783103704452515)
--(axis cs:2,0.841618478298187)
--(axis cs:1,0.996139526367188)
--(axis cs:0,0.711984097957611)
--cycle;

\addplot [very thick, color0]
table {%
0 0.711983740329742
1 0.757441580295563
2 0.585110485553741
3 0.558108687400818
4 0.543187916278839
5 0.540744483470917
6 0.541114330291748
7 0.537480592727661
8 0.530260801315308
9 0.528254151344299
10 0.527114510536194
11 0.527178645133972
12 0.52766615152359
13 0.528686344623566
14 0.528109729290009
15 0.529342293739319
16 0.529175162315369
17 0.528640389442444
18 0.525923907756805
19 0.525753259658813
20 0.525542497634888
21 0.522734701633453
22 0.522635519504547
23 0.522494256496429
24 0.523212313652039
25 0.523123979568481
26 0.523041248321533
27 0.523048341274261
28 0.522956609725952
29 0.522864758968353
30 0.522770822048187
31 0.52266252040863
32 0.522570073604584
33 0.522048056125641
34 0.521973013877869
35 0.521892905235291
36 0.521824419498444
37 0.521742343902588
38 0.521679878234863
39 0.521494090557098
40 0.521596968173981
};
\addplot [very thick, black, dashed]
table {%
0 0.5
40 0.5
};
\end{axis}

\end{tikzpicture}}

  \caption{Absolute logit---$\abs{f_{\vtheta_\text{MAP}}(\delta \vx)}$ for MAP and $|z(\delta \vx)|$ for Bayesian---and confidence of the toy dataset in \Cref{fig:laplaces} as functions of $\delta$. Each plot shows the mean and $\pm 3$ standard deviation over the test set.}
  \label{fig:bounded_z}
\end{figure}

We show, using the same toy dataset and Gaussian-based last-layer Bayesian method as in \Cref{fig:laplaces}, an illustration of the previous results in \Cref{fig:bounded_z}. Confirming the findings, for each input $\vx$, the Gaussian approximation drives $|z(\delta \vx)|$ to a constant for sufficiently large $\delta$. Note that on true data points ($\delta = 1$), the confidences remain high and the convergence occurs at some finite $\delta$.

Taken together, the results above formally validate the usage of the common Gaussian approximations of the weights distribution, both in Bayesian \citep[etc.]{mackay1992practical,graves_practical_2011,blundell_weight_2015} or non-Bayesian \citep[etc.]{franchi2019tradi,lu2020uncertainty} fashions, on ReLU networks for mitigating overconfidence problems. Furthermore, \Cref{thm:bounded_confidence_affine_relu} shows that a full-blown Gaussian approximation or Bayesian treatment (i.e. on all layers of a NN) is not required to achieve control over the confidence far away from the training data. Put simply, even being ``just a bit Bayesian'' is enough to overcome at least asymptotic overconfidence.

We will show in the experiments (\Cref{subsec:ood_exp}) that the same Bayesian treatment also mitigates asymptotic confidence in the multi-class case. However, extending the theoretical analysis to this case is not straightforward, even with analytic approximations such as those by \citet{gibbs1998bayesian} and \citet{wu2018deterministic}.

\subsection{Laplace Approximations}
\label{subsec:bayesian_laplace_results}

The results in the previous section imply that the asymptotic confidence of a Gaussian-approximated binary ReLU classifier---either via a full or last-layer approximation---can be driven closer to uniform by controlling the covariance. In this section, we analyze the case when a Bayesian method in the form of a Laplace approximation is employed for obtaining the Gaussian. Although Laplace approximations are currently less popular than variational Bayes (VB), they have useful practical benefits: (i) they can be applied to any pre-trained network, (ii) whenever the approximation \eqref{eq:z_full} can be employed, \Cref{prop:same_dec_bdry} holds, and (iii) no re-training is needed. Indeed, Laplace approximations can be attractive to practitioners who already have a working MAP-trained network, but want to enhance its uncertainty estimates further without decreasing performance.

The principle of Laplace approximations is as follows. Let $p(\vtheta \vert \D) \propto p(\vtheta) \prod_{\vx, t \in \D} p(y = t \vert \vx, \vtheta)$ be the posterior of a network $f_\vtheta$. Then we can obtain a Gaussian approximation $p(\vtheta \vert \D) \approx \N(\vtheta \vert \vmu, \b{\Sigma})$ of the posterior by setting $\vmu = \vtheta_\text{MAP}$ and $\b{\Sigma} := (-\nabla^2 \log p(\vtheta \vert \D) \vert_{\vtheta_\text{MAP}})^\inv$, the inverse Hessian of the negative log-posterior at the mode. In the binary classification case, the likelihood $p(y \vert \vx, \vw)$ is assumed to be a Bernoulli distribution $\mathcal{B}(\sigma(f_\vtheta(\vx)))$. The prior $p(\vtheta)$ is assumed to be an isotropic Gaussian $\N(\vtheta \vert \b{0}, \sigma^2_0 \b{I})$.

While the prior variance $\sigma^2_0$ is tied to the MAP estimation (it can be derived from the weight decay), it is often treated as a separate hyperparameter and tuned after training \citep{ritter_scalable_2018}. This treatment is useful in the case when one has only a pre-trained network and not the original training hyperparameters. Under this situation, in the following proposition, we analyze the effect of $\sigma^2_0$ on the asymptotic confidence presented by \Cref{thm:full_bounded_confidence_affine_relu}. The statement for the last-layer case is analogous and presented in \Cref{appendix:proofs}.

\vspace{0.25em}

\begin{restatable}[All-layer Laplace]{proposition}{proptwofive}
\label{prop:laplace_varprior_full}
  Let $f_\vtheta$ be a binary ReLU classification network modeling a Bernoulli distribution $p(y | \vx, \vtheta) = \mathcal{B}(\sigma(f_\vtheta(\vx)))$ with parameter $\vtheta \in \R^p$. Let $\N(\vtheta | \vmu, \mSigma)$ be the posterior obtained via a Laplace approximation with prior $\N(\vtheta | \b{0}, \sigma^2_0 \mI)$, $\mH$ be the Hessian of the negative log-likelihood at $\vmu$, and $\mJ$ be the Jacobian as in \Cref{thm:full_bounded_confidence_affine_relu}. Then for any input $\vx \in \R^n$, the confidence $\sigma(\abs{z(\vx)})$ is a decreasing function of $\sigma^2_0$ with limits
  \begin{align*}
    \lim_{\sigma^2_0 \to \infty} \sigma(\abs{z(\vx)}) &\leq \sigma \left( \frac{\abs{f_\vmu(\vx)}}{1 + \sqrt{\pi/8 \, \lambda_\text{\emph{max}}(\mH) \Vert \mJ \vx \Vert^2}} \right) \\
    \lim_{\sigma^2_0 \to 0} \sigma(\abs{z(\vx)}) &= \sigma(\abs{f_\vmu(\vx)}) \, .
  \end{align*}
\end{restatable}

The result above shows that the ``far-away'' confidence decreases (up to some limit) as the prior variance increases. Meanwhile, we recover the far-away confidence induced by the MAP estimate as the prior variance goes to zero. One could therefore pick a value of $\sigma_0^2$ as high as possible for mitigating overconfidence. However, this is undesirable since it also lowers the confidence of the training data and test data around them (i.e. the so-called \emph{in-distribution data}), thus, causing underconfident predictions. Another common way to set this hyperparameter is by maximizing the validation log-likelihood \citep{ritter_scalable_2018}. This is also inadequate for our purpose since it only considers points close to the training data.

Inspired by \citet{hendrycks2018deep} and \citet{hein2019relu}, we simultaneously prefer high confidence on the in-distribution validation set and low confidence (high entropy) on the out-of-distribution validation set. Let $\hat{\D} := \{ \hat{\vx}_i, \hat{t}_i \}_{i=1}^m$ be a validation set and $\tilde{\D} := \{ \tilde{\vx}_i \}_{i=1}^m$ be an out-of-distribution dataset. We then pick the optimal $\sigma_0^2$ by solving the following one-parameter optimization problem:
\begin{align} \label{eq:opt_var_prior}
  \argmin_{\sigma_0^2} -\frac{1}{m} \sum_{i=1}^m \log p(y = \hat{t}_i | \hat{\vx}_i, \D) + \lambda H[p(y | \tilde{\vx}_i, \D)] \, ,
\end{align}
where $\lambda \in [0, 1]$ is controlling the trade-off between both terms. The first term in \eqref{eq:opt_var_prior} is the standard cross-entropy loss over $\hat{\D}$ while the second term is the negative predictive entropy over $\tilde{\D}$. Alternatively, the second term can be replaced by the cross-entropy loss where the target is the uniform probability vector. In all our experiments, we simply assume that $\tilde{\D}$ is a collection of uniform noise in the input space.

\section{Related Work}
\label{sec:related}

\begin{figure*}[t]
    \centering

    \subfloat{\includegraphics[width=0.22\textwidth, height=0.15\textwidth]{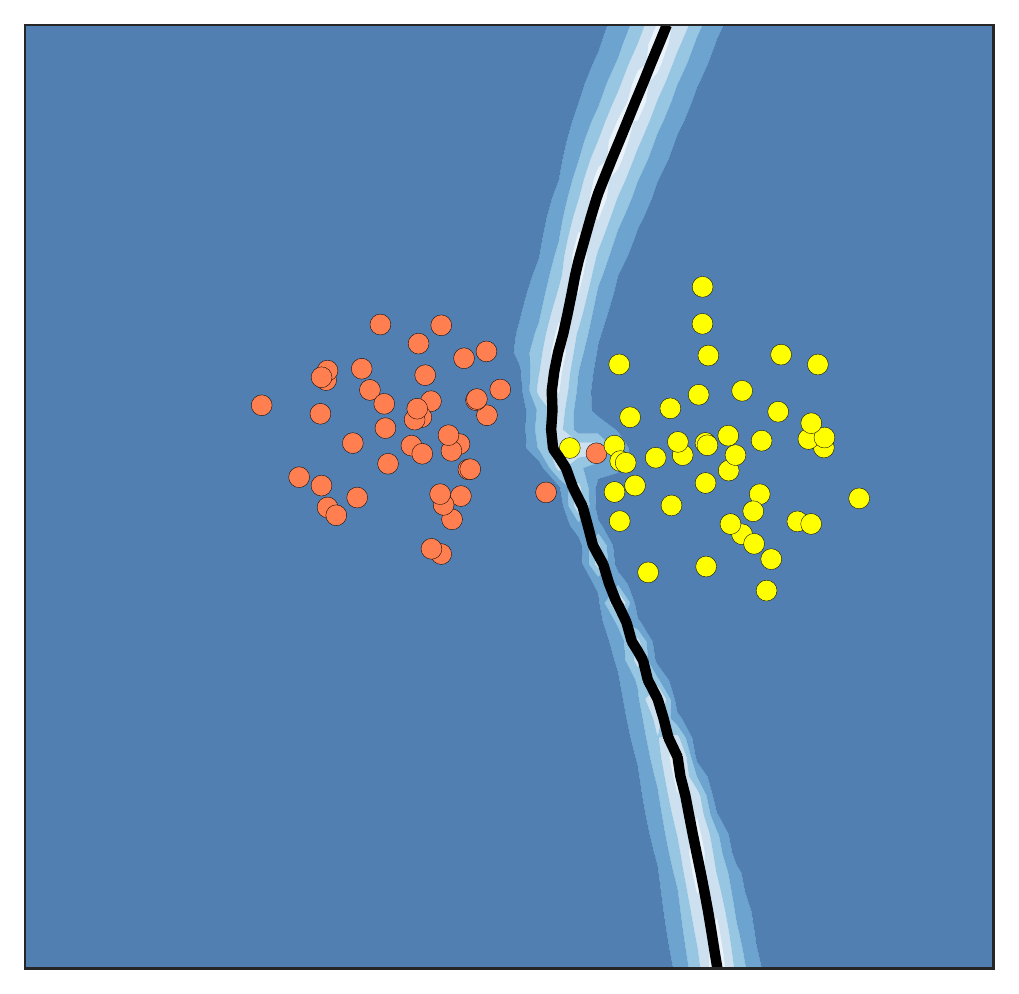}}
    \quad
    \subfloat{\includegraphics[width=0.22\textwidth, height=0.15\textwidth]{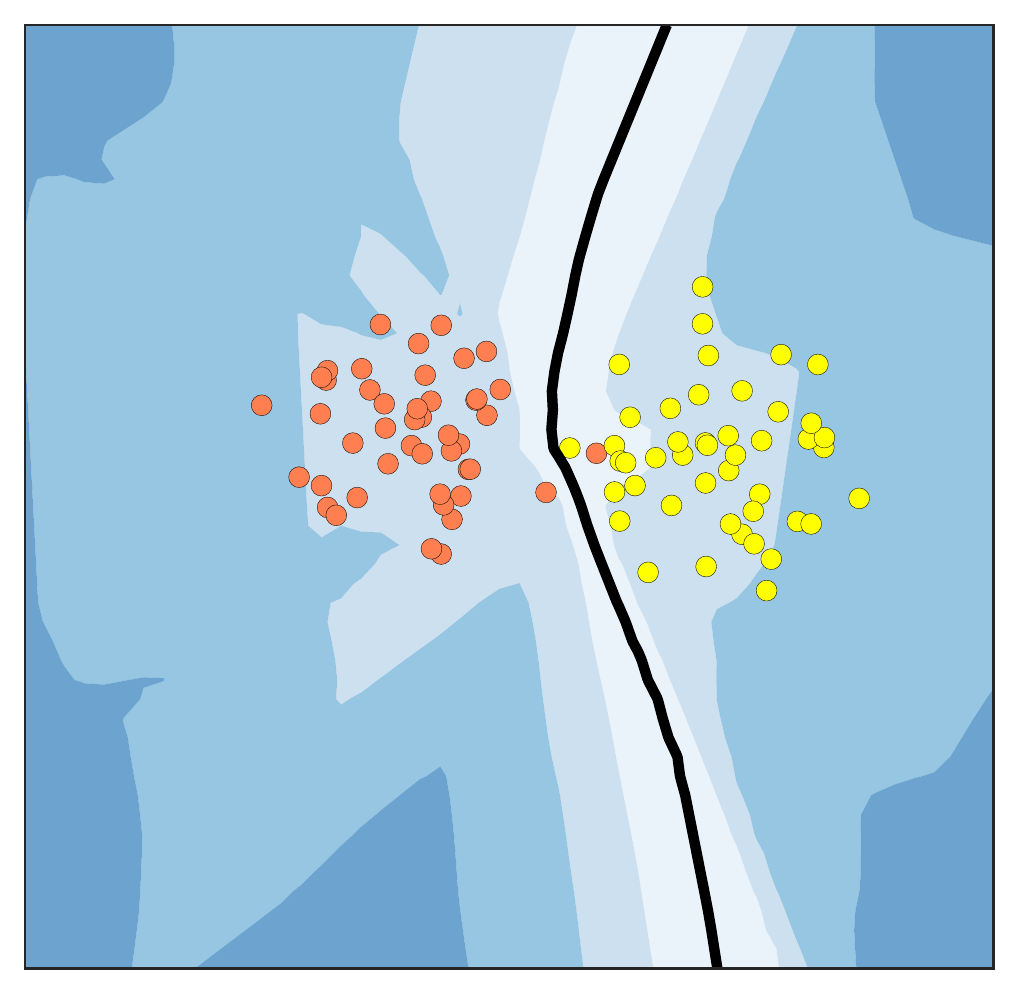}}
    \quad
    \subfloat{\includegraphics[width=0.22\textwidth, height=0.15\textwidth]{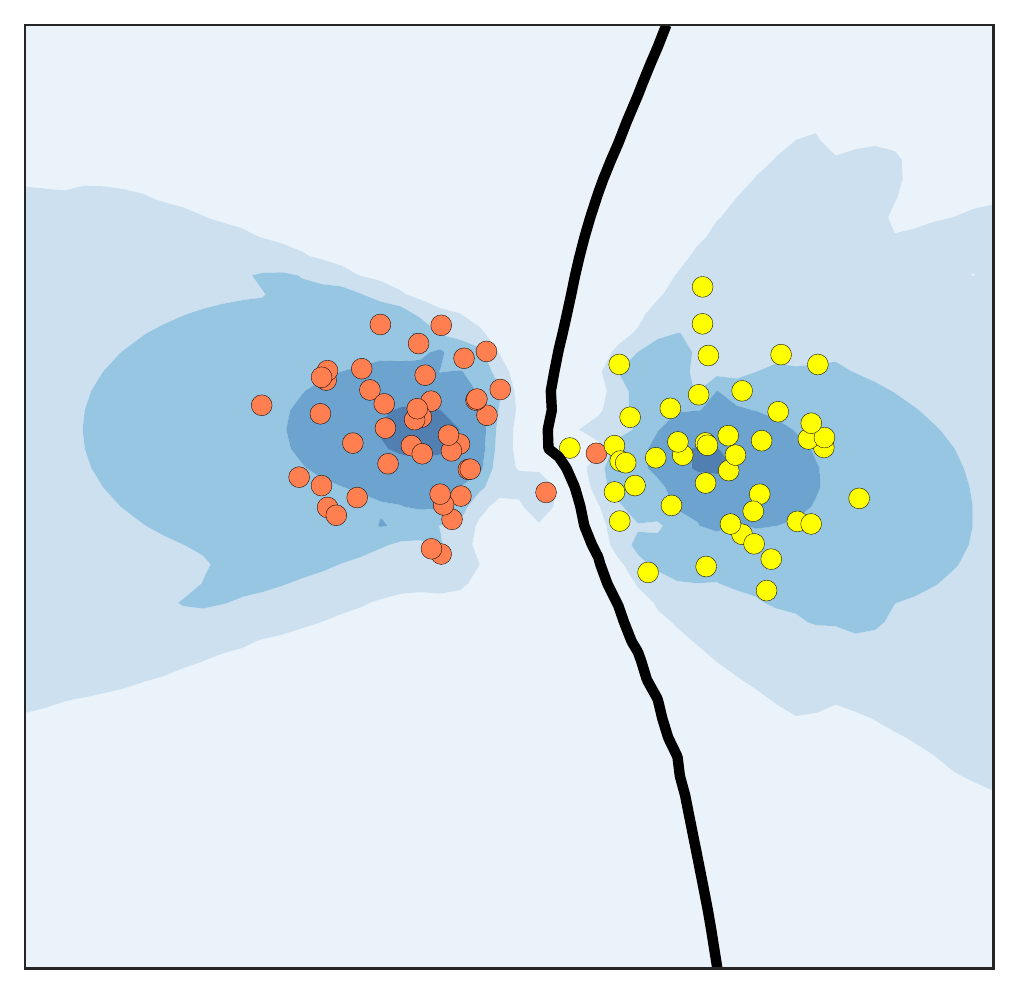}}
    \quad
    \subfloat{\includegraphics[width=0.22\textwidth, height=0.15\textwidth]{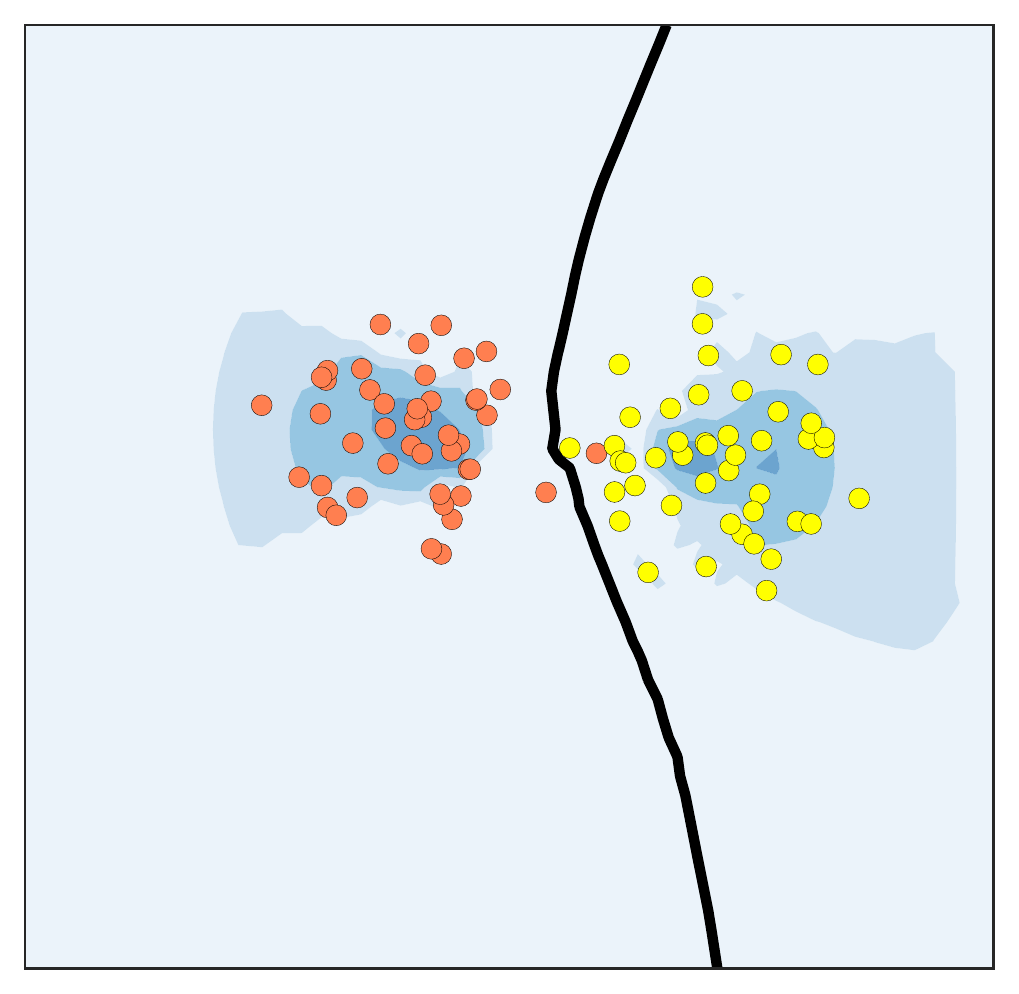}}
    \quad
    \subfloat{\includegraphics[width=0.03\textwidth, height=0.15\textwidth]{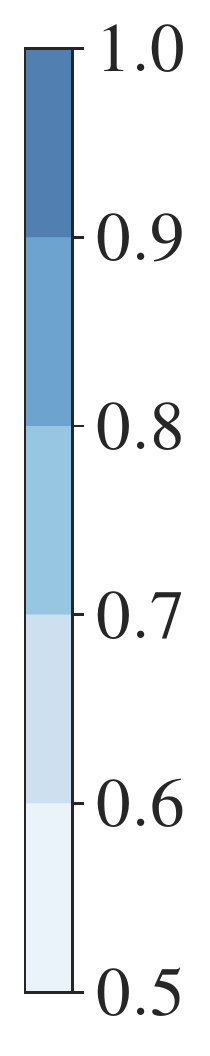}}

    \setcounter{subfigure}{0}

    \subfloat[MAP]{\includegraphics[width=0.22\textwidth, height=0.15\textwidth]{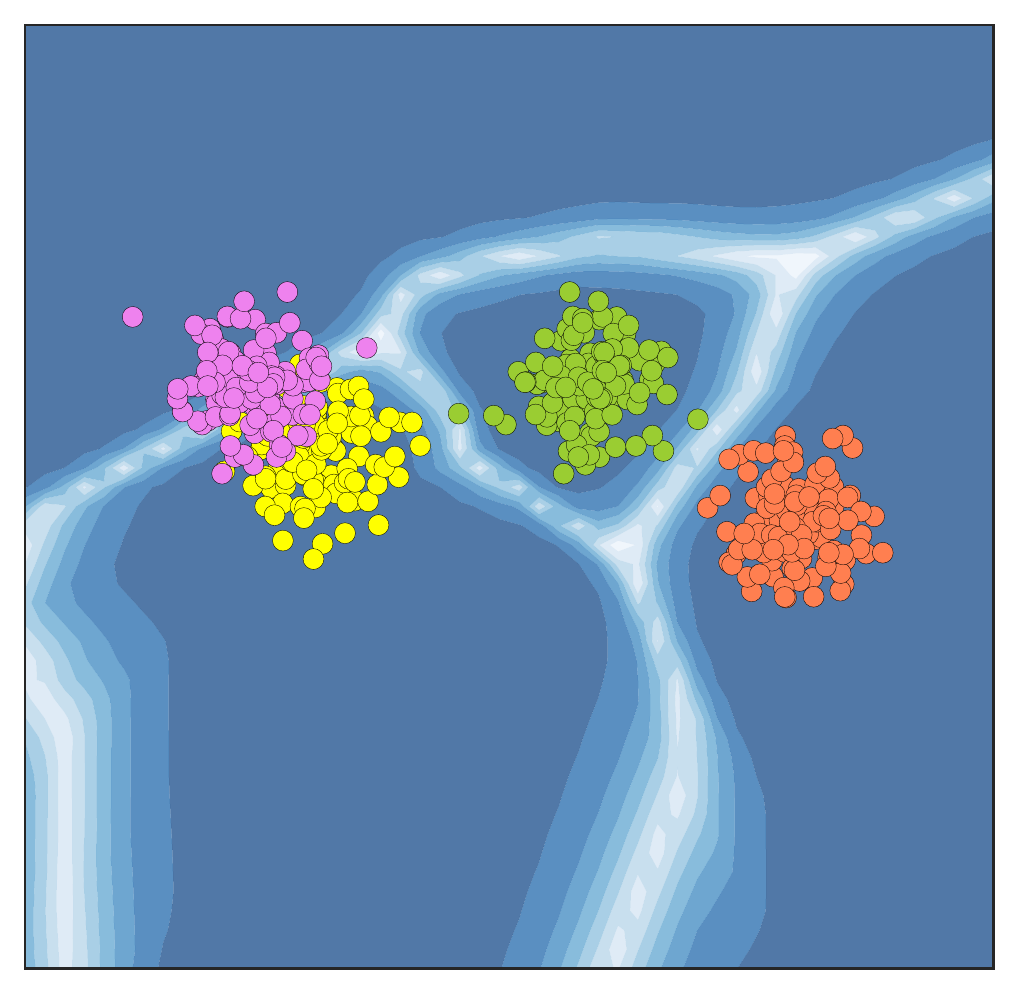}}
    \quad
    \subfloat[Temp. scaling]{\includegraphics[width=0.22\textwidth, height=0.15\textwidth]{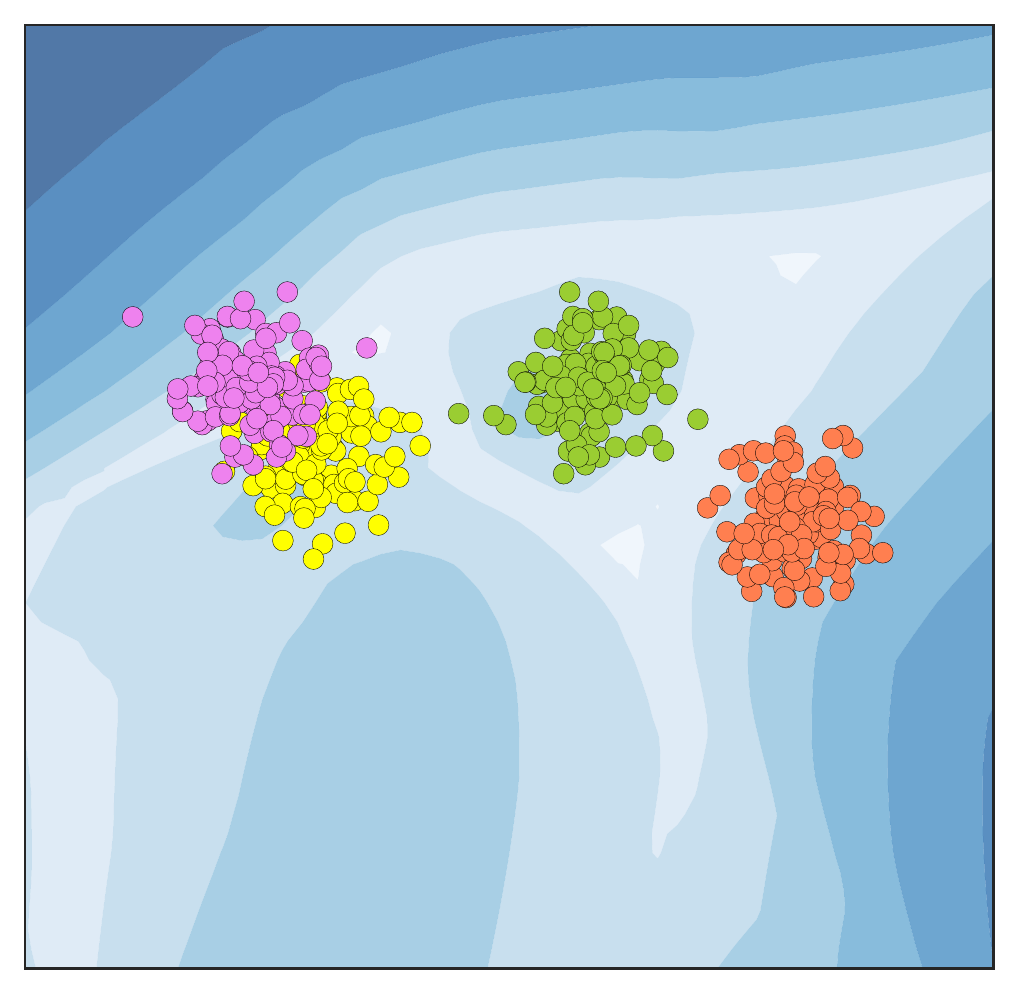}}
    \quad
    \subfloat[LLLA]{\includegraphics[width=0.22\textwidth, height=0.15\textwidth]{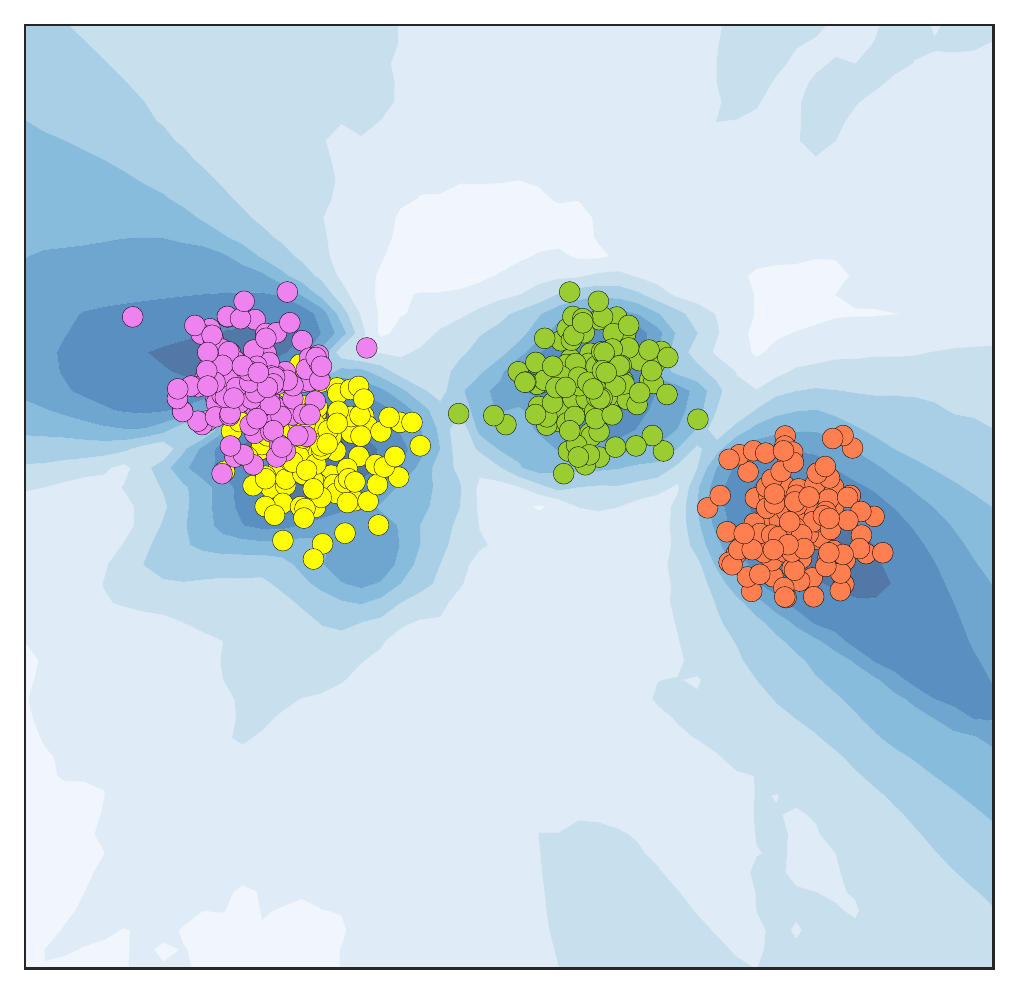}}
    \quad
    \subfloat[Full Laplace]{\includegraphics[width=0.22\textwidth, height=0.15\textwidth]{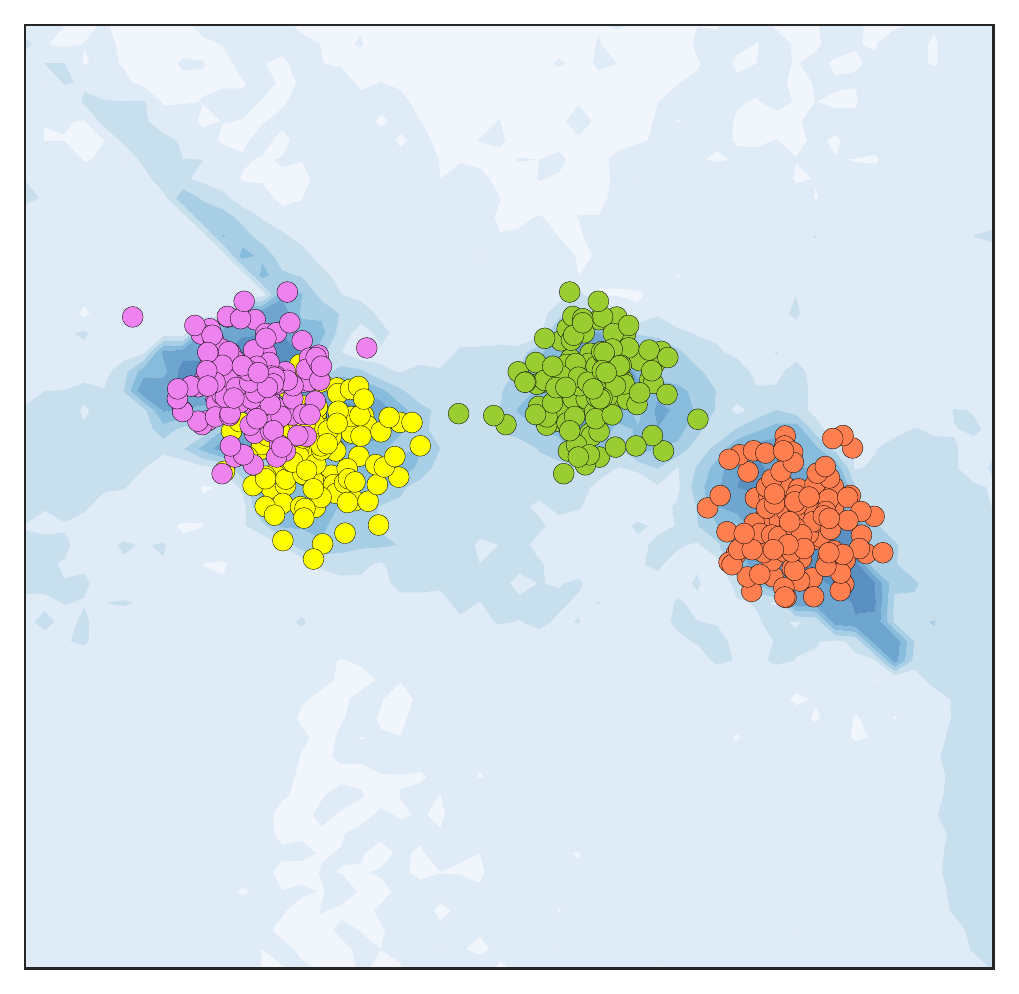}}
    \quad
    \subfloat{\includegraphics[width=0.03\textwidth, height=0.15\textwidth]{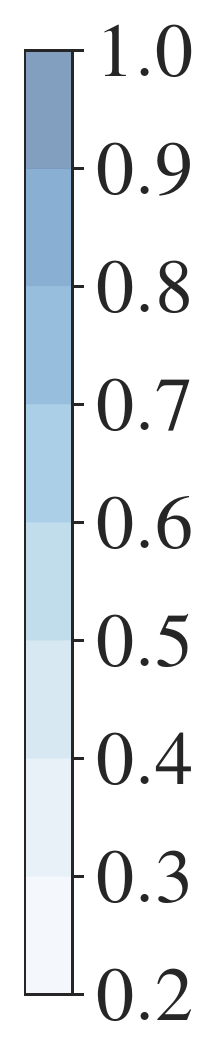}}

    \caption{Binary (top) and multi-class (bottom) toy classification problem. Background color represents confidence.}

    \label{fig:two_class_toy}
  \end{figure*}

The overconfidence problem of deep neural networks, and thus ReLU networks, has long been known in the deep learning community \citep{nguyen2015deep}, although a formal description was only delivered recently. Many methods have been proposed to combat or at least detect this issue. \emph{Post-hoc} heuristics based on temperature or Platt scaling \citep{platt1999probabilistic,guo17calibration,liang2018enhancing} are unable to detect inputs with arbitrarily high confidence far away from the training data \citep{hein2019relu}.

Many works on uncertainty quantification in deep learning have recently been proposed. \citet{gast2018lightweight} proposed lightweight probabilistic networks via assumed density filtering. \citet{malinin2018predictive,malinin2019reverse,sensoy2018evidential} employ a Dirichlet distribution to model the distribution of a network's output. \citet{lakshminarayanan2017simple} quantify predictive uncertainty based on the idea of model ensembling and frequentist calibration. \citet{hein2019relu} proposed enhanced training objectives based on robust optimization to mitigate this issue. \citet{meinke2020towards} proposed a similar approach with provable guarantees. However, they either lack in their theoretical analysis or do not employ probabilistic or Bayesian approximations. Our results, meanwhile, provide a theoretical justification to the commonly-used Gaussian approximations of NNs' weights, both Bayesian \citep[etc.]{graves_practical_2011,blundell_weight_2015,louizos_structured_2016,maddox2019simple} and non-Bayesian \citep[etc.]{franchi2019tradi,lu2020uncertainty}.

Bayesian methods have long been thought to mitigate the overconfidence problem on any neural network \citep{mackay1992evidence}. Empirical evidence supporting this intuition has also been presented \citep[etc.]{liu2018advbnn,wu2018deterministic}. Our results complement these with a theoretical justification for the ReLU-logistic case. Furthermore, our theoretical results show that, in some cases, an expensive Bayesian treatment over all layers of a network is not necessary (\Cref{thm:bounded_confidence_affine_relu}). Our results are thus theoretically validating the usage of Bayesian generalized linear models \citep{gelman2008weakly} (especially in conjunction with ReLU features) and last-layer Bayesian methods \citep{snoek2015scalable,wilson2016deep,wilson2016stochastic,riquelme2018deep}; and complementing the empirical analyses of \citet{ober2019benchmarking} and \citet{brosse2020last}.

\section{Experiments}
\label{sec:experiments}

\begin{figure*}[t!]
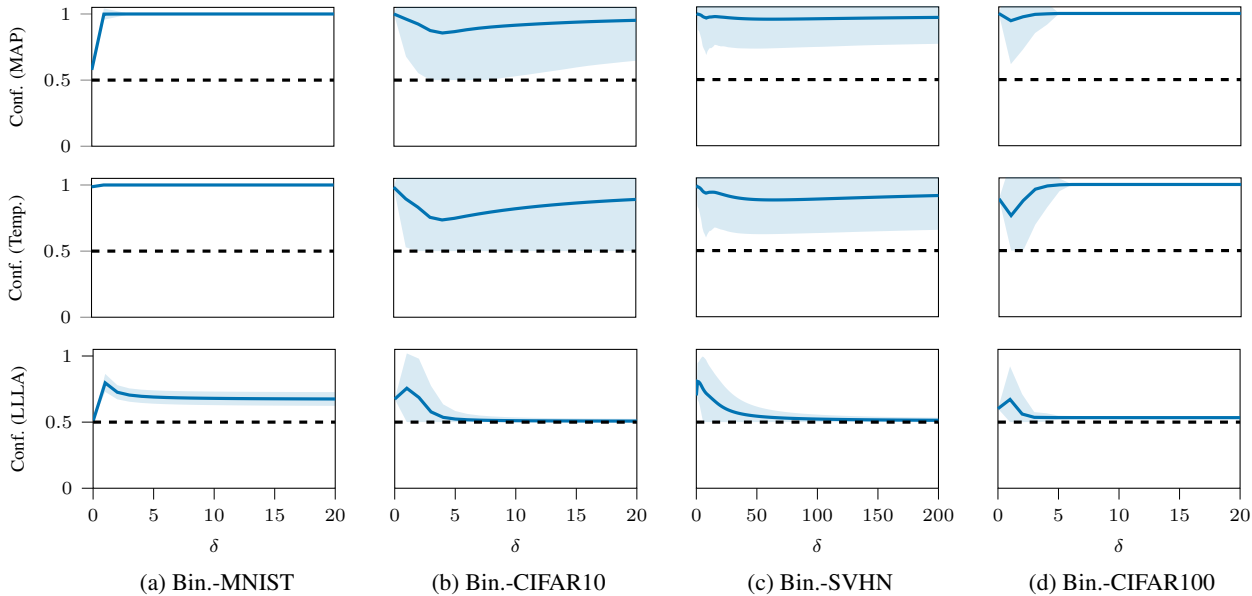

  \centering

  \hspace{3em}
  \subfloat{
\begin{tikzpicture}[baseline, trim axis left, trim axis right]

\tikzstyle{every node}=[font=\scriptsize]

\definecolor{color0}{rgb}{0.00392156862745098,0.450980392156863,0.698039215686274}

\begin{axis}[
width=0.28\textwidth,
height=0.2\textwidth,
tick align=outside,
tick pos=left,
x grid style={white!69.01960784313725!black},
xmin=0, xmax=20,
xtick style={color=black},
y grid style={white!69.01960784313725!black},
ylabel={Conf. (MAP)},
ymin=0, ymax=1.05,
xmajorticks=false,
]
\path [fill=color0, fill opacity=0.15, semithick]
(axis cs:0,0.577265381813049)
--(axis cs:0,0.577265381813049)
--(axis cs:1,0.956375002861023)
--(axis cs:2,0.975884258747101)
--(axis cs:3,0.995291948318481)
--(axis cs:4,0.999293982982635)
--(axis cs:5,0.999874114990234)
--(axis cs:6,0.999963700771332)
--(axis cs:7,0.999985754489899)
--(axis cs:8,0.999994158744812)
--(axis cs:9,0.999997615814209)
--(axis cs:10,0.999998986721039)
--(axis cs:11,0.999999582767487)
--(axis cs:12,0.999999821186066)
--(axis cs:13,0.999999940395355)
--(axis cs:14,1)
--(axis cs:15,1)
--(axis cs:16,1)
--(axis cs:17,1)
--(axis cs:18,1)
--(axis cs:19,1)
--(axis cs:20,1)
--(axis cs:20,1)
--(axis cs:20,1)
--(axis cs:19,1)
--(axis cs:18,1)
--(axis cs:17,1)
--(axis cs:16,1)
--(axis cs:15,1)
--(axis cs:14,1)
--(axis cs:13,1.00000011920929)
--(axis cs:12,1.00000011920929)
--(axis cs:11,1.00000035762787)
--(axis cs:10,1.00000095367432)
--(axis cs:9,1.00000238418579)
--(axis cs:8,1.0000057220459)
--(axis cs:7,1.00001394748688)
--(axis cs:6,1.000035405159)
--(axis cs:5,1.00012195110321)
--(axis cs:4,1.00068628787994)
--(axis cs:3,1.00459146499634)
--(axis cs:2,1.02352273464203)
--(axis cs:1,1.04164779186249)
--(axis cs:0,0.577265381813049)
--cycle;

\addplot [very thick, color0]
table {%
0 0.577265381813049
1 0.999011397361755
2 0.999703466892242
3 0.99994170665741
4 0.999990165233612
5 0.999998033046722
6 0.999999523162842
7 0.999999821186066
8 0.999999940395355
9 1
10 1
11 1
12 1
13 1
14 1
15 1
16 1
17 1
18 1
19 1
20 1
};
\addplot [very thick, black, dashed]
table {%
0 0.5
20 0.5
};
\end{axis}

\end{tikzpicture}}
  \qquad
  \subfloat{
\begin{tikzpicture}[baseline]

\tikzstyle{every node}=[font=\scriptsize]

\definecolor{color0}{rgb}{0.00392156862745098,0.450980392156863,0.698039215686274}

\begin{axis}[
width=0.28\textwidth,
height=0.2\textwidth,
tick align=outside,
tick pos=left,
x grid style={white!69.01960784313725!black},
xmin=0, xmax=20,
xtick style={color=black},
y grid style={white!69.01960784313725!black},
ymin=0, ymax=1.05,
xmajorticks=false, ymajorticks=false,
ytick style={color=black}
]
\path [fill=color0, fill opacity=0.15, semithick]
(axis cs:0,0.999915719032288)
--(axis cs:0,0.99991500377655)
--(axis cs:1,0.677247583866119)
--(axis cs:2,0.553297877311707)
--(axis cs:3,0.5)
--(axis cs:4,0.5)
--(axis cs:5,0.5)
--(axis cs:6,0.5)
--(axis cs:7,0.5)
--(axis cs:8,0.502380311489105)
--(axis cs:9,0.516180872917175)
--(axis cs:10,0.52839195728302)
--(axis cs:11,0.540452599525452)
--(axis cs:12,0.55387556552887)
--(axis cs:13,0.567001223564148)
--(axis cs:14,0.580338954925537)
--(axis cs:15,0.594748735427856)
--(axis cs:16,0.607540905475616)
--(axis cs:17,0.618395209312439)
--(axis cs:18,0.62794953584671)
--(axis cs:19,0.637705862522125)
--(axis cs:20,0.646351099014282)
--(axis cs:20,1.25800323486328)
--(axis cs:20,1.25800323486328)
--(axis cs:19,1.26202774047852)
--(axis cs:18,1.26665019989014)
--(axis cs:17,1.2707211971283)
--(axis cs:16,1.27545475959778)
--(axis cs:15,1.28131926059723)
--(axis cs:14,1.28792929649353)
--(axis cs:13,1.29294013977051)
--(axis cs:12,1.29696846008301)
--(axis cs:11,1.30021810531616)
--(axis cs:10,1.30127239227295)
--(axis cs:9,1.30097460746765)
--(axis cs:8,1.30017447471619)
--(axis cs:7,1.30028676986694)
--(axis cs:6,1.30348932743073)
--(axis cs:5,1.30456793308258)
--(axis cs:4,1.29805254936218)
--(axis cs:3,1.2839058637619)
--(axis cs:2,1.29458165168762)
--(axis cs:1,1.24536561965942)
--(axis cs:0,0.999915719032288)
--cycle;

\addplot [very thick, color0]
table {%
0 0.999915361404419
1 0.961306631565094
2 0.923939764499664
3 0.87463790178299
4 0.856374502182007
5 0.866474092006683
6 0.880843758583069
7 0.89234459400177
8 0.901277422904968
9 0.908577740192413
10 0.914832174777985
11 0.920335352420807
12 0.925422012805939
13 0.929970681667328
14 0.934134125709534
15 0.938033998012543
16 0.941497802734375
17 0.944558203220367
18 0.947299897670746
19 0.949866771697998
20 0.952177166938782
};
\addplot [very thick, black, dashed]
table {%
0 0.5
20 0.5
};
\end{axis}

\end{tikzpicture}}
  \qquad
  \subfloat{\input{figs/conf_map_svhn_binary}}
  \qquad
  \subfloat{
\begin{tikzpicture}

\tikzstyle{every node}=[font=\scriptsize]

\definecolor{color0}{rgb}{0.00392156862745098,0.450980392156863,0.698039215686274}

\begin{axis}[
width=0.28\textwidth,
height=0.2\textwidth,
tick align=outside,
tick pos=left,
x grid style={white!69.01960784313725!black},
xmin=0, xmax=20,
xtick style={color=black},
y grid style={white!69.01960784313725!black},
ymin=0, ymax=1.05,
ytick style={color=black},
xmajorticks=false,
ymajorticks=false
]
\path [fill=color0, fill opacity=0.15, semithick]
(axis cs:0,0.999985575675964)
--(axis cs:0,0.999984502792358)
--(axis cs:1,0.616365194320679)
--(axis cs:2,0.72725510597229)
--(axis cs:3,0.851654827594757)
--(axis cs:4,0.915990352630615)
--(axis cs:5,0.999987304210663)
--(axis cs:6,1)
--(axis cs:7,1)
--(axis cs:8,1)
--(axis cs:9,1)
--(axis cs:10,1)
--(axis cs:11,1)
--(axis cs:12,1)
--(axis cs:13,1)
--(axis cs:14,1)
--(axis cs:15,1)
--(axis cs:16,1)
--(axis cs:17,1)
--(axis cs:18,1)
--(axis cs:19,1)
--(axis cs:20,1)
--(axis cs:20,1)
--(axis cs:20,1)
--(axis cs:19,1)
--(axis cs:18,1)
--(axis cs:17,1)
--(axis cs:16,1)
--(axis cs:15,1)
--(axis cs:14,1)
--(axis cs:13,1)
--(axis cs:12,1)
--(axis cs:11,1)
--(axis cs:10,1)
--(axis cs:9,1)
--(axis cs:8,1)
--(axis cs:7,1)
--(axis cs:6,1)
--(axis cs:5,1.00001180171967)
--(axis cs:4,1.07854795455933)
--(axis cs:3,1.13463115692139)
--(axis cs:2,1.22226274013519)
--(axis cs:1,1.27414393424988)
--(axis cs:0,0.999985575675964)
--cycle;

\addplot [very thick, color0]
table {%
0 0.999985039234161
1 0.945254564285278
2 0.974758923053741
3 0.993143022060394
4 0.997269153594971
5 0.999999523162842
6 1
7 1
8 1
9 1
10 1
11 1
12 1
13 1
14 1
15 1
16 1
17 1
18 1
19 1
20 1
};
\addplot [very thick, black, dashed]
table {%
0 0.5
20 0.5
};
\end{axis}

\end{tikzpicture}}

  \vspace{-0.75em}
  \setcounter{subfigure}{0}

  \hspace{3em}
  \subfloat{
\begin{tikzpicture}[baseline, trim axis left, trim axis right]

\tikzstyle{every node}=[font=\scriptsize]

\definecolor{color0}{rgb}{0.00392156862745098,0.450980392156863,0.698039215686274}

\begin{axis}[
width=0.28\textwidth,
height=0.2\textwidth,
tick align=outside,
tick pos=left,
x grid style={white!69.01960784313725!black},
xmin=0, xmax=20,
xtick style={color=black},
y grid style={white!69.01960784313725!black},
ylabel={Conf. (Temp.)},
ymin=0, ymax=1.05,
xmajorticks=false,
]
\path [fill=color0, fill opacity=0.15, semithick]
(axis cs:0,0.986292123794556)
--(axis cs:0,0.986291766166687)
--(axis cs:1,0.999766886234283)
--(axis cs:2,1)
--(axis cs:3,1)
--(axis cs:4,1)
--(axis cs:5,1)
--(axis cs:6,1)
--(axis cs:7,1)
--(axis cs:8,1)
--(axis cs:9,1)
--(axis cs:10,1)
--(axis cs:11,1)
--(axis cs:12,1)
--(axis cs:13,1)
--(axis cs:14,1)
--(axis cs:15,1)
--(axis cs:16,1)
--(axis cs:17,1)
--(axis cs:18,1)
--(axis cs:19,1)
--(axis cs:20,1)
--(axis cs:20,1)
--(axis cs:20,1)
--(axis cs:19,1)
--(axis cs:18,1)
--(axis cs:17,1)
--(axis cs:16,1)
--(axis cs:15,1)
--(axis cs:14,1)
--(axis cs:13,1)
--(axis cs:12,1)
--(axis cs:11,1)
--(axis cs:10,1)
--(axis cs:9,1)
--(axis cs:8,1)
--(axis cs:7,1)
--(axis cs:6,1)
--(axis cs:5,1)
--(axis cs:4,1)
--(axis cs:3,1)
--(axis cs:2,1)
--(axis cs:1,1.00022804737091)
--(axis cs:0,0.986292123794556)
--cycle;

\addplot [very thick, color0]
table {%
0 0.986291944980621
1 0.999997496604919
2 1
3 1
4 1
5 1
6 1
7 1
8 1
9 1
10 1
11 1
12 1
13 1
14 1
15 1
16 1
17 1
18 1
19 1
20 1
};
\addplot [very thick, black, dashed]
table {%
0 0.5
20 0.5
};
\end{axis}

\end{tikzpicture}}
  \qquad
  \subfloat{
\begin{tikzpicture}[baseline]

\tikzstyle{every node}=[font=\fontsize{8}{9}\selectfont]

\definecolor{color0}{rgb}{0.00392156862745098,0.450980392156863,0.698039215686274}

\begin{axis}[
width=0.28\textwidth,
height=0.2\textwidth,
tick align=outside,
tick pos=left,
x grid style={white!69.01960784313725!black},
xmin=0, xmax=20,
xtick style={color=black},
y grid style={white!69.01960784313725!black},
ymin=0, ymax=1.05,
xmajorticks=false, ymajorticks=false,
ytick style={color=black}
]
\path [fill=color0, fill opacity=0.15, semithick]
(axis cs:0,0.983205616474152)
--(axis cs:0,0.983204543590546)
--(axis cs:1,0.529015839099884)
--(axis cs:2,0.5)
--(axis cs:3,0.5)
--(axis cs:4,0.5)
--(axis cs:5,0.5)
--(axis cs:6,0.5)
--(axis cs:7,0.5)
--(axis cs:8,0.5)
--(axis cs:9,0.5)
--(axis cs:10,0.5)
--(axis cs:11,0.5)
--(axis cs:12,0.5)
--(axis cs:13,0.5)
--(axis cs:14,0.5)
--(axis cs:15,0.5)
--(axis cs:16,0.5)
--(axis cs:17,0.5)
--(axis cs:18,0.5)
--(axis cs:19,0.5)
--(axis cs:20,0.5)
--(axis cs:20,1.30481195449829)
--(axis cs:20,1.30481195449829)
--(axis cs:19,1.30450809001923)
--(axis cs:18,1.30382347106934)
--(axis cs:17,1.3023202419281)
--(axis cs:16,1.30019521713257)
--(axis cs:15,1.29741787910461)
--(axis cs:14,1.29372978210449)
--(axis cs:13,1.28825187683105)
--(axis cs:12,1.28089702129364)
--(axis cs:11,1.27146470546722)
--(axis cs:10,1.25905978679657)
--(axis cs:9,1.24369812011719)
--(axis cs:8,1.22525823116302)
--(axis cs:7,1.20414352416992)
--(axis cs:6,1.18189096450806)
--(axis cs:5,1.15965914726257)
--(axis cs:4,1.14074242115021)
--(axis cs:3,1.16499757766724)
--(axis cs:2,1.26260197162628)
--(axis cs:1,1.26090788841248)
--(axis cs:0,0.983205616474152)
--cycle;

\addplot [very thick, color0]
table {%
0 0.983205080032349
1 0.894961893558502
2 0.832221984863281
3 0.755507230758667
4 0.735786139965057
5 0.747840285301208
6 0.764990746974945
7 0.780908048152924
8 0.795273542404175
9 0.808150172233582
10 0.819747924804688
11 0.830208837985992
12 0.839737176895142
13 0.848342955112457
14 0.856180787086487
15 0.86336475610733
16 0.869893252849579
17 0.875833690166473
18 0.881265640258789
19 0.886293649673462
20 0.890912234783173
};
\addplot [very thick, black, dashed]
table {%
0 0.5
20 0.5
};
\end{axis}

\end{tikzpicture}}
  \qquad
  \subfloat{\input{figs/conf_temp_svhn_binary}}
  \qquad
  \subfloat{
\begin{tikzpicture}

\tikzstyle{every node}=[font=\scriptsize]

\definecolor{color0}{rgb}{0.00392156862745098,0.450980392156863,0.698039215686274}

\begin{axis}[
width=0.28\textwidth,
height=0.2\textwidth,
tick align=outside,
tick pos=left,
x grid style={white!69.01960784313725!black},
xmin=0, xmax=20,
xtick style={color=black},
y grid style={white!69.01960784313725!black},
ymin=0, ymax=1.05,
ytick style={color=black},
xmajorticks=false,
ymajorticks=false
]
\path [fill=color0, fill opacity=0.15, semithick]
(axis cs:0,0.891654491424561)
--(axis cs:0,0.891653776168823)
--(axis cs:1,0.5)
--(axis cs:2,0.5)
--(axis cs:3,0.696532309055328)
--(axis cs:4,0.82928866147995)
--(axis cs:5,0.949530065059662)
--(axis cs:6,0.996741652488708)
--(axis cs:7,0.999860823154449)
--(axis cs:8,0.999993860721588)
--(axis cs:9,0.999999701976776)
--(axis cs:10,1)
--(axis cs:11,1)
--(axis cs:12,1)
--(axis cs:13,1)
--(axis cs:14,1)
--(axis cs:15,1)
--(axis cs:16,1)
--(axis cs:17,1)
--(axis cs:18,1)
--(axis cs:19,1)
--(axis cs:20,1)
--(axis cs:20,1)
--(axis cs:20,1)
--(axis cs:19,1)
--(axis cs:18,1)
--(axis cs:17,1)
--(axis cs:16,1)
--(axis cs:15,1)
--(axis cs:14,1)
--(axis cs:13,1)
--(axis cs:12,1)
--(axis cs:11,1)
--(axis cs:10,1)
--(axis cs:9,1.00000035762787)
--(axis cs:8,1.00000548362732)
--(axis cs:7,1.00012397766113)
--(axis cs:6,1.00291740894318)
--(axis cs:5,1.04503738880157)
--(axis cs:4,1.14853477478027)
--(axis cs:3,1.23214054107666)
--(axis cs:2,1.28547966480255)
--(axis cs:1,1.17830538749695)
--(axis cs:0,0.891654491424561)
--cycle;

\addplot [very thick, color0]
table {%
0 0.891654133796692
1 0.7650186419487
2 0.878090620040894
3 0.964336395263672
4 0.988911688327789
5 0.997283756732941
6 0.999829530715942
7 0.999992370605469
8 0.999999701976776
9 1
10 1
11 1
12 1
13 1
14 1
15 1
16 1
17 1
18 1
19 1
20 1
};
\addplot [very thick, black, dashed]
table {%
0 0.5
20 0.5
};
\end{axis}

\end{tikzpicture}}

  \vspace{-0.75em}
  \setcounter{subfigure}{0}

  \hspace{3em}
  \subfloat[Bin.-MNIST]{
\begin{tikzpicture}[trim axis left, trim axis right]

\tikzstyle{every node}=[font=\scriptsize]

\definecolor{color0}{rgb}{0.00392156862745098,0.450980392156863,0.698039215686274}

\begin{axis}[
width=0.28\textwidth,
height=0.2\textwidth,
tick align=outside,
tick pos=left,
x grid style={white!69.01960784313725!black},
xlabel={\(\displaystyle \delta\)},
xmin=0, xmax=20,
xtick style={color=black},
y grid style={white!69.01960784313725!black},
ylabel={Conf. (LLLA)},
ymin=0, ymax=1.05,
ytick style={color=black}
]
\path [fill=color0, fill opacity=0.15, semithick]
(axis cs:0,0.507924914360046)
--(axis cs:0,0.507924199104309)
--(axis cs:1,0.728665828704834)
--(axis cs:2,0.672055184841156)
--(axis cs:3,0.654027163982391)
--(axis cs:4,0.644757449626923)
--(axis cs:5,0.639215648174286)
--(axis cs:6,0.635541200637817)
--(axis cs:7,0.632920742034912)
--(axis cs:8,0.630966484546661)
--(axis cs:9,0.629452288150787)
--(axis cs:10,0.628244638442993)
--(axis cs:11,0.627259492874146)
--(axis cs:12,0.626440942287445)
--(axis cs:13,0.625749468803406)
--(axis cs:14,0.625158429145813)
--(axis cs:15,0.62464702129364)
--(axis cs:16,0.624199867248535)
--(axis cs:17,0.623806118965149)
--(axis cs:18,0.623456537723541)
--(axis cs:19,0.623144030570984)
--(axis cs:20,0.622863054275513)
--(axis cs:20,0.727196335792542)
--(axis cs:20,0.727196335792542)
--(axis cs:19,0.727399587631226)
--(axis cs:18,0.727626383304596)
--(axis cs:17,0.727881193161011)
--(axis cs:16,0.728169202804565)
--(axis cs:15,0.728497624397278)
--(axis cs:14,0.728875637054443)
--(axis cs:13,0.729315042495728)
--(axis cs:12,0.729831397533417)
--(axis cs:11,0.730448007583618)
--(axis cs:10,0.731197834014893)
--(axis cs:9,0.732129633426666)
--(axis cs:8,0.733319342136383)
--(axis cs:7,0.734891414642334)
--(axis cs:6,0.737063527107239)
--(axis cs:5,0.740275681018829)
--(axis cs:4,0.745451867580414)
--(axis cs:3,0.755156219005585)
--(axis cs:2,0.779523313045502)
--(axis cs:1,0.864915251731873)
--(axis cs:0,0.507924914360046)
--cycle;

\addplot [very thick, color0]
table {%
0 0.507924556732178
1 0.796790540218353
2 0.725789248943329
3 0.704591691493988
4 0.695104658603668
5 0.689745664596558
6 0.686302363872528
7 0.683906078338623
8 0.682142913341522
9 0.680790960788727
10 0.679721236228943
11 0.678853750228882
12 0.678136169910431
13 0.677532255649567
14 0.677017033100128
15 0.676572322845459
16 0.67618453502655
17 0.67584365606308
18 0.675541460514069
19 0.675271809101105
20 0.675029695034027
};
\addplot [very thick, black, dashed]
table {%
0 0.5
20 0.5
};
\end{axis}

\end{tikzpicture}}
  \qquad
  \subfloat[Bin.-CIFAR10]{
\begin{tikzpicture}[trim axis left, trim axis right]

\tikzstyle{every node}=[font=\scriptsize]

\definecolor{color0}{rgb}{0.00392156862745098,0.450980392156863,0.698039215686274}

\begin{axis}[
width=0.28\textwidth,
height=0.2\textwidth,
tick align=outside,
tick pos=left,
x grid style={white!69.01960784313725!black},
xlabel={\(\displaystyle \delta\)},
xmin=0, xmax=20,
xtick style={color=black},
y grid style={white!69.01960784313725!black},
ymajorticks=false,
ymin=0, ymax=1.05,
ytick style={color=black}
]
\path [fill=color0, fill opacity=0.15, semithick]
(axis cs:0,0.674011290073395)
--(axis cs:0,0.674011290073395)
--(axis cs:1,0.5)
--(axis cs:2,0.5)
--(axis cs:3,0.5)
--(axis cs:4,0.5)
--(axis cs:5,0.5)
--(axis cs:6,0.5)
--(axis cs:7,0.5)
--(axis cs:8,0.5)
--(axis cs:9,0.5)
--(axis cs:10,0.5)
--(axis cs:11,0.5)
--(axis cs:12,0.5)
--(axis cs:13,0.5)
--(axis cs:14,0.5)
--(axis cs:15,0.5)
--(axis cs:16,0.5)
--(axis cs:17,0.5)
--(axis cs:18,0.5)
--(axis cs:19,0.5)
--(axis cs:20,0.5)
--(axis cs:20,0.525096237659454)
--(axis cs:20,0.525096237659454)
--(axis cs:19,0.525542497634888)
--(axis cs:18,0.526053369045258)
--(axis cs:17,0.526642978191376)
--(axis cs:16,0.527329742908478)
--(axis cs:15,0.528143703937531)
--(axis cs:14,0.529123485088348)
--(axis cs:13,0.530305027961731)
--(axis cs:12,0.531758248806)
--(axis cs:11,0.533593118190765)
--(axis cs:10,0.535964071750641)
--(axis cs:9,0.539149343967438)
--(axis cs:8,0.543635189533234)
--(axis cs:7,0.550427436828613)
--(axis cs:6,0.561957478523254)
--(axis cs:5,0.584208130836487)
--(axis cs:4,0.635019183158875)
--(axis cs:3,0.774278044700623)
--(axis cs:2,0.979620814323425)
--(axis cs:1,1.0198540687561)
--(axis cs:0,0.674011290073395)
--cycle;

\addplot [very thick, color0]
table {%
0 0.674011290073395
1 0.756257712841034
2 0.686343789100647
3 0.577509999275208
4 0.536078095436096
5 0.523309290409088
6 0.517828464508057
7 0.514831185340881
8 0.512975215911865
9 0.511726140975952
10 0.510834038257599
11 0.510167837142944
12 0.509653747081757
13 0.509243905544281
14 0.508910834789276
15 0.508634924888611
16 0.508402764797211
17 0.508204340934753
18 0.50803279876709
19 0.507883787155151
20 0.507752418518066
};
\addplot [very thick, black, dashed]
table {%
0 0.5
20 0.5
};
\end{axis}

\end{tikzpicture}}
  \qquad
  \subfloat[Bin.-SVHN]{\input{figs/conf_laplace_svhn_binary}}
  \qquad
  \subfloat[Bin.-CIFAR100]{
\begin{tikzpicture}[trim axis left, trim axis right]

\tikzstyle{every node}=[font=\scriptsize]

\definecolor{color0}{rgb}{0.00392156862745098,0.450980392156863,0.698039215686274}

\begin{axis}[
width=0.28\textwidth,
height=0.2\textwidth,
tick align=outside,
tick pos=left,
x grid style={white!69.01960784313725!black},
xlabel={\(\displaystyle \delta\)},
xmin=0, xmax=20,
xtick style={color=black},
y grid style={white!69.01960784313725!black},
ymin=0, ymax=1.05,
ytick style={color=black},
ymajorticks=false
]
\path [fill=color0, fill opacity=0.15, semithick]
(axis cs:0,0.60160756111145)
--(axis cs:0,0.601607203483582)
--(axis cs:1,0.5)
--(axis cs:2,0.5)
--(axis cs:3,0.5)
--(axis cs:4,0.503103733062744)
--(axis cs:5,0.518903136253357)
--(axis cs:6,0.521266341209412)
--(axis cs:7,0.52255117893219)
--(axis cs:8,0.523337602615356)
--(axis cs:9,0.523868143558502)
--(axis cs:10,0.524250745773315)
--(axis cs:11,0.524539947509766)
--(axis cs:12,0.524770021438599)
--(axis cs:13,0.524956881999969)
--(axis cs:14,0.525113940238953)
--(axis cs:15,0.52524721622467)
--(axis cs:16,0.525361239910126)
--(axis cs:17,0.525458872318268)
--(axis cs:18,0.525544047355652)
--(axis cs:19,0.525619328022003)
--(axis cs:20,0.525685608386993)
--(axis cs:20,0.542167365550995)
--(axis cs:20,0.542167365550995)
--(axis cs:19,0.542246520519257)
--(axis cs:18,0.542336106300354)
--(axis cs:17,0.542437374591827)
--(axis cs:16,0.542553126811981)
--(axis cs:15,0.542688012123108)
--(axis cs:14,0.542845487594604)
--(axis cs:13,0.543030560016632)
--(axis cs:12,0.543250560760498)
--(axis cs:11,0.543519854545593)
--(axis cs:10,0.543855786323547)
--(axis cs:9,0.544295132160187)
--(axis cs:8,0.544890880584717)
--(axis cs:7,0.545756220817566)
--(axis cs:6,0.547112822532654)
--(axis cs:5,0.549527287483215)
--(axis cs:4,0.567056894302368)
--(axis cs:3,0.575985729694366)
--(axis cs:2,0.711506605148315)
--(axis cs:1,0.922095239162445)
--(axis cs:0,0.60160756111145)
--cycle;

\addplot [very thick, color0]
table {%
0 0.601607382297516
1 0.671642124652863
2 0.561062335968018
3 0.536244332790375
4 0.535080313682556
5 0.534215211868286
6 0.534189581871033
7 0.534153699874878
8 0.534114241600037
9 0.534081637859344
10 0.534053266048431
11 0.534029901027679
12 0.534010291099548
13 0.533993721008301
14 0.533979713916779
15 0.533967614173889
16 0.533957183361053
17 0.533948123455048
18 0.533940076828003
19 0.53393292427063
20 0.533926486968994
};
\addplot [very thick, black, dashed]
table {%
0 0.5
20 0.5
};
\end{axis}

\end{tikzpicture}}

  \caption{Confidence of MAP (top row), temperature scaling (middle row), and LLLA (bottom row) as functions of $\delta$ over the test sets of binary classification datasets. Thick blue lines and shades correspond to means and $\pm 3$ standard deviations, respectively. Dashed lines signify the desirable confidence for $\delta$ sufficiently high.}
  \label{fig:exp_far_away}
\end{figure*}

\begin{table*}[t]
  \caption{OOD detection for far-away points in binary classification settings. The in-distribution datasets are Binary-MNIST, Binary-CIFAR10, Binary-SVHN, and Binary-CIFAR100. Each OOD dataset is obtained by scaling uniform noise images in the corresponding input space of the in-distribution dataset with $\delta = 100$. All values are means and standard deviations over 10 trials.}
  \label{tab:ood_faraway}

  \vspace{1em}

  \centering
  \scriptsize

  \begin{tabular}{lrrrr|rr}
    \toprule

      & \multicolumn{2}{c}{\bf MAP} & \multicolumn{2}{c|}{\bf +Temp.} & \multicolumn{2}{c}{\bf +LLLA} \\
    \cmidrule(r){2-3} \cmidrule(r){4-5} \cmidrule(l){6-7}
      & MMC $\downarrow$ & AUR $\uparrow$ & MMC $\downarrow$ & AUR $\uparrow$ & MMC $\downarrow$ & AUR $\uparrow$ \\

    \midrule

    Binary-MNIST    & 99.9$\pm$0.0 & - & 100.0$\pm$0.0 & - & 79.4$\pm$0.9 & - \\
    Noise ($\delta = 100$)         & 100.0$\pm$0.0 & 0.2$\pm$0.1 & 100.0$\pm$0.0 & 45.1$\pm$5.8 & \textbf{67.5$\pm$0.8} & \textbf{99.6$\pm$0.1} \\

    \midrule

    Binary-CIFAR10   & 96.3$\pm$0.3 & - & 90.5$\pm$0.6 & - & 76.4$\pm$0.3 & - \\
    Noise ($\delta = 100$)        & 98.9$\pm$1.0 & 11.3$\pm$10.3 & 97.6$\pm$2.2 & 11.3$\pm$10.3 & \textbf{50.6$\pm$0.1} & \textbf{99.5$\pm$0.1} \\

    \midrule

    Binary-SVHN   & 99.4$\pm$0.0 & - & 98.2$\pm$0.1 & - & 80.7$\pm$0.1 & - \\
    Noise ($\delta = 100$)       & 98.8$\pm$0.6 & 50.5$\pm$42.3 & 95.9$\pm$3.0 & 50.5$\pm$42.3 & \textbf{51.2$\pm$0.6} & \textbf{99.8$\pm$0.1} \\

    \midrule

    Binary-CIFAR100   & 94.5$\pm$0.5 & - & 74.5$\pm$2.9 & - & 66.7$\pm$0.5 & - \\
    Noise ($\delta = 100$)           & 100.0$\pm$0.0 & 1.5$\pm$0.7 & 100.0$\pm$0.0 & 0.0$\pm$0.0 & \textbf{53.5$\pm$0.1} & \textbf{93.6$\pm$1.8} \\

    \bottomrule

  \end{tabular}
\end{table*}

\begin{table*}[t]
  \caption{Multi-class OOD detection results for MAP, last-layer Laplace (LLLA), (all-layers) diagonal Laplace (DLA), and (all-layers) Kronecker-Factored Laplace (KFLA). Each ``far-away'' Noise dataset is constructed as in \Cref{tab:ood_faraway} with $\delta = 2000$. All values are averages and standard deviations over 10 trials.}
  \label{tab:ood_results}

  \vspace{1em}

  \centering
  \scriptsize
  \renewcommand{\tabcolsep}{5pt}

  \begin{tabular}{lrrrr|rrrrrr}
    \toprule

      & \multicolumn{2}{c}{\bf MAP} & \multicolumn{2}{c|}{\bf +Temp.} & \multicolumn{2}{c}{\bf +LLLA} & \multicolumn{2}{c}{\bf +DLA} & \multicolumn{2}{c}{\bf +KFLA} \\
    \cmidrule(r){2-3} \cmidrule(r){4-5} \cmidrule(l){6-7} \cmidrule(l){8-9} \cmidrule(l){10-11}
      & MMC $\downarrow$ & AUR $\uparrow$ & MMC $\downarrow$ & AUR $\uparrow$ & MMC $\downarrow$ & AUR $\uparrow$ & MMC $\downarrow$ & AUR $\uparrow$ & MMC $\downarrow$ & AUR $\uparrow$ \\

    \midrule

    MNIST - MNIST & 99.2$\pm$0.0 & - & 99.5$\pm$0.1 & - & 98.4$\pm$0.2 & - & 84.5$\pm$0.2 & - & 92.9$\pm$0.3 & - \\
    MNIST - EMNIST & 82.3$\pm$0.0 & 89.2$\pm$0.1 & 87.6$\pm$1.4 & 88.9$\pm$0.2 & 70.2$\pm$1.9 & \textbf{92.0$\pm$0.4} & \textbf{54.5$\pm$0.3} & 87.7$\pm$0.4 & 58.7$\pm$0.4 & 89.6$\pm$0.3 \\
    MNIST - FMNIST & 66.3$\pm$0.0 & 97.4$\pm$0.0 & 75.2$\pm$2.5 & 97.1$\pm$0.1 & 56.0$\pm$1.8 & 98.2$\pm$0.2 & 42.5$\pm$0.1 & 96.3$\pm$0.1 & \textbf{39.9$\pm$0.5} & \textbf{98.6$\pm$0.1} \\
    MNIST - Noise ($\delta = 2000$) & 100.0$\pm$0.0 & 0.1$\pm$0.0 & 100.0$\pm$0.0 & 6.8$\pm$4.1 & 99.9$\pm$0.0 & 9.6$\pm$0.7 & 84.9$\pm$1.3 & 53.7$\pm$3.1 & \textbf{55.6$\pm$2.0} & \textbf{97.3$\pm$0.4} \\

    \midrule

    CIFAR10 - CIFAR10 & 97.1$\pm$0.1 & - & 95.4$\pm$0.2 & - & 92.8$\pm$1.1 & - & 88.4$\pm$0.1 & - & 86.5$\pm$0.1 & - \\
    CIFAR10 - SVHN & 62.5$\pm$0.0 & 95.8$\pm$0.1 & 54.6$\pm$0.6 & 96.1$\pm$0.0 & 45.9$\pm$1.6 & \textbf{96.4$\pm$0.1} & 43.3$\pm$0.1 & 95.5$\pm$0.1 & \textbf{43.0$\pm$0.1} & 94.8$\pm$0.1 \\
    CIFAR10 - LSUN & 74.5$\pm$0.0 & 91.9$\pm$0.1 & 66.9$\pm$0.6 & 92.2$\pm$0.1 & 57.4$\pm$1.9 & 92.7$\pm$0.4 & 49.0$\pm$0.5 & \textbf{92.8$\pm$0.3} & \textbf{47.6$\pm$0.4} & 92.2$\pm$0.2 \\
    CIFAR10 - Noise ($\delta = 2000$) & 98.7$\pm$0.2 & 10.9$\pm$0.4 & 98.4$\pm$0.2 & 10.0$\pm$0.5 & \textbf{17.4$\pm$0.0} & \textbf{100.0$\pm$0.0} & 60.7$\pm$2.0 & 89.6$\pm$1.1 & 61.8$\pm$1.5 & 87.6$\pm$0.9 \\

    \midrule

    SVHN - SVHN & 98.5$\pm$0.0 & - & 97.4$\pm$0.2 & - & 93.2$\pm$1.0 & - & 88.8$\pm$0.0 & - & 90.8$\pm$0.0 & - \\
    SVHN - CIFAR10 & 70.4$\pm$0.0 & 95.4$\pm$0.0 & 64.1$\pm$0.9 & 95.4$\pm$0.0 & 43.4$\pm$2.1 & 97.2$\pm$0.1 & \textbf{38.0$\pm$0.1} & \textbf{97.6$\pm$0.0} & 41.2$\pm$0.1 & 97.5$\pm$0.0 \\
    SVHN - LSUN & 71.7$\pm$0.0 & 95.5$\pm$0.0 & 65.4$\pm$1.0 & 95.6$\pm$0.0 & 44.3$\pm$2.3 & 97.3$\pm$0.1 & \textbf{39.5$\pm$0.7} & \textbf{97.5$\pm$0.2} & 42.0$\pm$0.6 & \textbf{97.5$\pm$0.1} \\
    SVHN - Noise ($\delta = 2000$) & 98.7$\pm$0.1 & 11.9$\pm$0.6 & 98.4$\pm$0.1 & 11.0$\pm$0.6 & \textbf{27.5$\pm$0.1} & \textbf{99.6$\pm$0.0} & 60.8$\pm$1.6 & 92.8$\pm$0.6 & 62.4$\pm$2.0 & 94.0$\pm$0.5 \\

    \midrule

    CIFAR100 - CIFAR100 & 81.2$\pm$0.1 & - & 78.9$\pm$0.8 & - & 74.6$\pm$0.2 & - & 76.4$\pm$0.2 & - & 73.4$\pm$0.2 & - \\
    CIFAR100 - SVHN & 53.5$\pm$0.0 & 78.8$\pm$0.1 & 49.2$\pm$1.2 & 79.2$\pm$0.1 & 42.7$\pm$0.3 & \textbf{80.4$\pm$0.2} & 46.0$\pm$0.1 & 79.6$\pm$0.2 & \textbf{41.4$\pm$0.1} & 80.1$\pm$0.2 \\
    CIFAR100 - LSUN & 50.7$\pm$0.0 & 81.0$\pm$0.1 & 46.8$\pm$1.1 & 81.1$\pm$0.1 & 39.8$\pm$0.2 & \textbf{82.6$\pm$0.2} & 43.5$\pm$0.3 & 81.5$\pm$0.2 & \textbf{39.7$\pm$0.4} & 81.6$\pm$0.3 \\
    CIFAR100 - Noise ($\delta = 2000$) & 99.5$\pm$0.1 & 2.8$\pm$0.2 & 99.4$\pm$0.1 & 2.6$\pm$0.2 & \textbf{5.9$\pm$0.0} & \textbf{99.9$\pm$0.0} & 41.5$\pm$1.5 & 84.2$\pm$0.9 & 37.1$\pm$1.3 & 84.2$\pm$0.8 \\

    \bottomrule
  \end{tabular}
\end{table*}

We corroborate our theoretical results via four experiments using various Gaussian-based Bayesian methods. In \Cref{subsec:toy_exp} we visualize the confidence of 2D binary and multi-class toy datasets. In \Cref{subsec:binary_exp} we empirically validate our main result that the confidence of binary classification datasets approaches finite constants as $\delta$ increases. Furthermore, we show empirically that this property also holds in the multi-class case, along with the usefulness of Bayesian methods in standard OOD detection tasks in \Cref{subsec:ood_exp}. Finally, in \Cref{subsec:gps}, we show that our results also hold in the case of last-layer Gaussian processes.

Unless stated otherwise, we use LeNet (for MNIST) or ResNet-18 (for CIFAR-10, SVHN, CIFAR-100) architectures. We train these networks by following the procedure described by \citet{meinke2020towards} (\Cref{appendix:training}). To obtain the optimal hyperparameter $\sigma_0^2$, we follow \eqref{eq:opt_var_prior} with $\lambda$ set to 0.25. We mainly use a \textit{last-layer Laplace approximation} (LLLA)\footnote{\fontsize{8pt}{8pt}\selectfont \url{https://github.com/wiseodd/last_layer_laplace}.} where a Laplace approximation with an exact Hessian or its Kronecker factors is applied only to the last layer of a network (\Cref{appendix:applications}). Whenever the approximations of predictive distribution in \eqref{eq:z} and \eqref{eq:z_full} cannot be used, we compute them via Monte Carlo integrations with 100 posterior samples. Other Laplace approximations that we use will be introduced in the subsection where they are first employed. Besides the vanilla MAP method, we use the temperature scaling method \citep{guo17calibration} as a baseline since it claims to give calibrated predictions in the frequentist sense. In particular, the optimal temperature is found via a validation log-likelihood maximization using PyCalib \citep{wenger2019nonparametric}. For each dataset that we use, we obtain a validation set via a random split from the respective test set.\footnote{We use 50, 1000, and 2000 points for the toy, binary, and multi-class classification cases, respectively.}  Lastly, all numbers reported in this section are averages along with their standard deviations over 10 trials.

\subsection{Toy Dataset}
\label{subsec:toy_exp}

Here, the dataset is constructed by sampling the input points from $k$ independent Gaussians. The corresponding targets indicate from which Gaussian the point was sampled.  We use a 3-layer ReLU network with 20 hidden units at each layer. We use the exact Hessian and the full generalized-Gauss-Newton (GGN) approximation of the Hessian for the case of LLLA and all-layer Laplace approximations, respectively.

We show the results for the binary and multi-class cases in \Cref{fig:two_class_toy}. The MAP predictions have high confidence everywhere except at the region close to the decision boundary. Temperature scaling assigns low confidence to the training data, while assigning high confidence far away from them. LLLA, albeit simple, yields high confidence close to the training points and high uncertainty otherwise, while maintaining the MAP's decision boundary. Furthermore, we found that the all-layer Laplace approximation makes the aforementioned finding stronger: the boundaries of the high-confidence regions are now closer to the training data.

\subsection{Binary Classification}
\label{subsec:binary_exp}

We validate our theoretical finding by plotting the test confidence of various binary classification datasets as functions of $\delta$. Each dataset is constructed by picking two classes which are most difficult to distinguish, based on the confusion matrix of the corresponding multi-class problem.

As shown in \Cref{fig:exp_far_away}, both MAP (top row) and temperature scaling (middle row) methods are overconfident for sufficiently large $\delta$. Meanwhile, LLLA which represents Bayesian methods, mitigates this issue: As $\delta$ increases, the confidence converges to some constant close to the uniform confidence (one-half). Moreover, when $\delta = 1$ (the case of in-distribution data), LLLA retains higher confidence.

\Cref{tab:ood_faraway} further quantifies the results where we treat collections of 2000 uniform noise images scaled by $\delta = 100$ as the OOD datasets. Note that, while the resulting data points are not in the image space anymore, this construction is useful to assess the effectiveness of the Bayesian methods in unbounded problems. We report the standard metrics proposed by \citet{hendrycks17baseline}: mean-maximum-confidence (MMC) and area-under-ROC-curve (AUR). Confirming our finding in \Cref{fig:exp_far_away}, LLLA is able to detect OOD data with high accuracy: for the chosen values of $\delta$, the MMC and AUR values are close to the ideal values of 50 and 100, respectively. Both MAP and temperature scaling fail to do so since their confidence estimates saturate to one. These results (i) confirm our theoretical analysis in \Cref{sec:theory}, (ii) show that even a simple Bayesian method yields good uncertainty estimates, and (iii) temperature scaling is not calibrated for outliers far-away from the training data.\footnote{This confirms the theoretical arguments of \citet{hein2019relu}.}

\subsection{Multi-class Classification}
\label{subsec:ood_exp}

We also show empirically that Bayesian methods yield a similar behavior in multi-class settings. On top of LLLA, representing Bayesian methods, we employ various  other scalable Laplace approximation techniques: diagonal Laplace approximation (DLA) where a diagonal Gaussian is used to approximate the posterior over all layers of a network, and Kronecker-factored Laplace approximation (KFLA) \citep{ritter_scalable_2018} where a matrix-variate normal is used to approximate the posterior over all layers. We use 20 posterior samples for both DLA and KFLA. We refer the reader to \Cref{appendix:applications} for details.

For each training dataset we evaluate all methods both in the non-asymptotic (the corresponding OOD test datasets, e.g. SVHN and LSUN for CIFAR-10) and asymptotic (Noise datasets) regime. Each ``far-away'' Noise dataset is constructed by scaling 2000 uniform noise images in the corresponding input space with $\delta = 2000$. As in the previous section, we report the MMC and AUR metrics.

As presented in \Cref{tab:ood_results}, all the Bayesian methods improve the OOD detection performance of the base models both in the non-asymptotic and asymptotic regime. Especially in the asymptotic regime, all the Bayesian methods perform well, empirically confirming our hypothesis that our theoretical analysis carries over to the multi-class setting. Meanwhile, both MAP's and temperature scaling's MMC and AUR are close to 100 and 0, respectively.\footnote{I.e. the worst values for those metrics.} Moreover, while LLLA is the simplest Bayesian method in this experiment, it often outperforms DLA and KFLA. Our finding agrees with the prior observation that last-layer Bayesian approximations are often sufficient \citep{ober2019benchmarking,brosse2020last}. Furthermore, in \Cref{appendix:more_experiments} we also show that we can apply Laplace approximations on top of prior OOD detection methods such as ACET \citep{hein2019relu} and Outlier-Exposure \citep{hendrycks2018deep} to achieve state-of-the-art performance in the non-asymptotic regime, while also having high uncertainty in the asymptotic regime.

\begin{table}[h]
  \caption{Wall-clock time (in second) of posterior inferences and predictions over test sets.}
  \label{tab:timing}

  \vspace{1em}

  \centering
  \scriptsize

  \begin{tabular}{lrrrr}

    \toprule

       & {\bf MNIST} & {\bf CIFAR-10} & {\bf SVHN} & {\bf CIFAR-100} \\

      \midrule

      {\bf Inference} \\

      MAP    & -     & -     & -     & -     \\
      +Temp. & 0.0   & 0.0   & 0.0   & 0.0   \\
      +LLLA  & 1.8   & 23.0  & 33.7  & 23.1  \\
      +DLA   & 1.3   & 22.9  & 33.6  & 23.0  \\
      +KFLA  & 4.3   & 78.1  & 115.1 & 78.6  \\

      \midrule

      {\bf Prediction} \\

      MAP    & 0.4   & 1.2   & 2.7   & 1.2   \\
      +Temp. & 0.4   & 1.2   & 2.7   & 1.2   \\
      +LLLA  & 0.9   & 1.7   & 4.3   & 1.6   \\
      +DLA   & 7.3   & 100.0 & 260.6 & 100.4 \\
      +KFLA  & 21.5  & 151.0 & 392.8 & 151.7 \\

      \bottomrule
  \end{tabular}
\end{table}

In \Cref{tab:timing}, we present the computational cost analysis in terms of wall-clock time. We measure the time required for each method to do posterior inference (or finding the optimal temperature) and to make predictions. While MAP and temperature scaling are fast, as we have shown in the previous results, they are overconfident. Among the Bayesian methods, since the cost of LLLA is constant w.r.t. the network depth, we found that it is up to two orders of magnitude faster than DLA and KFLA when making predictions. All in all, this finding, combined with the previous results, makes this simple Bayesian method attractive in applications.

\begin{table}[t]
  \caption{Multi-class OOD detection results for deep kernel learning (DKL). Each ``far-away'' Noise dataset is constructed as in \Cref{tab:ood_results} with $\delta = 2000$. All values are averages and standard deviations over 10 trials.}
  \label{tab:dkl}

  \vspace{1em}

  \centering
  \scriptsize
  \renewcommand{\tabcolsep}{10pt}

  \begin{tabular}{lrr}
    \toprule

    \textbf{Train - Test}  & \textbf{MMC $\downarrow$} & \textbf{AUR $\uparrow$} \\

    \midrule

    MNIST - MNIST & 99.6$\pm$0.0 & - \\
    MNIST - EMNIST & 83.9$\pm$0.0 & 94.4$\pm$0.1 \\
    MNIST - FMNIST & 70.6$\pm$0.1 & 98.8$\pm$0.0 \\
    MNIST - Noise & 58.6$\pm$0.5 & 99.7$\pm$0.0 \\

    \midrule

    CIFAR10 - CIFAR10 & 97.5$\pm$0.0 & - \\
    CIFAR10 - SVHN & 50.6$\pm$0.1 & 98.6$\pm$0.0 \\
    CIFAR10 - LSUN & 77.9$\pm$0.3 & 93.4$\pm$0.1 \\
    CIFAR10 - Noise & 56.5$\pm$0.7 & 98.5$\pm$0.1 \\

    \midrule

    SVHN - SVHN & 98.6$\pm$0.0 & - \\
    SVHN - CIFAR10 & 72.7$\pm$0.0 & 96.0$\pm$0.0 \\
    SVHN - LSUN & 76.7$\pm$0.1 & 95.1$\pm$0.1 \\
    SVHN - Noise & 48.6$\pm$0.7 & 99.4$\pm$0.0 \\

    \midrule

    CIFAR100 - CIFAR100 & 80.5$\pm$0.0 & - \\
    CIFAR100 - SVHN & 72.7$\pm$0.1 & 63.1$\pm$0.1 \\
    CIFAR100 - LSUN & 66.8$\pm$0.4 & 69.7$\pm$0.3 \\
    CIFAR100 - Noise & 43.1$\pm$1.1 & 90.3$\pm$0.7 \\

    \bottomrule
  \end{tabular}

\end{table}

\subsection{Last-layer Gaussian Processes}
\label{subsec:gps}

It has been shown that Gaussian-approximated linear models with infinitely many features, e.g. two-layer networks with infinitely wide hidden layers, are equivalent to Gaussian processes \citep{neal1996priors}. In the language of \Cref{thm:bounded_confidence_affine_relu}, this is the case when the dimension of the feature space $\R^d$ goes to infinity. It is therefore interesting to see, at least empirically, whether low asymptotic confidence is also attained in this case.

While unlike LLLA, deep kernel learning \citep[DKL,][]{wilson2016deep,wilson2016stochastic} is not a \emph{post-hoc} method, it is a suitable model for showcasing our theory in the case of last-layer Gaussian processes. We therefore train stochastic variational DKL models \citep{wilson2016stochastic} which use the same networks used in the previous experiment (minus the top layer) as their feature extractors, following the implementation provided by GPyTorch \cite{gardner2018gpytorch}. The training protocol is identical as before (cf. \Cref{appendix:training}). To compute each prediction, we use 20 samples from the Gaussian process posterior. We are mainly interested in the performance of DKL in term of multi-class OOD detection (both in asymptotic and non-asymptotic regimes), similar to the previous section.

The results are presented in \Cref{tab:dkl}. When compared to the results of the MAP estimation in \Cref{tab:ood_results}, we found that DKL is able to mitigate asymptotic overconfidence (see results against the Noise dataset). These results empirically verify that our analysis also holds in the non-parametric infinite-width regime. Nevertheless, we found that LLLA generally outperforms DKL both in term of MMC and AUR metrics. This finding, along with the simplicity and efficiency of LLLA make it more attractive than DKL, especially since DKL requires retraining and thus cannot simply be applied to pre-trained ReLU networks.

\section{Conclusion}
\label{sec:conclusion}

We have shown analytically that Gaussian approximations of weights distributions, when applied on binary ReLU classification networks, can mitigate the asymptotic overconfidence problem that plagues deep learning. While this behavior does not seem surprising---indeed, Gaussian-based approximate Bayesian methods have empirically been known to give good uncertainty estimates---formal statements regarding this property had been missing. Our results provide some of these statements. Furthermore, we have shown, both theoretically and empirically, that a sufficient condition for good uncertainty estimates in ReLU networks is to be ``a bit Bayesian'': apply a Gaussian-based probabilistic method, in particular a Bayesian one, to the last layer of the network. Our analysis further validates the common usage of approximate inference methods---both Bayesian and non-Bayesian---which leverage the Gaussian distribution.

\section*{Acknowledgements}
The authors gratefully acknowledge financial support by the European Research Council through ERC StG Action 757275 / PANAMA; the DFG Cluster of Excellence ``Machine Learning - New Perspectives for Science'', EXC 2064/1, project number 390727645; the German Federal Ministry of Education and Research (BMBF) through the T\"{u}bingen AI Center (FKZ: 01IS18039A); and funds from the Ministry of Science, Research and Arts of the State of Baden-W\"{u}rttemberg. AK is grateful to Alexander Meinke for the pre-trained models and the International Max Planck Research School for Intelligent Systems (IMPRS-IS) for support. AK also thanks all members of Methods of Machine Learning group for helpful feedback.

\clearpage

\bibliography{paper}
\bibliographystyle{icml2020}

\clearpage

\begin{appendices}
\crefalias{section}{appsec}

\section{Proofs}
\label{appendix:proofs}

In the followings, the norm $\Vert \cdot \Vert$ is the standard $\ell^2$-norm.

\vspace{0.5em}

\proptwotwo*

\begin{proof}
  Let $\vx \in \R^n$ be arbitrary and denote $\mu_f := f_{\vtheta_\text{MAP}}(\vx)$ and $v_f := \vd(\vx)^\top \mSigma \vd(\vx)$. For the forward direction, suppose that $\sigma(\mu_f) = 0.5$. This implies that $\mu_f = 0$, and we have $\sigma(0/(1 + \pi/8 \, v_f)^{1/2}) = \sigma(0) = 0.5$. For the reverse direction, suppose that $\sigma(\mu_f/(1 + \pi/8 \, v_f)^{1/2}) = 0.5$. This implies $\mu_f/(1 + \pi/8 \, v_f)^{1/2} = 0$. Since the denominator of the l.h.s. is positive, it follows that $\mu_f$ must be $0$, implying that $\sigma(\mu_f) = 0.5$.
\end{proof}

\vspace{0.5em}

\begin{lemma}[\citeauthor{hein2019relu}, \citeyear{hein2019relu}] \label{lemma:hein_linear_region}
  Let $\{ Q_i \}_{l=1}^R$ be the set of linear regions associated to the ReLU network $f: \R^n \to \R^k$. For any $\vx \in \R^n$ there exists an $\alpha > 0$ and $t \in \{ 1, \dots, R \}$ such that $\delta \vx \in Q_t$ for all $\delta \geq \alpha$. Furthermore, the restriction of $f$ to $Q_t$ can be written as an affine function $\mU^\top \vx + \vc$ for some suitable $\mU \in \R^{n \times k}$ and $\vc \in \R^k$. \qed
\end{lemma}

\vspace{0.5em}

\thmtwothree*

\begin{proof}
  By Lemma 3.1 of \citet{hein2019relu} (also presented in \Cref{lemma:hein_linear_region}) there exists an $\alpha > 0$ and a linear region $R$, along with $\b{u} \in \R^n$ and $c \in \R$, such that for any $\delta \geq \alpha$, we have that $\delta \vx \in R$ and the restriction $f_\vtheta \vert_R$ can be written as $\vu^\top \vx + c$. Note that, for any such $\delta$, the vector $\vu$ and scalar $c$ are constant w.r.t. $\delta\vx$. Therefore for any such $\delta$, we can write the gradient $\vd(\delta \vx)$ as follows:
  \begin{align}
    \vd(\delta \vx) &= \left. \frac{\partial (\delta \vu^\top \vx)}{\partial \vtheta} \right\vert_\vmu + \left. \frac{\partial c}{\partial \vtheta}  \right\vert_\vmu = \left. \delta \frac{\partial \vu}{\partial \vtheta} \right\vert_\vmu^\top  \vx  + \left. \frac{\partial c}{\partial \vtheta} \right\vert_\vmu \nonumber \\
          &= \delta \left( \mJ^\top \vx + \frac{1}{\delta} \nabla_\vtheta \, c \vert_\vmu \right) \, .
  \end{align}
  Hence, by \eqref{eq:z_full},
  \begin{align*}
    \abs{z&(\delta \vx)} = \frac{\abs{\delta \vu^\top \vx + c}}{\sqrt{1 + \pi/8 \, \vd(\delta \vx)^\top \b{\Sigma} \vd(\delta \vx)}} \\
          &= \frac{\abs{\delta (\vu^\top \vx + \frac{1}{\delta} c)}}{\sqrt{1 + \pi/8 \, \delta^2 (\mJ^\top \vx + \frac{1}{\delta} \nabla_\vtheta \, c \vert_\vmu)^\top \b{\Sigma} (\mJ^\top \vx + \frac{1}{\delta} \nabla_\vtheta \, c \vert_\vmu)}} \\
          &= \frac{\cancel{\delta} \abs{(\vu^\top \vx + \frac{1}{\delta} c)}}{\cancel{\delta} \sqrt{\frac{1}{\delta^2} + \pi/8 \, (\mJ^\top \vx + \frac{1}{\delta} \nabla_\vtheta \, c \vert_\vmu)^\top \b{\Sigma} (\mJ^\top \vx + \frac{1}{\delta} \nabla_\vtheta \, c \vert_\vmu)}}
  \end{align*}

  Now, notice that as $\delta \to \infty$, $1/\delta^2$ and $1/\delta$ goes to zero. So, in the limit, we have that
  \begin{equation*}
    \lim_{\delta \to \infty} \abs{z(\delta \vx)} = \frac{\abs{\vu^\top \vx}}{\sqrt{\pi/8 \, (\mJ^\top \vx)^\top \b{\Sigma} (\mJ^\top \vx)}} \, .
  \end{equation*}
  Using the Cauchy-Schwarz inequality and \Cref{lemma:min_eigval}, we can upper-bound this limit with
  \begin{equation*}
    \lim_{\delta \to \infty} \abs{z(\delta \vx)} \leq \frac{\Vert \vu \Vert \Vert \vx \Vert}{\sqrt{\pi/8 \, \lambda_\text{min}(\mSigma) \Vert \mJ^\top \vx \Vert^2}} \, .
  \end{equation*}

  The following lemma is needed to get the desired result.

  \begin{lemma} \label{lemma:singval_bound}
     Let $\mA \in \R^{m \times n}$ and $\vz \in \R^{n}$ with $m \geq n$, then $\Vert \mA \vz \Vert^2 \geq s_\text{\emph{min}}^2(\mA) \Vert \vz \Vert^2$.
  \end{lemma}

  \begin{proof}
    By SVD, $\mA = \mU \mS \mV^\top$. Notice that $\mU, \mV$ are orthogonal and thus are isometries, and that $\mS$ is a rectangular diagonal matrix with $n$ non-zero elements. Therefore,
    \begin{align}
      \Vert \mU (\mS \mV^\top \vz) \Vert^2 &= \Vert \mS \mV^\top \vz \Vert^2 = \sum_{i=1}^n s_i^2(\mA) (\mV^\top \vz)_i^2 \nonumber \\
            &\geq s_\text{min}^2(\mA) \sum_{i=1}^n (\mV^\top \vz)_i^2 \nonumber \\
            &= s_\text{min}^2(\mA) \Vert \mV^\top \vz \Vert^2 = s_\text{min}^2(\mA) \Vert \vz \Vert^2 \, ,
    \end{align}
    thus the proof is complete.
  \end{proof}

  Notice that $\mJ^\top \in \R^{p \times n}$ with $p \geq n$ by our hypothesis. Therefore, using the previous lemma on $\Vert \mJ^\top \vx \Vert^2$ in conjunction with $s_\text{min}(\mJ) = s_\text{min}(\mJ^\top)$, we conclude that
  \begin{align}
    \lim_{\delta \to \infty} \abs{z(\delta \vx)} &\leq \frac{\Vert \vu \Vert \Vert \vx \Vert}{\sqrt{\pi/8 \, \lambda_\text{min}(\mSigma) \, s_\text{min}^2(\mJ) \Vert \vx \Vert^2}} \nonumber \\
          &= \frac{\Vert \vu \Vert}{s_\text{min}(\mJ) \sqrt{\pi/8 \, \lambda_\text{min}(\mSigma)}} \, ,
  \end{align}
  thus the first result is proved.

  To prove the second statement, let $L := \lim_{\delta \to \infty} \abs{z(\delta \vx)}$. Since $L$ is the limit of $\abs{z \vert_Q (\delta \vx)}$ in the linear region $Q$ given by \Cref{lemma:hein_linear_region}, it is sufficient to show that the function $(0, \infty] \to \R$ defined by $\delta \mapsto \abs{z \vert_Q (\delta \vx)}$ is increasing.

  For some suitable choices of $\vu \in \R^n$ that depends on $\vmu$, we can write the restriction of the ``point-estimated'' ReLU network $f_\vmu \vert_Q (\vx)$ as $\vu^\top \vx$ by definition of ReLU network and since we assume that $f$ has no bias parameters. Furthermore, we let the matrix $\mJ := \frac{\partial \vu}{\partial \vtheta} \vert_\vmu$ to be the Jacobian of $\vu$ w.r.t. $\vtheta$ at $\vmu$. Therefore for any $\delta \geq \alpha$, we can write as a function of $\delta$:
  \begin{align*}
    \abs{z \vert_Q (\delta \vx)} &= \frac{\abs{\delta \vu^\top \vx}}{\sqrt{1 + \pi/8 \, \delta^2 \, (\mJ^\top \vx)^\top \mSigma (\mJ^\top \vx)}} \\
          &=: \frac{\abs{\delta a}}{\sqrt{1 + \pi/8 \, \delta^2 \, b}} \, ,
   \end{align*}
  where for simplicity we have let $a := \vu^\top \vx$ and $b := (\mJ^\top \vx)^\top \mSigma (\mJ^\top \vx)$. The derivative is therefore given by
  \begin{equation*}
    \frac{d}{d \delta} \abs{z \vert_Q (\delta \vx)} = \frac{\delta \abs{a}}{(1 + \delta^2 b)^\frac{3}{2} \abs{\delta}}
  \end{equation*}
  and since $\mSigma$ is positive-definite, it is non-negative for $\delta \in (0, \infty]$. Thus we conclude that $\abs{z \vert_Q (\delta \vx)}$ is an increasing function.
\end{proof}

\vspace{0.5em}

\thmtwofour*

\begin{proof}
  By Lemma 3.1 of \citet{hein2019relu} there exists $\alpha > 0$ and a linear region $R$, along with $\b{U} \in \R^{d \times n}$ and $\vc \in \R^d$, such that for any $\delta \geq \alpha$, we have that $\delta \vx \in R$ and the restriction $\phi \vert_R$ can be written as $\b{U} \vx + \vc$. Therefore, for any such $\delta$,
  \begin{align*}
    \abs{z \circ \phi \vert_R (\delta \vx)} &= \frac{\abs{\vmu^\top (\delta \b{U} \vx + \vc)}}{\sqrt{1 + \pi/8 \, (\delta \b{U} \vx + \vb)^\top \b{\Sigma} (\delta \b{U} \vx + \vc)}} \\
        &= \frac{\abs{\vmu^\top (\b{U} \vx + \frac{1}{\delta} \vc)}}{\sqrt{\frac{1}{\delta^2} + \pi/8 \, (\b{U} \vx + \frac{1}{\delta} \vc)^\top \b{\Sigma} (\b{U} \vx + \frac{1}{\delta} \vc)}}
  \end{align*}

  Now, notice that as $\delta \to \infty$, $1/\delta^2$ and $1/\delta$ goes to zero. So, in the limit, we have that
  \begin{equation*}
    \lim_{\delta \to \infty} \abs{z \circ \phi \vert_R (\delta \vx)} = \frac{\abs{\vmu^\top (\b{U} \vx)}}{\sqrt{\pi/8 \, (\b{U} \vx)^\top \b{\Sigma} (\b{U} \vx)}} \, .
  \end{equation*}
  We need the following lemma to obtain the bound.
  \begin{lemma} \label{lemma:min_eigval}
    Let $\vx \in \R^n$ be a vector and $\b{A} \in \R^{n \times n}$ be an SPD matrix. If $\lambda_\text{\emph{min}}(\b{A})$ is the minimum eigenvalue of $\b{A}$, then $\vx^\top \b{A} \vx \geq \lambda_\text{\emph{min}} \norm{\vx}^2$.
  \end{lemma}
  \begin{proof}
    Since $\b{A}$ is SPD, it admits an eigendecomposition $\b{A} = \b{Q} \bm{\Lambda} \b{Q}^\top$ and $\b{\Lambda} = \bm{\Lambda}^\frac{1}{2} \bm{\Lambda}^\frac{1}{2}$ makes sense. Therefore, by keeping in mind that $\b{Q}^\top \vx$ is a vector in $\R^n$, we have
    \begin{align*}
      \vx^\top \b{A} \vx &= \vx^\top \b{Q} \bm{\Lambda}^\frac{1}{2} \bm{\Lambda}^\frac{1}{2} \b{Q}^\top \vx = \norm{\bm{\Lambda}^\frac{1}{2} \b{Q}^\top \vx}^2 \\
            &= \sum_{i=1}^n \lambda_i(\b{A}) (\b{Q}^\top \vx)^2_i \geq \lambda_\text{min}(\b{A}) \sum_{i=1}^n (\b{Q}^\top \vx)^2_i \\
            &= \lambda_\text{min}(\b{A}) \norm{\b{Q}^\top \vx}^2 = \lambda_\text{min}(\b{A}) \norm{\vx}^2 \, ,
    \end{align*}
    where the last equality is obtained since $\norm{\b{Q}^\top \vx}^2 = \vx^\top \b{Q}^\top \b{Q} \vx$ and by noting that $\b{Q}$ is an orthogonal matrix.
  \end{proof}
  Using the Cauchy-Schwarz inequality and the previous lemma, we can upper-bound the limit with
  \begin{align*}
    \lim_{\delta \to \infty} \abs{z \circ \phi \vert_R (\delta \vx)} &\leq \frac{\norm{\vmu} \norm{\b{U} \vx}}{\sqrt{\pi/8 \, \lambda_\text{min}(\b{\Sigma}) \norm{\b{U} \vx}^2}} \\
          &= \frac{\norm{\vmu} }{\sqrt{\pi/8 \, \lambda_\text{min}(\b{\Sigma})}} \, ,
  \end{align*}
  which concludes the proof for the first statement.

  For the second statement, since the previous limit is the limit of $\abs{z \vert_R (\delta \vx)}$ in the linear region $R$, it is sufficient to show that the function $(0, \infty] \to \R$ defined by $\delta \mapsto \abs{z \vert_R (\delta \vx)}$ is increasing. For some $\mU \in \R^{d \times n}$ that depends on the fixed parameter of $\phi$, we write the restriction $\phi \vert_R (\vx)$ as $\mU \vx$ by definition of ReLU network and since $\phi$ is assumed to have no bias parameters. Therefore for any $\delta \geq \alpha$, we can write as a function of $\delta$:
  \begin{align*}
    \abs{z \vert_Q (\delta \vx)} &= \frac{\abs{\delta \vmu^\top \mU \vx}}{\sqrt{1 + \pi/8 \, \delta^2 \, (\mU \vx)^\top \mSigma (\mU \vx)}} \\
          &=: \frac{\abs{\delta a}}{\sqrt{1 + \pi/8 \, \delta^2 \, b}} \, ,
   \end{align*}
  where for simplicity we have let $a := \vmu^\top \mU \vx$ and $b := (\mU \vx)^\top \mSigma (\mU \vx)$. The derivative is therefore given by
  \begin{equation*}
    \frac{d}{d \delta} \abs{z \vert_Q (\delta \vx)} = \frac{\delta \abs{a}}{(1 + \delta^2 b)^\frac{3}{2} \abs{\delta}}
  \end{equation*}
  and since $\mSigma$ is positive-definite, it is non-negative for $\delta \in (0, \infty]$. Thus we conclude that $\abs{z \vert_Q (\delta \vx)}$ is an increasing function.
\end{proof}

\vspace{0.5em}

\proptwofive*

\begin{proof}
  The assumption on the prior implies that $-\log p(\vtheta) = 1/2 \, \vtheta^\top (1/\sigma^2_0 \b{I}) \vtheta + \text{const}$, which has Hessian $1/\sigma^2_0 \b{I}$. Thus, the Hessian of the negative log posterior $-\log p(\vtheta \vert \D) = -\log p(\vtheta) - \log \prod_{\vx, t \in \D} p(y \vert \vx, \vtheta)$ is $1/\sigma^2_0 \b{I} + \b{H}$. This implies that the posterior covariance $\b{\Sigma}$ of the Laplace approximation is given by
  \begin{equation}\label{eq_appendix:laplace_sigma}
    \b{\Sigma} = \left( \frac{1}{\sigma^2_0} \b{I} + \b{H} \right)^\inv \, .
  \end{equation}
  Therefore, the $i$th eigenvalue of $\b{\Sigma}$ for any $i = 1, \dots, n$ is
  $$
  \lambda_i(\b{\Sigma}) = \frac{1}{1/\sigma^2_0 + \lambda_i(\b{H})} = \frac{\sigma^2_0}{1 + \sigma^2_0 \lambda_i(\b{H})} \, .
  $$
  For all $i = 1, \dots, n$, the derivative of $\lambda_i(\b{\Sigma})$ w.r.t. $\sigma^2_0$ is $1/(1 + \sigma^2_0 \lambda_i(\b{H}))^2$ which is non-negative. This tells us that $\lambda_i(\b{\Sigma})$ is a non-decreasing function of $\sigma^2_0$. Furthermore, it is also clear that $\sigma^2_0/(1 + \sigma^2_0 \lambda_i(\b{H}))$ goes to $1/\lambda_i(\b{H})$ as $\sigma^2_0$ goes to infinity, while it goes to 0 as $\sigma^2_0$ goes to zero.

  Now, we can write.
  \begin{equation}
    \abs{z(\vx)} = \frac{\abs{f_\vmu(\vx)}}{\sqrt{1 + \pi/8 \, \sum_{i=1}^d \lambda_i(\b{\Sigma}) (\b{Q}^\top \vd)^2_i}} \, ,
  \end{equation}
  where $\b{\Sigma} = \b{Q} \, \diag{\lambda_i(\b{\Sigma}), \dots, \lambda_d(\b{\Sigma})} \, \b{Q}^\top$ is the eigendecomposition of $\b{\Sigma}$. It is therefore clear that the denominator of the r.h.s. is a non-decreasing function of $\sigma^2_0$. This implies $\abs{z(\vx)}$ is a non-increasing function of $\sigma^2_0$.

  For the limits, it is clear that $\lambdamin(\b{\Sigma})$ has limits $1/\lambdamax(\b{H})$ and $0$ whenever $\sigma^2_0 \to \infty$ and $\sigma^2 \to 0$, respectively. From these facts, the right limit is immediate from \Cref{lemma:min_eigval} while the left limit is directly obtained by noticing that the denominator goes to $1$ as $\sigma^2_0 \to 0$.
\end{proof}

\vspace{0.5em}

\begin{proposition}[Last-layer Laplace]
  Let $g: \R^d \to \R$ be a binary linear classifier defined by $g \circ \phi(\vx) := \vw^\top \phi(\vx)$ where $\phi: \R^n \to \R^d$ is a ReLU network, modeling a Bernoulli distribution $p(y | \vx, \vw) = \mathcal{B}(\sigma(g \circ \phi(\vx)))$ with parameter $\vw \in \R^d$. Let $\N(\vw | \vmu, \mSigma)$ be the posterior obtained via a Laplace approximation with prior $\N(\vtheta | \b{0}, \sigma^2_0 \mI)$ and $\mH$ be the Hessian of the negative log-likelihood at $\vmu$. Then for any input $\vx \in \R^n$, the confidence $\sigma(\abs{z(\vx)})$ is a non-increasing function of $\sigma^2_0$ with limits
  \begin{align*}
    \lim_{\sigma^2_0 \to \infty} \sigma(\abs{z(\vx)}) &\leq \sigma \left( \frac{\abs{\vmu^\top \vphi}}{1 + \sqrt{\pi/8 \, \lambda_\text{\emph{max}}(\mH) \Vert \vphi \Vert^2}} \right) \\
    \lim_{\sigma^2_0 \to 0} \sigma(\abs{z(\vx)}) &= \sigma(\abs{\vmu^\top \vphi}) \, .
  \end{align*}
\end{proposition}

\begin{proof}
  The assumption on the prior implies that $-\log p(\vw) = 1/2 \, \vw^\top (1/\sigma^2_0 \b{I}) \vw + \text{const}$, which has Hessian $1/\sigma^2_0 \b{I}$. Thus, the Hessian of the negative log posterior $-\log p(\vw \vert \D) = -\log p(\vw) - \log \prod_{\vx, t \in \D} p(y \vert \vx, \vw)$ is $1/\sigma^2_0 \b{I} + \b{H}$. This implies that the posterior covariance $\b{\Sigma}$ of the Laplace approximation is given by
  \begin{equation}\label{eq_appendix:laplace_sigma}
    \b{\Sigma} = \left( \frac{1}{\sigma^2_0} \b{I} + \b{H} \right)^\inv \, .
  \end{equation}
  Therefore, the $i$th eigenvalue of $\b{\Sigma}$ for any $i = 1, \dots, n$ is
  $$
  \lambda_i(\b{\Sigma}) = \frac{1}{1/\sigma^2_0 + \lambda_i(\b{H})} = \frac{\sigma^2_0}{1 + \sigma^2_0 \lambda_i(\b{H})} \, .
  $$
  For all $i = 1, \dots, n$, the derivative of $\lambda_i(\b{\Sigma})$ w.r.t. $\sigma^2_0$ is $1/(1 + \sigma^2_0 \lambda_i(\b{H}))^2$ which is non-negative. This tells us that $\lambda_i(\b{\Sigma})$ is a non-decreasing function of $\sigma^2_0$. Furthermore, it is also clear that $\sigma^2_0/(1 + \sigma^2_0 \lambda_i(\b{H}))$ goes to $1/\lambda_i(\b{H})$ as $\sigma^2_0$ goes to infinity, while it goes to 0 as $\sigma^2_0$ goes to zero.

  Now, we can write.
  \begin{equation}
    \abs{z(\vx)} = \frac{\abs{\vmu^\top \vphi}}{\sqrt{1 + \pi/8 \, \sum_{i=1}^d \lambda_i(\b{\Sigma}) (\b{Q}^\top \vphi)^2_i}} \, ,
  \end{equation}
  where $\b{\Sigma} = \b{Q} \, \diag{\lambda_i(\b{\Sigma}), \dots, \lambda_d(\b{\Sigma})} \, \b{Q}^\top$ is the eigendecomposition of $\b{\Sigma}$. It is therefore clear that the denominator of the r.h.s. is a non-decreasing function of $\sigma^2_0$. This implies $\abs{z(\vx)}$ is a non-increasing function of $\sigma^2_0$.

  For the limits, it is clear that $\lambdamin(\b{\Sigma})$ has limits $1/\lambdamax(\b{H})$ and $0$ whenever $\sigma^2_0 \to \infty$ and $\sigma^2 \to 0$, respectively. From these facts, the right limit is immediate from \Cref{lemma:min_eigval} while the left limit is directly obtained by noticing that the denominator goes to $1$ as $\sigma^2_0 \to 0$.
\end{proof}

\section{Laplace Approximations}
\label{appendix:applications}

The theoretical results in the main text essentially tell us that if we have a Gaussian approximate posterior that comes from a Laplace approximation, then using \cref{eq:two_class_pred} (and \cref{eq:multi_class_pred}) to make predictions can remedy the overconfidence problem on any ReLU network. In this section, we describe LLLA, DLA, and KFLA: the Laplace methods being used in the main text. For the sake of clarity, we omit biases in the following and revisit the case where biases are included at the end of this section.

\subsection{LLLA}

In the case of LLLA, we simply perform a Laplace approximation to get the posterior of the weight of the last layer $\vw$ while assuming the previous layer to be fixed. I.e.~we infer $p(\vw \vert \D) = \N(\vw \vert \vw_\text{MAP}, \b{H}^{-1})$ where $\b{H}$ is the Hessian of the negative log-posterior w.r.t. $\vw$ at $\vw_\text{MAP}$. This Hessian could be easily obtained via automatic differentiation. We emphasize that we only deal with the weight at the last layer and not the weight of the whole network, thus the inversion of $\b{H}$ is rarely a problem. For instance, even for large models such as DenseNet-201 \citep{huang2017densely} and ResNet-152 \citep{he2016deep} have $d = 1920$ and $d = 2048$ respectively,\footnote{Based on the implementations available in the TorchVision package.} implying that we only need to do the inversion of a single $1920 \times 1920$ or $2048 \times 2048$ matrix once.

In the case of multi-class classification, we now have $f: \R^d \to \R^k$ defined by $\vphi \mapsto \b{W}_\text{MAP} \vphi$. We obtain the posterior over a random matrix $\b{W} \in \R^{k \times d}$ in the form $\N(\vec(\b{W}) \vert \vec(\b{W}_\text{MAP}), \b{\Sigma})$ for some $\b{\Sigma} \in \R^{dk \times dk}$ SPD. The procedure is still similar to the one described above, since the exact Hessian of the linear multi-class classifier can still be easily and efficiently obtained via automatic differentiation. Note that in this case we need to invert a $dk \times dk$ matrix, which, depending on the size of $k$, can be quite large.\footnote{For example, the ImageNet dataset has $k = 1000$.}

For a more efficient procedure, we can make a further approximation to the posterior in the multi-class case by assuming the posterior is a matrix Gaussian distribution. We can use the Kronecker-factored Laplace approximation (KFLA) \cite{ritter_scalable_2018}, but only for the last layer of the network. That is, we find the Kronecker factorization of the Hessian $\b{H}^\inv \approx \b{V}^\inv \otimes \b{U}^\inv$ via automatic differentiation \citep{dangel2020backpack}.\footnote{In practice, we take the running average of the Kronecker factors of the Hessian over the mini-batches.} Then by definition of a matrix Gaussian \citep{gupta1999matrix}, we immediately obtain the posterior $\MN(\b{W} \vert \b{W}_\text{MAP}, \b{U}, \b{V})$. The distribution of the latent functions is Gaussian, since $\b{f} := \b{W} \vphi$ and $p(\b{W} \vert \D) = \MN(\b{W} \vert \b{W}_\text{MAP}, \b{U}, \b{V})$ imply
\begin{align} \label{eq:pred_dist_multiclass}
  p(\b{f} \vert \D) &= \MN(\b{f} \vert \b{W}_\text{MAP} \vphi, \b{U}, \vphi^\top \b{V} \vphi) \nonumber \\
        &= \N(\b{f} \vert \b{W}_\text{MAP} \vphi, (\vphi^\top \b{V} \vphi) \otimes \b{U}) \nonumber \\
        &= \N(\b{f} \vert \b{W}_\text{MAP} \vphi, (\vphi^\top \b{V} \vphi) \b{U}) \, ,
\end{align}
where the last equality follows since $(\vphi^\top \b{V} \vphi)$ is a scalar. We then have the following integral
\begin{align*}
  p(y = i \vert &\vx, \D) = \\
      &\int \mathrm{softmax}(\b{f}, i) \, \N(\b{f} \vert \b{W}_\text{MAP} \vphi, (\vphi^\top \b{V} \vphi) \b{U}) \, d\b{f} \, ,
\end{align*}
which can be approximated via a MC-integral.

While one can always assume that the bias trick is already used, i.e.~it is absorbed in the weight matrix/vector, in practice when dealing with pre-trained networks, one does not have such liberty. In this case, one can simply assume that the bias $b$ or $\vb$ is independent of the weight $\vw$ or $\b{W}$, respectively in the two- and multi-class cases. By using the same Laplace approximation procedure, one can easily get $p(b \vert \D) := \N(b | \mu_b, \sigma^2_b)$ or $p(\vb \vert \D) := \N(\vb \vert \vmu_b, \b{\Sigma}_b)$. This implies $\vw^\top \vphi + b =: f$ and $\b{W} \vphi + \vb =: \b{f}$ are also Gaussians given by
\begin{align}
  \N(f \vert \vmu^\top \vphi+ \mu_b, \vphi^\top \b{H}^\inv \vphi + \sigma^2_b) \\
  \shortintertext{or}
  \N(\b{f} \vert \b{M} \vphi + \vb, (\vphi^\top \otimes \b{I}) \b{\Sigma} (\vphi \otimes \b{I}) + \b{\Sigma}_b) \, ,
\end{align}
respectively, with $\b{I} \in \R^{k \times k}$ if $\b{W} \in \R^{k \times d}$ and $\vphi \in \R^d$. Similarly, in the case when the Kronecker-factored approximation is used, we have
\begin{equation}
  p(\b{f} \vert \D) = \N(\b{f} \vert \b{W}_\text{MAP} \vphi + \vmu_b, (\vphi^\top \b{V} \vphi) \b{U} + \b{\Sigma}_b) \, .
\end{equation}

We present the pseudocodes of LLLA in \Cref{algo:llla_exact_twoclass,algo:llla_kf_multiclass}.

\begin{algorithm}[h!]
  \caption{LLLA with exact Hessian for binary classification.}
  \label{algo:llla_exact_twoclass}
  \begin{algorithmic}[1]
        \Require
            \Statex A pre-trained network $f \circ \phi$ with $\b{w}_\text{MAP}$ as the weight of $f$, (averaged) cross-entropy loss $\mathcal{L}$, training set $\D_\text{train}$, test set $\D_\text{test}$, mini-batch size $m$, running average weighting $\rho$, and prior precision $\tau_0 = 1/\sigma^2_0$.
        \Ensure
            \Statex Predictions $\mathcal{P}$ containing $p(y = 1 \vert \vx, \D_\text{train}) \, \forall \vx \in \D_\text{test}$.
        \State $\b{\Lambda} = 0 \in \R^{d \times d}$

        \For{$i = 1, \dots, \abs{\D_\text{train}}/m$}
          \State $\b{X}_i, \b{y}_i = \mathrm{sampleMinibatch}(\D_\text{train}, m)$
          \State $\b{A}_i, \b{B}_i = \mathrm{getHessian}(\mathcal{L}(f \circ \phi(\b{X}_i), \b{y}_i), \b{w}_\text{MAP})$
          \State $\b{\Lambda} = \rho \b{\Lambda} + (1-\rho) \b{\Lambda}_i$
        \EndFor

        \State $\b{\Sigma} = (\abs{\D_\text{train}} \, \b{\Lambda} + \tau_0 \, \b{I})^\inv$
        \State $p(\b{w} \vert \D) = \N(\b{w} \vert \b{w}_\text{MAP}, \b{\Sigma})$
        \State $\mathcal{Y} = \varnothing$

        \ForAll{$\vx \in \D_\text{test}$}
          \State $y = \sigma(\vw_\text{MAP}^\top \vphi / (1 + \pi/8 \, \vphi^\top \b{\Sigma} \vphi)^{1/2})$
          \State $\mathcal{Y} = \mathcal{P} \cup \{ y \}$
        \EndFor
  \end{algorithmic}
\end{algorithm}

\begin{algorithm}[t!]
  \caption{LLLA with Kronecker-factored Hessian for multi-class classification.}
  \label{algo:llla_kf_multiclass}
  \begin{algorithmic}[1]
        \Require
            \Statex A pre-trained network $f \circ \phi$ with $\b{W}_\text{MAP}$ as the weight of $f$, (averaged) cross-entropy loss $\mathcal{L}$, training set $\D_\text{train}$, test set $\D_\text{test}$, mini-batch size $m$, number of samples $s$, running average weighting $\rho$, and prior precision $\tau_0 = 1/\sigma^2_0$.
        \Ensure
            \Statex Predictions $\mathcal{P}$ containing $p(y = i \vert \vx, \D_\text{train}) \, \forall \vx \in \D_\text{test} \, \forall i \in \{ 1, \dots, k \}$.
        \State $\b{A} = 0 \in \R^{k \times k}, \b{B} = 0 \in \R^{d \times d}$

        \For{$i = 1, \dots, \abs{\D_\text{train}}/m$}
          \State $\b{X}_i, \b{y}_i = \mathrm{sampleMinibatch}(\D_\text{train}, m)$
          \State $\b{A}_i, \b{B}_i = \mathrm{KronFactors}(\mathcal{L}(f \circ \phi(\b{X}_i), \b{y}_i), \b{W}_\text{MAP})$
          \State $\b{A} = \rho \b{A} + (1-\rho) \b{A}_i$
          \State $\b{B} = \rho \b{B} + (1-\rho) \b{B}_i$
        \EndFor

        \State $\b{U} = (\sqrt{\abs{\D_\text{train}}} \, \b{A} + \sqrt{\tau_0} \, \b{I})^\inv$
        \State $\b{V} = (\sqrt{\abs{\D_\text{train}}} \, \b{B} + \sqrt{\tau_0} \, \b{I})^\inv$
        \State $p(\b{W} \vert \D) = \MN(\b{W} \vert \b{W}_\text{MAP}, \b{U}, \b{V})$
        \State $\mathcal{Y} = \varnothing$

        \ForAll{$\vx \in \D_\text{test}$}
          \State $p(\b{f} \vert \D) = \N(\b{f} \vert \b{W}_\text{MAP} \vphi, (\vphi^\top \b{V} \vphi) \b{U})$
          \State $\b{y} = \b{0}$
          \For{$j = 1, \dots, s$}
            \State $\b{f}_j \sim p(\b{f} \vert \D)$
            \State $\b{y} = \b{y} + \mathrm{softmax}(\b{f}_j)$
          \EndFor
          \State $\b{y} = \b{y} / m$
          \State $\mathcal{Y} = \mathcal{Y} \cup \{ \b{y} \}$
        \EndFor
  \end{algorithmic}
\end{algorithm}

\subsection{DLA}

In this method, we aim at inferring the \emph{diagonal} of the covariance of the Gaussian over the whole layer of a network. Instead of using the exact diagonal Hessian, we use the diagonal of the Fisher information matrix $\mF$ of the network \citep{ritter_scalable_2018} as follows
\begin{align*}
  \diag{\mSigma} &\approx (\sigma^2_0 + \diag{\mF})^\inv \\
      &= (\sigma^2_0 + \E_{\vy \sim p(\vy | \vx, \vtheta), \vx \sim \D}(\nabla_\vtheta \, p(\vy | \vx, \vtheta))^2)^\inv \, .
\end{align*}
Thus, one simply needs to do several backpropagation to compute the gradients of all weight matrices of the network. This gives rise to the Gaussian posterior $\N(\vtheta | \vtheta_\text{MAP}, \diag{\mSigma})$. During prediction, an MC-integration scheme is employed: we repeatedly sample a whole network and average their predictions. That is, for each layer $l \in \{ 1, \dots, L \}$, we sample the $l$th layer's weight matrix $\mW^l \sim \N(\mW^l | \mW^l_\text{MAP}, \diag{\mSigma})$ by computing
\begin{align*}
  \ve &\sim \N(\b{0}, \mI) \\
  \vec(\mW^l) &= \vec(\mW^l_\text{MAP} + \ve \odot \diag{\mSigma_l}^{\frac{1}{2}}) \, ,
\end{align*}
where we have denoted the covariance matrix of the $l$th layer as $\mSigma_l$. Note that, the computational cost for doing prediction scales with the size of the network, thus this scheme is already orders of magnitude more expensive than LLLA, cf. \Cref{tab:timing}.

\subsection{KFLA}

LLLA with a matrix normal distribution as described in the previous section is a special case of KFLA \citep{ritter_scalable_2018}. In KFLA, similar to DLA, we aim to infer the posterior of the whole network parameters and not just those of the last layer. Concretely, for each layer $l \in \{ 1, \dots, L \}$, we infer the posterior
\begin{align*}
  p(\mW^l | \D) &\approx \MN(\mW^l | \mW^l_\text{MAP}, \mU^l, \mV^l) \\
      &= \N(\vec(\mW^l) | \vec(\mW^l_\text{MAP}), \mV^l \otimes \mU^l) \, ,
\end{align*}
where
\begin{align*}
  \mU^l &= (\sqrt{\abs{\D}} \mA + 1/\sigma^2_0 \mI)^\inv \, , \\
  \mV^l &= (\sqrt{\abs{\D}} \mB + 1/\sigma^2_0 \mI)^\inv \, ,
\end{align*}
and $\mA, \mB$ are the Kronecker-factors---e.g. obtained KFAC \citep{martens2015optimizing}---of the Hessian of the loss w.r.t. $\mW^l$.

During predictions, as in DLA, we also use MC-integration to compute the posterior predictive distribution. That is, at each layer $l \in \{ 1, \dots, L \}$, we sample the $l$th layer's weight matrix $\mW^l$ via
\begin{align*}
  \mE &\sim \N(\b{0}, \mI) \\
  \mW^l &= \mW^l_\text{MAP} + (\mU^l)^\frac{1}{2} \mE (\mV^l)^\frac{1}{2} \, ,
\end{align*}
where $\mS, \mT$ are the Cholesky factors such that $(\mU^l)^\frac{1}{2}(\mU^\frac{1}{2})^\top = \mU$ and $(\mV^\frac{1}{2})^\top\mV^\frac{1}{2} = \mV$. Again, the cost for doing prediction scales with the size of the network, and it is clear that KFLA is more expensive than DLA.

\section{Training Detail}
\label{appendix:training}

We train all networks we use in \Cref{tab:ood_results} for 100 epochs with batch size of 128. We use ADAM and SGD with 0.9 momentum with the initial learning rates of 0.001 and 0.1 for MNIST and CIFAR-10/SVHN/CIFAR1-00 experiments, respectively, and we divide them by 10 at epoch 50, 75, and 95. Standard data augmentations, i.e.~random crop and standardization are also used for training the network on CIFAR-10. We use a graphic card with $11$GB memory for all computation.

\section{Further Experiments}
\label{appendix:more_experiments}

\subsection{Non-Bayesian Baselines}

\begin{figure*}[t]
  \centering

  \subfloat{\includegraphics[width=0.22\textwidth, height=0.15\textwidth]{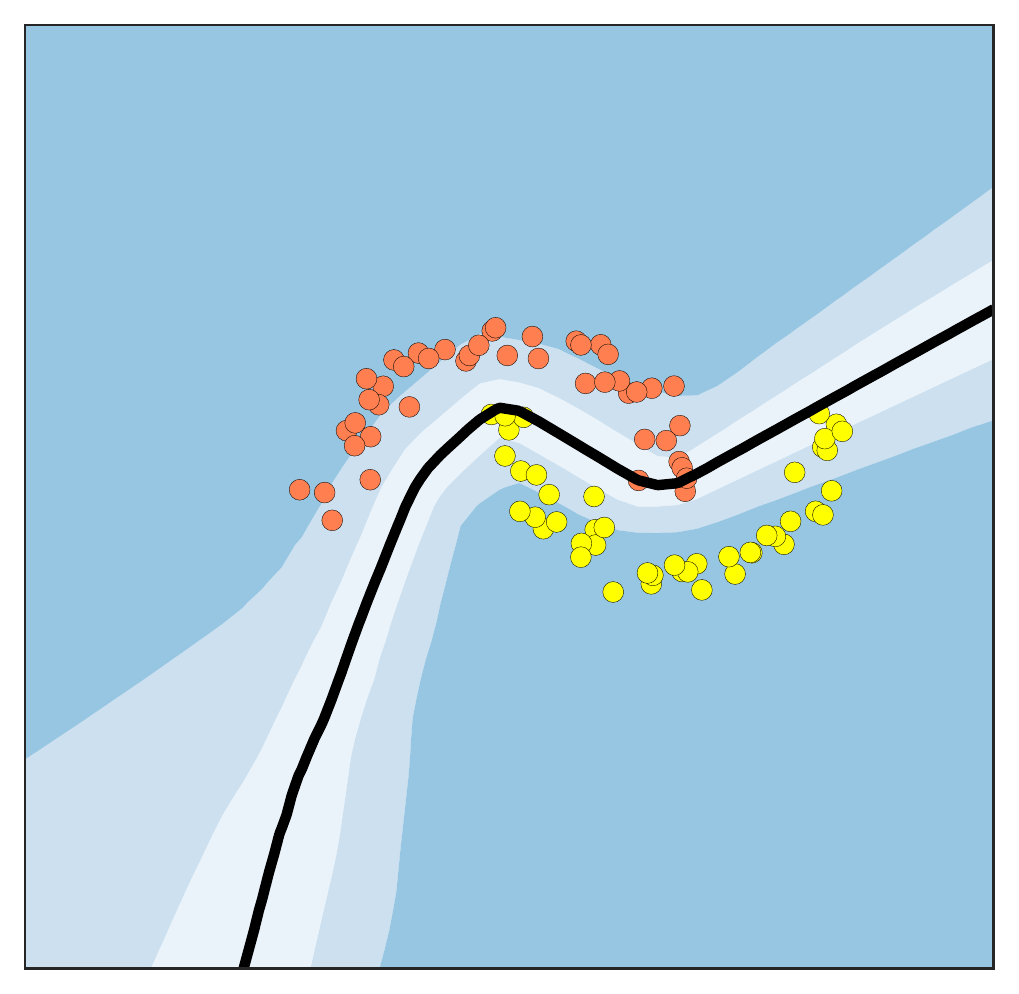}}
  \quad
  \subfloat{\includegraphics[width=0.22\textwidth, height=0.15\textwidth]{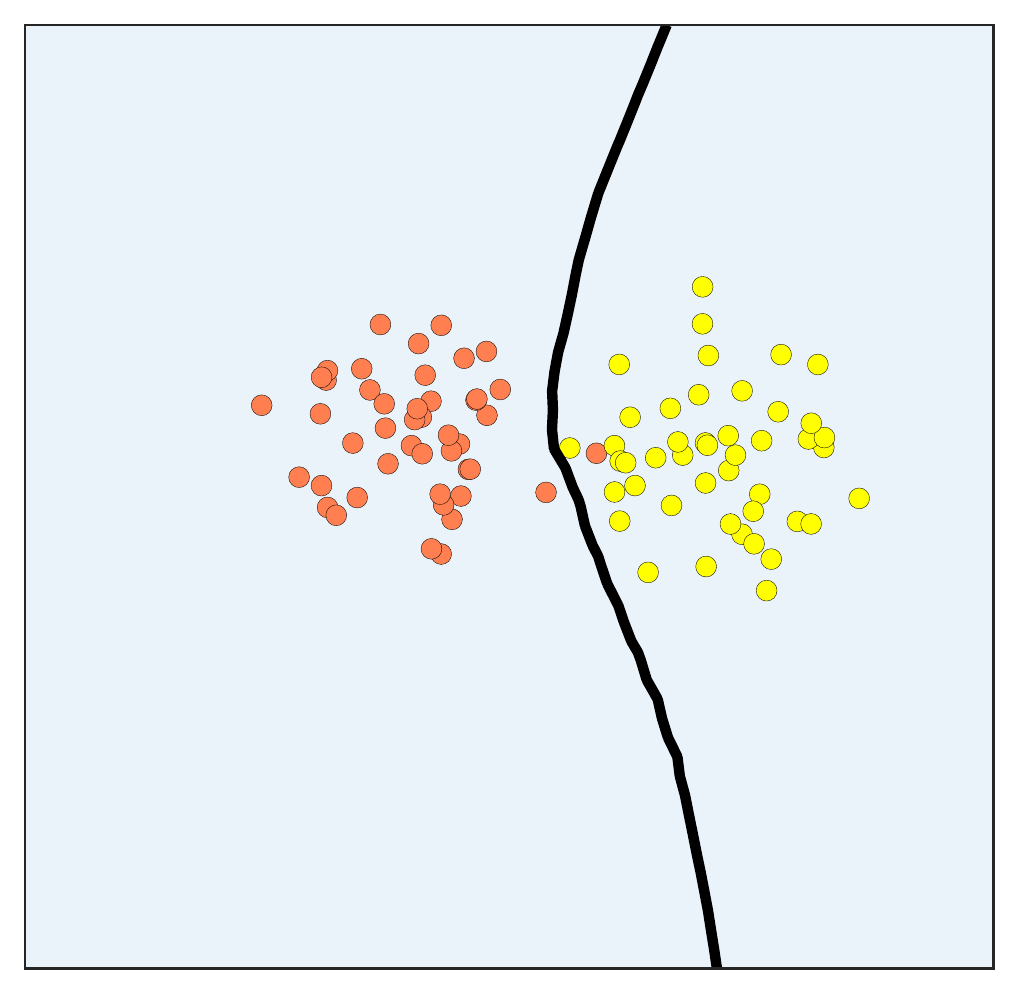}}
  \quad
  \subfloat{\includegraphics[width=0.03\textwidth, height=0.15\textwidth]{toy_2d_bnn_colorbar}}
  \qquad
  \qquad
  \subfloat{\includegraphics[width=0.22\textwidth, height=0.15\textwidth]{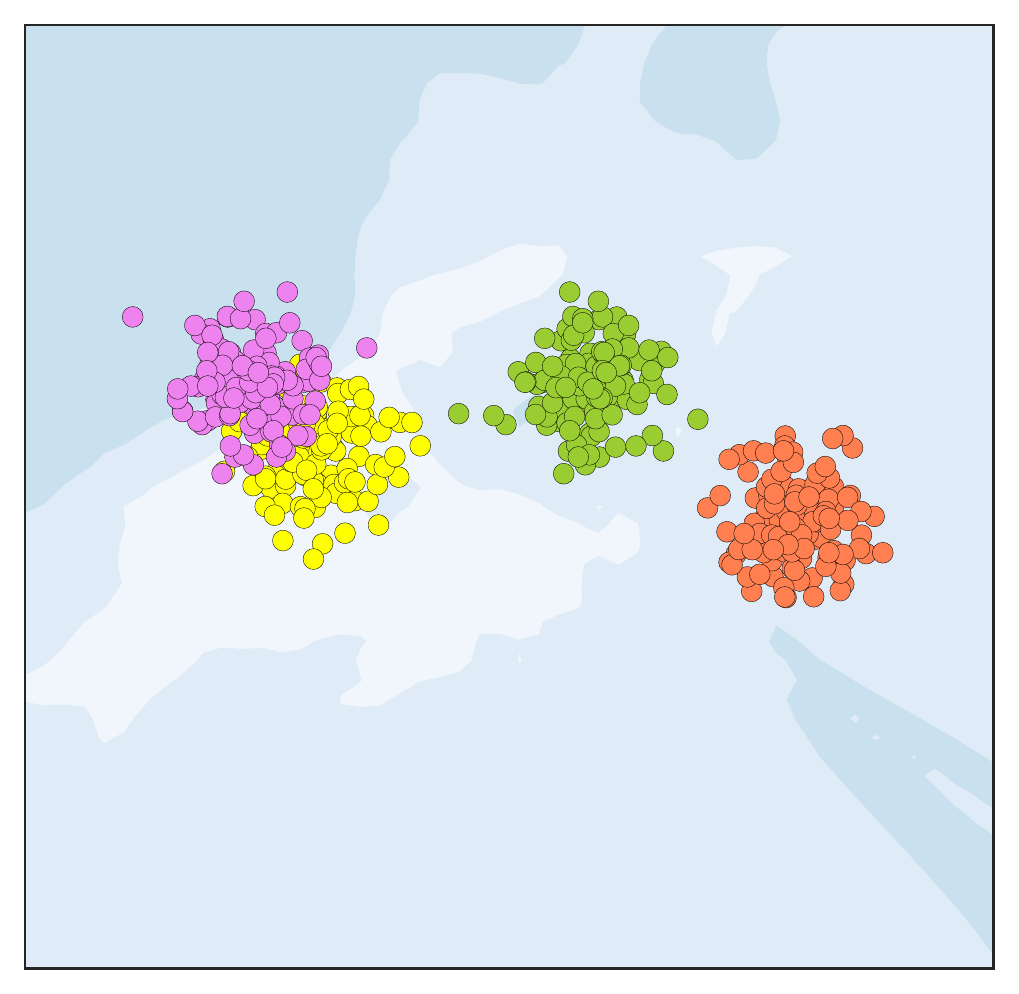}}
  \quad
  \subfloat{\includegraphics[width=0.03\textwidth, height=0.15\textwidth]{toy_2d_nn_multiclass_colorbar}}

  \caption{Binary and multi-class toy classification results using a simple isotropic Gaussian baseline. Black lines represent decision boundary, shades represent confidence.}

  \label{fig:toy_iso}
\end{figure*}

To represent non-Bayesian Gaussian approximations, we use the following simple baseline: Given a ReLU network, we assume that the distribution over the last-layer's weights is an isotropic Gaussian $\N(\mathbf{0}, \sigma^2_0 \mI)$, where $\sigma^2_0$ is found via cross-validation, optimizing \eqref{eq:opt_var_prior}. The results on toy datasets are presented in \Cref{fig:toy_iso}. We found that our theoretical analysis hold under this setup, in the sense that far-away from the training data, the confidence is constant less than one. However, predictions around the training data lack structure, unlike the predictions of Bayesian methods. This is because an isotropic Gaussian is too simple and does not capture the structure of the training data. In contrast, Bayesian methods, in particular Laplace approximations, capture this structure in the Hessian of the negative log-likelihood.

\subsection{Histograms}

To give a more fine-grained perspective of the results in \Cref{tab:ood_faraway,tab:ood_results}, we show the histograms in \Cref{fig:histograms_binary,fig:histograms_multiclass}. The histograms of both the in-distribution data and far-away OOD data are close together in both MAP and temperature scaling methods, leading to low AUR scores. Meanwhile LLLA (representing Bayesian methods) yields clear separations.

\subsection{Asymptotic Confidence of Multi-class Problems}

\begin{figure*}[t!]
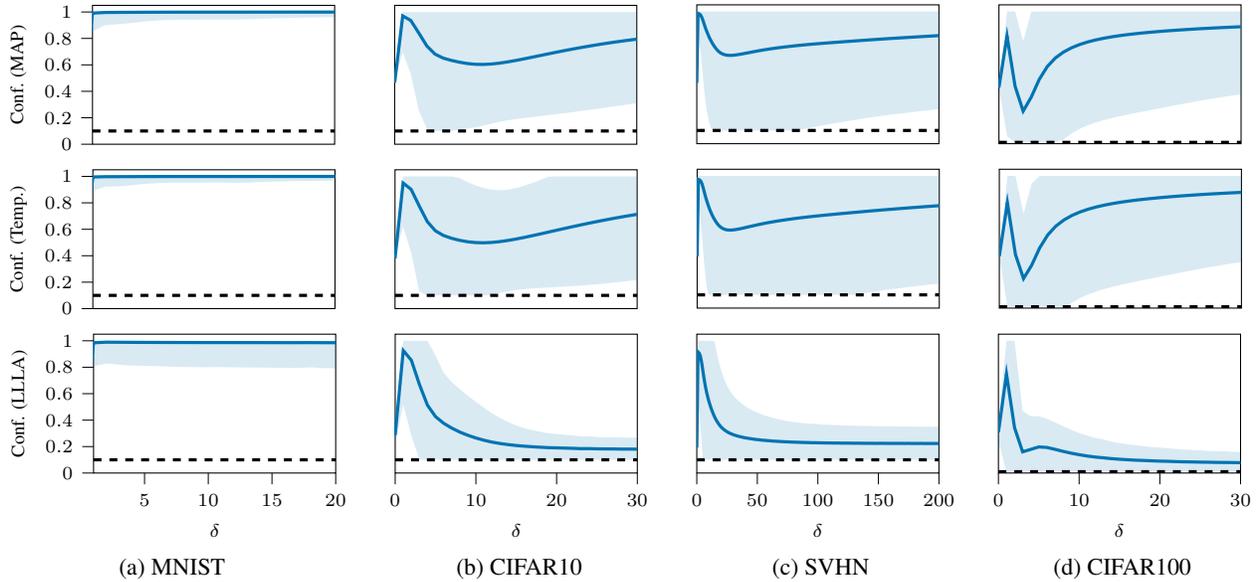

  \centering

  \subfloat{
\begin{tikzpicture}[baseline, trim axis right]

\tikzstyle{every node}=[font=\scriptsize]

\definecolor{color0}{rgb}{0.00392156862745098,0.450980392156863,0.698039215686274}

\begin{axis}[
width=0.28\textwidth,
height=0.2\textwidth,
tick align=outside,
tick pos=left,
x grid style={white!69.01960784313725!black},
xmin=1, xmax=20,
xtick style={color=black},
xtick={0,5,10,15,20},
y grid style={white!69.01960784313725!black},
ylabel={Conf. (MAP)},
ymin=0, ymax=1.05,
ytick style={color=black},
ytick={0,0.2,0.4,0.6,0.8,1,1.2},
xmajorticks=false,
]
\path [fill=color0, fill opacity=0.15, semithick]
(axis cs:0,0.118548311293125)
--(axis cs:0,0.118548177182674)
--(axis cs:1,0.854592084884644)
--(axis cs:2,0.900472700595856)
--(axis cs:3,0.907536268234253)
--(axis cs:4,0.91640031337738)
--(axis cs:5,0.927592098712921)
--(axis cs:6,0.935017347335815)
--(axis cs:7,0.93975442647934)
--(axis cs:8,0.940648555755615)
--(axis cs:9,0.941580593585968)
--(axis cs:10,0.943039059638977)
--(axis cs:11,0.943501889705658)
--(axis cs:12,0.942792177200317)
--(axis cs:13,0.944563388824463)
--(axis cs:14,0.948237836360931)
--(axis cs:15,0.951359808444977)
--(axis cs:16,0.953939437866211)
--(axis cs:17,0.956147909164429)
--(axis cs:18,0.957683086395264)
--(axis cs:19,0.959211647510529)
--(axis cs:20,0.961299777030945)
--(axis cs:20,1)
--(axis cs:20,1)
--(axis cs:19,1)
--(axis cs:18,1)
--(axis cs:17,1)
--(axis cs:16,1)
--(axis cs:15,1)
--(axis cs:14,1)
--(axis cs:13,1)
--(axis cs:12,1)
--(axis cs:11,1)
--(axis cs:10,1)
--(axis cs:9,1)
--(axis cs:8,1)
--(axis cs:7,1)
--(axis cs:6,1)
--(axis cs:5,1)
--(axis cs:4,1)
--(axis cs:3,1)
--(axis cs:2,1)
--(axis cs:1,1)
--(axis cs:0,0.118548311293125)
--cycle;

\addplot [very thick, color0]
table {%
0 0.1185482442379
1 0.991712749004364
2 0.99634861946106
3 0.997079074382782
4 0.997576594352722
5 0.998046875
6 0.998381435871124
7 0.998600363731384
8 0.998714804649353
9 0.998799324035645
10 0.998873233795166
11 0.99892783164978
12 0.998964846134186
13 0.999029338359833
14 0.999112069606781
15 0.999183475971222
16 0.999244332313538
17 0.999296486377716
18 0.999337911605835
19 0.999374985694885
20 0.999413013458252
};
\addplot [very thick, black, dashed]
table {%
1 0.1
20 0.1
};
\end{axis}

\end{tikzpicture}}
  \qquad
  \subfloat{
\begin{tikzpicture}[baseline]

\tikzstyle{every node}=[font=\scriptsize]

\definecolor{color0}{rgb}{0.00392156862745098,0.450980392156863,0.698039215686274}

\begin{axis}[
width=0.28\textwidth,
height=0.2\textwidth,
tick align=outside,
tick pos=left,
x grid style={white!69.01960784313725!black},
xmin=0, xmax=30,
xtick style={color=black},
xtick={0,10,20,30},
y grid style={white!69.01960784313725!black},
ymin=0, ymax=1.05,
ytick style={color=black},
ytick={0,0.2,0.4,0.6,0.8,1,1.2},
xmajorticks=false,
ymajorticks=false,
]
\path [fill=color0, fill opacity=0.15, semithick]
(axis cs:0,0.466141253709793)
--(axis cs:0,0.466141074895859)
--(axis cs:1,0.69603419303894)
--(axis cs:2,0.529141664505005)
--(axis cs:3,0.254213631153107)
--(axis cs:4,0.116498827934265)
--(axis cs:5,0.100000001490116)
--(axis cs:6,0.100000001490116)
--(axis cs:7,0.10140734910965)
--(axis cs:8,0.11016458272934)
--(axis cs:9,0.118322849273682)
--(axis cs:10,0.12809956073761)
--(axis cs:11,0.141211926937103)
--(axis cs:12,0.155360162258148)
--(axis cs:13,0.168927609920502)
--(axis cs:14,0.180571854114532)
--(axis cs:15,0.190185070037842)
--(axis cs:16,0.198096692562103)
--(axis cs:17,0.205271244049072)
--(axis cs:18,0.212209582328796)
--(axis cs:19,0.219620287418365)
--(axis cs:20,0.22686368227005)
--(axis cs:21,0.234084606170654)
--(axis cs:22,0.241879761219025)
--(axis cs:23,0.250024676322937)
--(axis cs:24,0.258216440677643)
--(axis cs:25,0.266704589128494)
--(axis cs:26,0.275466203689575)
--(axis cs:27,0.284411191940308)
--(axis cs:28,0.293108940124512)
--(axis cs:29,0.301727563142776)
--(axis cs:30,0.310366958379745)
--(axis cs:30,1)
--(axis cs:30,1)
--(axis cs:29,1)
--(axis cs:28,1)
--(axis cs:27,1)
--(axis cs:26,1)
--(axis cs:25,1)
--(axis cs:24,1)
--(axis cs:23,1)
--(axis cs:22,1)
--(axis cs:21,1)
--(axis cs:20,1)
--(axis cs:19,1)
--(axis cs:18,1)
--(axis cs:17,1)
--(axis cs:16,1)
--(axis cs:15,1)
--(axis cs:14,1)
--(axis cs:13,1)
--(axis cs:12,1)
--(axis cs:11,1)
--(axis cs:10,1)
--(axis cs:9,1)
--(axis cs:8,1)
--(axis cs:7,1)
--(axis cs:6,1)
--(axis cs:5,1)
--(axis cs:4,1)
--(axis cs:3,1)
--(axis cs:2,1)
--(axis cs:1,1)
--(axis cs:0,0.466141253709793)
--cycle;

\addplot [very thick, color0]
table {%
0 0.466141164302826
1 0.970477879047394
2 0.935900390148163
3 0.840181648731232
4 0.742160618305206
5 0.681387364864349
6 0.651420772075653
7 0.634106814861298
8 0.620611727237701
9 0.61015397310257
10 0.604320049285889
11 0.603319227695465
12 0.606497406959534
13 0.613252937793732
14 0.622448205947876
15 0.633267343044281
16 0.645105719566345
17 0.657545864582062
18 0.670207738876343
19 0.682900607585907
20 0.695312082767487
21 0.707337617874146
22 0.718997776508331
23 0.730195701122284
24 0.740859091281891
25 0.751024544239044
26 0.76069700717926
27 0.76988023519516
28 0.77853786945343
29 0.786716282367706
30 0.794462144374847
};
\addplot [very thick, black, dashed]
table {%
0 0.1
30 0.1
};
\end{axis}

\end{tikzpicture}}
  \qquad
  \subfloat{\input{figs/conf_map_svhn}}
  \qquad
  \subfloat{
\begin{tikzpicture}

\tikzstyle{every node}=[font=\scriptsize]

\definecolor{color0}{rgb}{0.00392156862745098,0.450980392156863,0.698039215686274}

\begin{axis}[
width=0.28\textwidth,
height=0.2\textwidth,
tick align=outside,
tick pos=left,
x grid style={white!69.01960784313725!black},
xmin=0, xmax=30,
xtick style={color=black},
xtick={0,10,20,30},
xticklabels={\(\displaystyle 0\),\(\displaystyle 10\),\(\displaystyle 20\),\(\displaystyle 30\)},
ymin=0, ymax=1.05,
ytick style={color=black},
ytick={0,0.2,0.4,0.6,0.8,1,1.2},
xmajorticks=false, ymajorticks=false,
]
\path [fill=color0, fill opacity=0.15, semithick]
(axis cs:0,0.426094442605972)
--(axis cs:0,0.426093906164169)
--(axis cs:1,0.0564219951629639)
--(axis cs:2,0.00999999977648258)
--(axis cs:3,0.00999999977648258)
--(axis cs:4,0.00999999977648258)
--(axis cs:5,0.00999999977648258)
--(axis cs:6,0.00999999977648258)
--(axis cs:7,0.00999999977648258)
--(axis cs:8,0.0307151079177856)
--(axis cs:9,0.0771908760070801)
--(axis cs:10,0.111881077289581)
--(axis cs:11,0.138957023620605)
--(axis cs:12,0.160605669021606)
--(axis cs:13,0.178580641746521)
--(axis cs:14,0.193971037864685)
--(axis cs:15,0.207625985145569)
--(axis cs:16,0.220550894737244)
--(axis cs:17,0.232825994491577)
--(axis cs:18,0.244471549987793)
--(axis cs:19,0.256434738636017)
--(axis cs:20,0.268547832965851)
--(axis cs:21,0.280450403690338)
--(axis cs:22,0.291935741901398)
--(axis cs:23,0.303245306015015)
--(axis cs:24,0.3146111369133)
--(axis cs:25,0.325919032096863)
--(axis cs:26,0.336328744888306)
--(axis cs:27,0.345979690551758)
--(axis cs:28,0.35566771030426)
--(axis cs:29,0.365026295185089)
--(axis cs:30,0.373785436153412)
--(axis cs:30,1)
--(axis cs:30,1)
--(axis cs:29,1)
--(axis cs:28,1)
--(axis cs:27,1)
--(axis cs:26,1)
--(axis cs:25,1)
--(axis cs:24,1)
--(axis cs:23,1)
--(axis cs:22,1)
--(axis cs:21,1)
--(axis cs:20,1)
--(axis cs:19,1)
--(axis cs:18,1)
--(axis cs:17,1)
--(axis cs:16,1)
--(axis cs:15,1)
--(axis cs:14,1)
--(axis cs:13,1)
--(axis cs:12,1)
--(axis cs:11,1)
--(axis cs:10,1)
--(axis cs:9,1)
--(axis cs:8,1)
--(axis cs:7,1)
--(axis cs:6,1)
--(axis cs:5,1)
--(axis cs:4,0.9973104596138)
--(axis cs:3,0.77384877204895)
--(axis cs:2,1)
--(axis cs:1,1)
--(axis cs:0,0.426094442605972)
--cycle;

\addplot [very thick, color0]
table {%
0 0.426094174385071
1 0.814106822013855
2 0.434225171804428
3 0.246428161859512
4 0.351588368415833
5 0.486854076385498
6 0.584090650081635
7 0.648577094078064
8 0.692956328392029
9 0.725358247756958
10 0.749941647052765
11 0.769253969192505
12 0.784803569316864
13 0.797611474990845
14 0.808331429958344
15 0.817448854446411
16 0.825365662574768
17 0.832311749458313
18 0.838459968566895
19 0.844063103199005
20 0.849205195903778
21 0.85392838716507
22 0.858268678188324
23 0.862304151058197
24 0.866105258464813
25 0.869696617126465
26 0.873000025749207
27 0.876057386398315
28 0.878991484642029
29 0.881772994995117
30 0.884386003017426
};
\addplot [very thick, black, dashed]
table {%
0 0.01
30 0.01
};
\end{axis}

\end{tikzpicture}}

  \vspace{-1em}
  \setcounter{subfigure}{0}

  \subfloat{
\begin{tikzpicture}[baseline, trim axis right]

\tikzstyle{every node}=[font=\scriptsize]

\definecolor{color0}{rgb}{0.00392156862745098,0.450980392156863,0.698039215686274}

\begin{axis}[
width=0.28\textwidth,
height=0.2\textwidth,
tick align=outside,
tick pos=left,
x grid style={white!69.01960784313725!black},
xmin=1, xmax=20,
xtick style={color=black},
xtick={0,5,10,15,20},
y grid style={white!69.01960784313725!black},
ylabel={Conf. (Temp.)},
ymin=0, ymax=1.05,
ytick style={color=black},
ytick={0,0.2,0.4,0.6,0.8,1,1.2},
xmajorticks=false,
]
\path [fill=color0, fill opacity=0.15, semithick]
(axis cs:0,0.128901824355125)
--(axis cs:0,0.128901645541191)
--(axis cs:1,0.891842901706696)
--(axis cs:2,0.922606885433197)
--(axis cs:3,0.924082219600677)
--(axis cs:4,0.931966602802277)
--(axis cs:5,0.942775249481201)
--(axis cs:6,0.948576092720032)
--(axis cs:7,0.95259964466095)
--(axis cs:8,0.951796770095825)
--(axis cs:9,0.952187418937683)
--(axis cs:10,0.953813195228577)
--(axis cs:11,0.95338100194931)
--(axis cs:12,0.950611054897308)
--(axis cs:13,0.952053189277649)
--(axis cs:14,0.956585109233856)
--(axis cs:15,0.959962487220764)
--(axis cs:16,0.96235191822052)
--(axis cs:17,0.96424663066864)
--(axis cs:18,0.965183794498444)
--(axis cs:19,0.966412127017975)
--(axis cs:20,0.968945264816284)
--(axis cs:20,1)
--(axis cs:20,1)
--(axis cs:19,1)
--(axis cs:18,1)
--(axis cs:17,1)
--(axis cs:16,1)
--(axis cs:15,1)
--(axis cs:14,1)
--(axis cs:13,1)
--(axis cs:12,1)
--(axis cs:11,1)
--(axis cs:10,1)
--(axis cs:9,1)
--(axis cs:8,1)
--(axis cs:7,1)
--(axis cs:6,1)
--(axis cs:5,1)
--(axis cs:4,1)
--(axis cs:3,1)
--(axis cs:2,1)
--(axis cs:1,1)
--(axis cs:0,0.128901824355125)
--cycle;

\addplot [very thick, color0]
table {%
0 0.128901734948158
1 0.995771110057831
2 0.997726082801819
3 0.998000025749207
4 0.998333632946014
5 0.998709619045258
6 0.998954713344574
7 0.999100804328918
8 0.999144554138184
9 0.999178767204285
10 0.999218940734863
11 0.999238848686218
12 0.999236226081848
13 0.999282121658325
14 0.999362230300903
15 0.999429106712341
16 0.999482870101929
17 0.999526917934418
18 0.999558091163635
19 0.999586820602417
20 0.999620616436005
};
\addplot [very thick, black, dashed]
table {%
1 0.1
20 0.1
};
\end{axis}

\end{tikzpicture}}
  \qquad
  \subfloat{
\begin{tikzpicture}[baseline]

\tikzstyle{every node}=[font=\scriptsize]

\definecolor{color0}{rgb}{0.00392156862745098,0.450980392156863,0.698039215686274}

\begin{axis}[
width=0.28\textwidth,
height=0.2\textwidth,
tick align=outside,
tick pos=left,
x grid style={white!69.01960784313725!black},
xmin=0, xmax=30,
xtick style={color=black},
xtick={0,10,20,30},
xticklabels={\(\displaystyle 0\),\(\displaystyle 10\),\(\displaystyle 20\),\(\displaystyle 30\)},
y grid style={white!69.01960784313725!black},
ymin=0, ymax=1.05,
ytick style={color=black},
ytick={0,0.2,0.4,0.6,0.8,1,1.2},
xmajorticks=false,
ymajorticks=false,
yticklabels={\(\displaystyle 0.0\),\(\displaystyle 0.2\),\(\displaystyle 0.4\),\(\displaystyle 0.6\),\(\displaystyle 0.8\),\(\displaystyle 1.0\),\(\displaystyle 1.2\)}
]
\path [fill=color0, fill opacity=0.15, semithick]
(axis cs:0,0.379381686449051)
--(axis cs:0,0.379381328821182)
--(axis cs:1,0.622612118721008)
--(axis cs:2,0.42354691028595)
--(axis cs:3,0.131144642829895)
--(axis cs:4,0.100000001490116)
--(axis cs:5,0.100000001490116)
--(axis cs:6,0.100000001490116)
--(axis cs:7,0.100000001490116)
--(axis cs:8,0.100000001490116)
--(axis cs:9,0.100000001490116)
--(axis cs:10,0.100000001490116)
--(axis cs:11,0.100000001490116)
--(axis cs:12,0.101723611354828)
--(axis cs:13,0.117922306060791)
--(axis cs:14,0.130957305431366)
--(axis cs:15,0.140647768974304)
--(axis cs:16,0.147496581077576)
--(axis cs:17,0.152480214834213)
--(axis cs:18,0.156321376562119)
--(axis cs:19,0.159866511821747)
--(axis cs:20,0.163073539733887)
--(axis cs:21,0.16627824306488)
--(axis cs:22,0.170013785362244)
--(axis cs:23,0.174265265464783)
--(axis cs:24,0.178911358118057)
--(axis cs:25,0.184134066104889)
--(axis cs:26,0.189927041530609)
--(axis cs:27,0.196218460798264)
--(axis cs:28,0.202718138694763)
--(axis cs:29,0.209500998258591)
--(axis cs:30,0.216589093208313)
--(axis cs:30,1)
--(axis cs:30,1)
--(axis cs:29,1)
--(axis cs:28,1)
--(axis cs:27,1)
--(axis cs:26,1)
--(axis cs:25,1)
--(axis cs:24,1)
--(axis cs:23,1)
--(axis cs:22,1)
--(axis cs:21,1)
--(axis cs:20,1)
--(axis cs:19,0.997411787509918)
--(axis cs:18,0.97332775592804)
--(axis cs:17,0.950165033340454)
--(axis cs:16,0.929225444793701)
--(axis cs:15,0.91197943687439)
--(axis cs:14,0.900238692760468)
--(axis cs:13,0.89563524723053)
--(axis cs:12,0.899397552013397)
--(axis cs:11,0.911787211894989)
--(axis cs:10,0.931554973125458)
--(axis cs:9,0.95664381980896)
--(axis cs:8,0.987413763999939)
--(axis cs:7,1)
--(axis cs:6,1)
--(axis cs:5,1)
--(axis cs:4,1)
--(axis cs:3,1)
--(axis cs:2,1)
--(axis cs:1,1)
--(axis cs:0,0.379381686449051)
--cycle;

\addplot [very thick, color0]
table {%
0 0.379381507635117
1 0.950173735618591
2 0.901468634605408
3 0.776543140411377
4 0.658077776432037
5 0.588475167751312
6 0.553180515766144
7 0.531935036182404
8 0.516607165336609
9 0.505551517009735
10 0.499361574649811
11 0.497894555330276
12 0.500560581684113
13 0.50677877664566
14 0.515597999095917
15 0.526313602924347
16 0.538361012935638
17 0.551322638988495
18 0.56482458114624
19 0.578639149665833
20 0.592466533184052
21 0.606159150600433
22 0.619667589664459
23 0.632871329784393
24 0.645673930644989
25 0.658064305782318
26 0.670016765594482
27 0.681511521339417
28 0.692506968975067
29 0.703022718429565
30 0.713084042072296
};
\addplot [very thick, black, dashed]
table {%
0 0.1
30 0.1
};
\end{axis}

\end{tikzpicture}}
  \qquad
  \subfloat{\input{figs/conf_temp_svhn}}
  \qquad
  \subfloat{
\begin{tikzpicture}

\tikzstyle{every node}=[font=\scriptsize]

\definecolor{color0}{rgb}{0.00392156862745098,0.450980392156863,0.698039215686274}

\begin{axis}[
width=0.28\textwidth,
height=0.2\textwidth,
tick align=outside,
tick pos=left,
x grid style={white!69.01960784313725!black},
xmin=0, xmax=30,
xtick style={color=black},
xtick={0,10,20,30},
xticklabels={\(\displaystyle 0\),\(\displaystyle 10\),\(\displaystyle 20\),\(\displaystyle 30\)},
y grid style={white!69.01960784313725!black},
ymin=0, ymax=1.05,
ytick style={color=black},
ytick={0,0.2,0.4,0.6,0.8,1,1.2},
xmajorticks=false, ymajorticks=false,
yticklabels={\(\displaystyle 0.0\),\(\displaystyle 0.2\),\(\displaystyle 0.4\),\(\displaystyle 0.6\),\(\displaystyle 0.8\),\(\displaystyle 1.0\),\(\displaystyle 1.2\)}
]
\path [fill=color0, fill opacity=0.15, semithick]
(axis cs:0,0.392882406711578)
--(axis cs:0,0.392882227897644)
--(axis cs:1,0.00999999977648258)
--(axis cs:2,0.00999999977648258)
--(axis cs:3,0.00999999977648258)
--(axis cs:4,0.00999999977648258)
--(axis cs:5,0.00999999977648258)
--(axis cs:6,0.00999999977648258)
--(axis cs:7,0.00999999977648258)
--(axis cs:8,0.00999999977648258)
--(axis cs:9,0.0401720404624939)
--(axis cs:10,0.0751219987869263)
--(axis cs:11,0.10281503200531)
--(axis cs:12,0.125267744064331)
--(axis cs:13,0.144092917442322)
--(axis cs:14,0.160313367843628)
--(axis cs:15,0.174761176109314)
--(axis cs:16,0.188373804092407)
--(axis cs:17,0.201281130313873)
--(axis cs:18,0.21353542804718)
--(axis cs:19,0.225998222827911)
--(axis cs:20,0.238543629646301)
--(axis cs:21,0.250872015953064)
--(axis cs:22,0.262802958488464)
--(axis cs:23,0.274552226066589)
--(axis cs:24,0.286324977874756)
--(axis cs:25,0.298019528388977)
--(axis cs:26,0.308884501457214)
--(axis cs:27,0.319042682647705)
--(axis cs:28,0.329203128814697)
--(axis cs:29,0.339036166667938)
--(axis cs:30,0.348297834396362)
--(axis cs:30,1)
--(axis cs:30,1)
--(axis cs:29,1)
--(axis cs:28,1)
--(axis cs:27,1)
--(axis cs:26,1)
--(axis cs:25,1)
--(axis cs:24,1)
--(axis cs:23,1)
--(axis cs:22,1)
--(axis cs:21,1)
--(axis cs:20,1)
--(axis cs:19,1)
--(axis cs:18,1)
--(axis cs:17,1)
--(axis cs:16,1)
--(axis cs:15,1)
--(axis cs:14,1)
--(axis cs:13,1)
--(axis cs:12,1)
--(axis cs:11,1)
--(axis cs:10,1)
--(axis cs:9,1)
--(axis cs:8,1)
--(axis cs:7,1)
--(axis cs:6,1)
--(axis cs:5,1)
--(axis cs:4,0.940497219562531)
--(axis cs:3,0.714090585708618)
--(axis cs:2,1)
--(axis cs:1,1)
--(axis cs:0,0.392882406711578)
--cycle;

\addplot [very thick, color0]
table {%
0 0.392882317304611
1 0.796695947647095
2 0.405396997928619
3 0.222774535417557
4 0.322944223880768
5 0.455004066228867
6 0.553008317947388
7 0.619614839553833
8 0.666190266609192
9 0.700534164905548
10 0.726815581321716
11 0.747609257698059
12 0.764467775821686
13 0.778444051742554
14 0.790216445922852
15 0.800288200378418
16 0.809070408344269
17 0.816808223724365
18 0.823685526847839
19 0.829952538013458
20 0.835702478885651
21 0.840986549854279
22 0.845848143100739
23 0.850366473197937
24 0.854613780975342
25 0.858618676662445
26 0.862314224243164
27 0.865743696689606
28 0.869024038314819
29 0.872131407260895
30 0.875053644180298
};
\addplot [very thick, black, dashed]
table {%
0 0.01
30 0.01
};
\end{axis}

\end{tikzpicture}}

  \vspace{-1em}
  \setcounter{subfigure}{0}

  \subfloat[MNIST]{
\begin{tikzpicture}[trim axis right]

\tikzstyle{every node}=[font=\scriptsize]

\definecolor{color0}{rgb}{0.00392156862745098,0.450980392156863,0.698039215686274}

\begin{axis}[
width=0.28\textwidth,
height=0.2\textwidth,
tick align=outside,
tick pos=left,
x grid style={white!69.01960784313725!black},
xlabel={\(\displaystyle \delta\)},
xmin=1, xmax=20,
xtick style={color=black},
xtick={0,5,10,15,20},
y grid style={white!69.01960784313725!black},
ylabel={Conf. (LLLA)},
ymin=0, ymax=1.05,
ytick style={color=black},
ytick={0,0.2,0.4,0.6,0.8,1,1.2},
]
\path [fill=color0, fill opacity=0.15, semithick]
(axis cs:0,0.124328963458538)
--(axis cs:0,0.113472066819668)
--(axis cs:1,0.807134032249451)
--(axis cs:2,0.827885627746582)
--(axis cs:3,0.819434463977814)
--(axis cs:4,0.812751531600952)
--(axis cs:5,0.810450077056885)
--(axis cs:6,0.80817186832428)
--(axis cs:7,0.804801821708679)
--(axis cs:8,0.802288830280304)
--(axis cs:9,0.800735712051392)
--(axis cs:10,0.800395965576172)
--(axis cs:11,0.801536500453949)
--(axis cs:12,0.799333333969116)
--(axis cs:13,0.798056185245514)
--(axis cs:14,0.798003792762756)
--(axis cs:15,0.796345412731171)
--(axis cs:16,0.795200347900391)
--(axis cs:17,0.794735848903656)
--(axis cs:18,0.798914313316345)
--(axis cs:19,0.793443918228149)
--(axis cs:20,0.792385876178741)
--(axis cs:20,1)
--(axis cs:20,1)
--(axis cs:19,1)
--(axis cs:18,1)
--(axis cs:17,1)
--(axis cs:16,1)
--(axis cs:15,1)
--(axis cs:14,1)
--(axis cs:13,1)
--(axis cs:12,1)
--(axis cs:11,1)
--(axis cs:10,1)
--(axis cs:9,1)
--(axis cs:8,1)
--(axis cs:7,1)
--(axis cs:6,1)
--(axis cs:5,1)
--(axis cs:4,1)
--(axis cs:3,1)
--(axis cs:2,1)
--(axis cs:1,1)
--(axis cs:0,0.124328963458538)
--cycle;

\addplot [very thick, color0]
table {%
0 0.118900515139103
1 0.985087156295776
2 0.989446640014648
3 0.988812267780304
4 0.988231122493744
5 0.987810373306274
6 0.987497091293335
7 0.987154304981232
8 0.986962795257568
9 0.986769258975983
10 0.986554384231567
11 0.986532092094421
12 0.98644232749939
13 0.986252665519714
14 0.986239612102509
15 0.986051738262177
16 0.985933601856232
17 0.985929191112518
18 0.986176490783691
19 0.985672116279602
20 0.985662341117859
};
\addplot [very thick, black, dashed]
table {%
1 0.1
20 0.1
};
\end{axis}

\end{tikzpicture}}
  \qquad
  \subfloat[CIFAR10]{
\begin{tikzpicture}[trim axis left, trim axis right]

\tikzstyle{every node}=[font=\scriptsize]

\definecolor{color0}{rgb}{0.00392156862745098,0.450980392156863,0.698039215686274}

\begin{axis}[
width=0.28\textwidth,
height=0.2\textwidth,
tick align=outside,
tick pos=left,
x grid style={white!69.01960784313725!black},
xlabel={\(\displaystyle \delta\)},
xmin=0, xmax=30,
xtick style={color=black},
xtick={0,10,20,30},
y grid style={white!69.01960784313725!black},
ymin=0, ymax=1.05,
ytick style={color=black},
ytick={0,0.2,0.4,0.6,0.8,1,1.2},
ymajorticks=false,
]
\path [fill=color0, fill opacity=0.15, semithick]
(axis cs:0,0.362391769886017)
--(axis cs:0,0.212717890739441)
--(axis cs:1,0.51719331741333)
--(axis cs:2,0.283435583114624)
--(axis cs:3,0.100000001490116)
--(axis cs:4,0.100000001490116)
--(axis cs:5,0.100000001490116)
--(axis cs:6,0.100000001490116)
--(axis cs:7,0.100000001490116)
--(axis cs:8,0.100000001490116)
--(axis cs:9,0.100000001490116)
--(axis cs:10,0.100000001490116)
--(axis cs:11,0.100000001490116)
--(axis cs:12,0.100000001490116)
--(axis cs:13,0.100000001490116)
--(axis cs:14,0.100000001490116)
--(axis cs:15,0.100000001490116)
--(axis cs:16,0.100000001490116)
--(axis cs:17,0.100000001490116)
--(axis cs:18,0.100000001490116)
--(axis cs:19,0.100000001490116)
--(axis cs:20,0.100000001490116)
--(axis cs:21,0.100000001490116)
--(axis cs:22,0.100000001490116)
--(axis cs:23,0.100000001490116)
--(axis cs:24,0.100000001490116)
--(axis cs:25,0.100000001490116)
--(axis cs:26,0.100000001490116)
--(axis cs:27,0.100000001490116)
--(axis cs:28,0.100000001490116)
--(axis cs:29,0.100000001490116)
--(axis cs:30,0.100000001490116)
--(axis cs:30,0.266953557729721)
--(axis cs:30,0.266953557729721)
--(axis cs:29,0.267490208148956)
--(axis cs:28,0.268476217985153)
--(axis cs:27,0.271142274141312)
--(axis cs:26,0.270573288202286)
--(axis cs:25,0.272203743457794)
--(axis cs:24,0.277360767126083)
--(axis cs:23,0.27629229426384)
--(axis cs:22,0.284352153539658)
--(axis cs:21,0.288962304592133)
--(axis cs:20,0.294613063335419)
--(axis cs:19,0.300481706857681)
--(axis cs:18,0.311272501945496)
--(axis cs:17,0.324787259101868)
--(axis cs:16,0.338744580745697)
--(axis cs:15,0.360235691070557)
--(axis cs:14,0.38505032658577)
--(axis cs:13,0.416545003652573)
--(axis cs:12,0.451476693153381)
--(axis cs:11,0.493596941232681)
--(axis cs:10,0.537880182266235)
--(axis cs:9,0.584479570388794)
--(axis cs:8,0.634307026863098)
--(axis cs:7,0.687408208847046)
--(axis cs:6,0.760208010673523)
--(axis cs:5,0.888026833534241)
--(axis cs:4,1)
--(axis cs:3,1)
--(axis cs:2,1)
--(axis cs:1,1)
--(axis cs:0,0.362391769886017)
--cycle;

\addplot [very thick, color0]
table {%
0 0.287554830312729
1 0.926282584667206
2 0.853442132472992
3 0.670593976974487
4 0.512842893600464
5 0.426471382379532
6 0.376485794782639
7 0.340043872594833
8 0.309781610965729
9 0.285342395305634
10 0.264942914247513
11 0.247890889644623
12 0.233609780669212
13 0.223108053207397
14 0.214786008000374
15 0.208004832267761
16 0.202808409929276
17 0.198632270097733
18 0.194958493113518
19 0.191749051213264
20 0.190316006541252
21 0.188553497195244
22 0.18675272166729
23 0.184402003884315
24 0.184596881270409
25 0.182942286133766
26 0.182296916842461
27 0.182136133313179
28 0.18097497522831
29 0.180965259671211
30 0.180471315979958
};
\addplot [very thick, black, dashed]
table {%
0 0.1
30 0.1
};
\end{axis}

\end{tikzpicture}}
  \qquad
  \subfloat[SVHN]{\input{figs/conf_laplace_svhn}}
  \qquad
  \subfloat[CIFAR100]{
\begin{tikzpicture}[trim axis left, trim axis right]

\tikzstyle{every node}=[font=\scriptsize]

\definecolor{color0}{rgb}{0.00392156862745098,0.450980392156863,0.698039215686274}

\begin{axis}[
width=0.28\textwidth,
height=0.2\textwidth,
tick align=outside,
tick pos=left,
x grid style={white!69.01960784313725!black},
xlabel={\(\displaystyle \delta\)},
xmin=0, xmax=30,
xtick style={color=black},
xtick={0,10,20,30},
y grid style={white!69.01960784313725!black},
ymin=0, ymax=1.05,
ytick style={color=black},
ytick={0,0.2,0.4,0.6,0.8,1,1.2},
ymajorticks=false,
]
\path [fill=color0, fill opacity=0.15, semithick]
(axis cs:0,0.383912980556488)
--(axis cs:0,0.234251782298088)
--(axis cs:1,0.00999999977648258)
--(axis cs:2,0.00999999977648258)
--(axis cs:3,0.00999999977648258)
--(axis cs:4,0.00999999977648258)
--(axis cs:5,0.00999999977648258)
--(axis cs:6,0.00999999977648258)
--(axis cs:7,0.00999999977648258)
--(axis cs:8,0.00999999977648258)
--(axis cs:9,0.00999999977648258)
--(axis cs:10,0.00999999977648258)
--(axis cs:11,0.00999999977648258)
--(axis cs:12,0.00999999977648258)
--(axis cs:13,0.00999999977648258)
--(axis cs:14,0.00999999977648258)
--(axis cs:15,0.00999999977648258)
--(axis cs:16,0.00999999977648258)
--(axis cs:17,0.00999999977648258)
--(axis cs:18,0.00999999977648258)
--(axis cs:19,0.00999999977648258)
--(axis cs:20,0.00999999977648258)
--(axis cs:21,0.00999999977648258)
--(axis cs:22,0.00999999977648258)
--(axis cs:23,0.00999999977648258)
--(axis cs:24,0.00999999977648258)
--(axis cs:25,0.00999999977648258)
--(axis cs:26,0.00999999977648258)
--(axis cs:27,0.00999999977648258)
--(axis cs:28,0.00999999977648258)
--(axis cs:29,0.00999999977648258)
--(axis cs:30,0.00999999977648258)
--(axis cs:30,0.157879620790482)
--(axis cs:30,0.157879620790482)
--(axis cs:29,0.161319553852081)
--(axis cs:28,0.164355844259262)
--(axis cs:27,0.163664847612381)
--(axis cs:26,0.167166888713837)
--(axis cs:25,0.171324953436852)
--(axis cs:24,0.174262404441833)
--(axis cs:23,0.176953792572021)
--(axis cs:22,0.181230619549751)
--(axis cs:21,0.187478572130203)
--(axis cs:20,0.192242562770844)
--(axis cs:19,0.19602707028389)
--(axis cs:18,0.20292866230011)
--(axis cs:17,0.210460379719734)
--(axis cs:16,0.218101993203163)
--(axis cs:15,0.225959628820419)
--(axis cs:14,0.237104162573814)
--(axis cs:13,0.253606557846069)
--(axis cs:12,0.265474081039429)
--(axis cs:11,0.283560276031494)
--(axis cs:10,0.30264487862587)
--(axis cs:9,0.327451884746552)
--(axis cs:8,0.353993654251099)
--(axis cs:7,0.382810771465302)
--(axis cs:6,0.410089761018753)
--(axis cs:5,0.430607736110687)
--(axis cs:4,0.428839206695557)
--(axis cs:3,0.469538867473602)
--(axis cs:2,1)
--(axis cs:1,1)
--(axis cs:0,0.383912980556488)
--cycle;

\addplot [very thick, color0]
table {%
0 0.309082388877869
1 0.748850226402283
2 0.342455863952637
3 0.16013790667057
4 0.178780972957611
5 0.197251290082932
6 0.19343401491642
7 0.179291158914566
8 0.163973465561867
9 0.150356739759445
10 0.13853657245636
11 0.128590568900108
12 0.120715647935867
13 0.114921279251575
14 0.108729287981987
15 0.10427812486887
16 0.100353673100471
17 0.0970326364040375
18 0.094859667122364
19 0.0921768769621849
20 0.0899581983685493
21 0.0886036902666092
22 0.0861795693635941
23 0.0847868174314499
24 0.0836823135614395
25 0.0823343843221664
26 0.0816804990172386
27 0.0803880468010902
28 0.0797883272171021
29 0.0792527794837952
30 0.0781140029430389
};
\addplot [very thick, black, dashed]
table {%
0 0.01
30 0.01
};
\end{axis}

\end{tikzpicture}}

  \caption{The multi-class confidence of MAP (top row), temperature scaling (middle row), and LLLA (bottom row) as functions of $\delta$ over the test sets of the multi-class datasets. Thick blue lines and shades correspond to means and $\pm 3$ standard deviations. Dotted lines signify the desirable confidence for $\delta$ sufficiently high.}
  \label{fig:exp_far_away_multiclass}
\end{figure*}

In \Cref{fig:exp_far_away_multiclass}, we present the multi-class counterpart of \Cref{fig:exp_far_away}. We found that, as in the binary case, the Bayesian method (LLLA) mitigates overconfidence in the asymptotic regime. We observed, however, that LLLA is less effective in MNIST, which might be due to the architecture choice and the training procedure used: The eigenvalues of the Gaussian posterior's covariance might be too small such that \eqref{eq:z} is still large.

\subsection{Rotated MNIST}

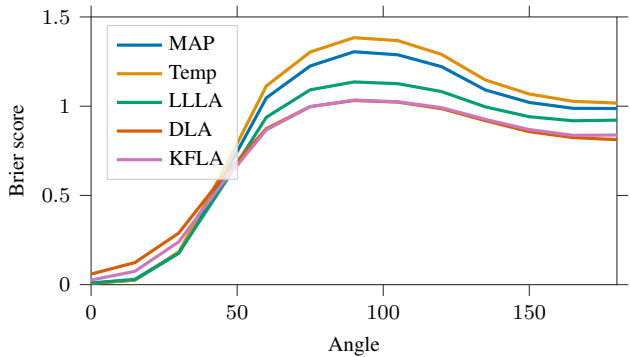
\begin{figure}[h]
  \centering

\begin{tikzpicture}

\definecolor{color0}{rgb}{0.00392156862745098,0.450980392156863,0.698039215686274}
\definecolor{color1}{rgb}{0.870588235294118,0.56078431372549,0.0196078431372549}
\definecolor{color2}{rgb}{0.00784313725490196,0.619607843137255,0.450980392156863}
\definecolor{color3}{rgb}{0.835294117647059,0.368627450980392,0}
\definecolor{color4}{rgb}{0.8,0.470588235294118,0.737254901960784}

\tikzstyle{every node}=[font=\fontsize{8}{9}\selectfont]

\begin{axis}[
width=0.5\textwidth,
height=0.3\textwidth,
axis line style={white!15.0!black},
legend cell align={left},
legend style={fill opacity=0.8, draw opacity=1, text opacity=1, at={(0.03,0.97)}, anchor=north west, draw=white!80.0!black},
tick align=outside,
x grid style={white!80.0!black},
xmin=0, xmax=180,
xtick style={color=white!15.0!black},
y grid style={white!80.0!black},
ymin=0, ymax=1.5,
ytick style={color=white!15.0!black},
xlabel={Angle},
ylabel={Brier score}
]
\addplot [very thick, color0]
table {%
0 0.00781412293650823
15 0.0259501588375447
30 0.176537412632689
45 0.594612529334294
60 1.04657915821736
75 1.22522100957458
90 1.30490721028484
105 1.287904170015
120 1.22116066005487
135 1.09074232083418
150 1.02092513877992
165 0.98789098647426
180 0.986961631915955
};
\addlegendentry{MAP}
\addplot [very thick, color1]
table {%
0 0.00757341162364879
15 0.0253799817553637
30 0.183623351712374
45 0.629574114566042
60 1.11315684172743
75 1.30347770997793
90 1.38379816954476
105 1.36735801544853
120 1.28990396725779
135 1.14646123668653
150 1.06815282941913
165 1.02726678090655
180 1.01738780868713
};
\addlegendentry{Temp}
\addplot [very thick, color2]
table {%
0 0.00923708850113416
15 0.0300196130524675
30 0.176866511134205
45 0.551196919971346
60 0.938088194720327
75 1.09160745387123
90 1.13573038927428
105 1.12611945145881
120 1.08166811174633
135 0.995944667295634
150 0.940955906957998
165 0.918333106599182
180 0.92176497973587
};
\addlegendentry{LLLA}
\addplot [very thick, color3]
table {%
0 0.0593668699665519
15 0.123624652220571
30 0.289016020475239
45 0.600480202610787
60 0.87411812929134
75 0.997567905269271
90 1.0318244716013
105 1.02277763687624
120 0.98614671122451
135 0.918896421210676
150 0.857149423781886
165 0.824006508575186
180 0.812544968176738
};
\addlegendentry{DLA}
\addplot [very thick, color4]
table {%
0 0.0257753595856141
15 0.0747279200848857
30 0.239708932315884
45 0.5690230380852
60 0.866850724531409
75 0.997352320032939
90 1.03448362811001
105 1.0247655589998
120 0.991887564234377
135 0.926128638092864
150 0.868974008531316
165 0.836844928886668
180 0.838264210815324
};
\addlegendentry{KFLA}
\end{axis}

\end{tikzpicture}

  \vspace{-1.5em}

  \caption{Brier scores (lower is better) over the rotated MNIST dataset. Values shown are means over ten trials. The standard deviations are very small and not visually observable.}
  \label{fig:rotated_mnist}
\end{figure}

Following \citet{ovadia2019can}, we further benchmark the methods in \Cref{subsec:ood_exp} on the rotated MNIST dataset. The goal is to see whether Bayesian methods could detect dataset shifts of increasing strength in term of Brier score \citep{brier1950verification}. Note that lower Brier score is favorable. We present the results in \Cref{fig:rotated_mnist}. We found that all Bayesian methods achieve lower Brier score compared to MAP and temperature scaling, signifying that Bayesian methods are better at detecting dataset shift.

\subsection{Adversarial Examples}

The adversarial datasets (``Adversarial'' and ``FarAwayAdv'', cf. \Cref{tab:ood_results}) are constructed as follows. For ``Adversarial'': We use the standard PGD attack \citep{madry2018towards} on a uniform noise dataset of size $2000$. The objective is to maximize the confidence of the MAP model (resp. ACET and OE below) inside of an $\ell^\infty$ ball with radius $\epsilon = 0.3$. The optimization is carried out for $40$ iterations with a step size of $0.1$. We ensure that the resulting adversarial examples are in the image space. For ``FarAwayAdv'': We use the same construction, but start from the ``far-away'' Noise datasets as used in \Cref{tab:ood_results} and we do not project the resulting adversarial examples onto the image space.

\subsection{Bayesian Methods on Top of State-of-the-art OOD Detectors}

We can also apply all methods we are considering here on top of the state-of-the-art models that are specifically trained to mitigate the overconfidence problem, namely ACET \citep{hein2019relu} and outlier exposure (OE) \citep{hendrycks2018deep}. The results are presented in \Cref{tab:ood_results_acet,tab:ood_results_oe}. In general, applying the Bayesian methods improves the models further, especially in the asymptotic regime.

\subsection{Frequentist Calibration}
\label{subsec:calibration_exp}

Although calibration is a frequentist approach for predictive uncertainty quantification, it is nevertheless interesting to get an insight on whether the properties of the Bayesian predictive distribution lead to a better calibration. To answer this, we use a standard metric \citep{naeini2015obtaining,guo17calibration}: the expected calibration error (ECE). We use the same models along with the same hyperparameters as we have used in the previous OOD experiments. We present the results in \Cref{tab:calibration}. We found that all the Bayesian methods are competitive to the temperature scaling method, which is specifically constructed for improving the frequentist calibration.

\begin{table}[t]
  \caption{Expected calibration errors (ECE).}
  \label{tab:calibration}

  \vspace{1em}

  \centering
  \scriptsize

    \begin{tabular}{lrrrr}
      \toprule

       & {\bf MNIST} & {\bf CIFAR10} & {\bf SVHN} & {\bf CIFAR100} \\

      \midrule

      MAP        & \textbf{6.7$\pm$0.3}  & 13.1$\pm$0.2 & 10.1$\pm$0.2  & 8.1$\pm$0.3  \\
      +Temp.     & 11.4$\pm$2.2 & \textbf{3.6$\pm$0.6}  & \textbf{2.1$\pm$0.5}  & 6.4$\pm$0.5  \\
      +LLLA      & 6.9$\pm$0.3  & \textbf{3.6$\pm$0.6}  & 5.2$\pm$0.8  & 4.8$\pm$0.3  \\
      +DLA       & 15.5$\pm$0.2  & 6.9$\pm$0.1  & 8.3$\pm$0.0  & \textbf{4.7$\pm$0.3}  \\
      +KFLA      & 9.7$\pm$0.3  & 7.9$\pm$0.1  & 6.5$\pm$0.1  & 5.6$\pm$0.4  \\

      \midrule

      ACET       & \textbf{5.9$\pm$0.2}  & 15.8$\pm$0.4 & 11.9$\pm$0.2 & 10.1$\pm$0.4 \\
      +Temp.     & 11.0$\pm$1.5 & \textbf{3.7$\pm$0.8}  & 2.3$\pm$0.4  & 6.4$\pm$0.4  \\
      +LLLA      & 6.1$\pm$0.2  & 12.3$\pm$0.7  & 9.3$\pm$0.5  & 6.9$\pm$0.3  \\
      +DLA       & 6.2$\pm$0.3  & 4.3$\pm$0.3  & \textbf{2.0$\pm$0.1}  & 6.0$\pm$0.3 \\
      +KFLA      & 6.1$\pm$0.3  & 4.3$\pm$0.2  & 2.1$\pm$0.1  & \textbf{4.6$\pm$0.2} \\

      \midrule

      OE         & 14.7$\pm$1.2 & 15.8$\pm$0.3 & 11.0$\pm$0.1 & 25.0$\pm$0.2 \\
      +Temp.     & 9.0$\pm$2.3  & 23.3$\pm$0.7 & \textbf{3.7$\pm$0.7}  & \textbf{19.4$\pm$0.2} \\
      +LLLA      & \textbf{6.5$\pm$0.6}  & \textbf{14.6$\pm$0.2} & 4.1$\pm$0.3  & 24.9$\pm$0.4 \\
      +DLA       & 9.1$\pm$0.6  & 15.8$\pm$0.3 & 7.2$\pm$0.1  & 29.0$\pm$0.2 \\
      +KFLA      & 10.1$\pm$0.9 & 15.9$\pm$0.3 & 6.4$\pm$0.1  & 29.0$\pm$0.2 \\

    \bottomrule
  \end{tabular}
\end{table}

\begin{figure*}[t!]
  \centering

  \subfloat{\input{figs/hist_confs_mnist_binary_map}}
  \quad 
  \subfloat{\input{figs/hist_confs_cifar10_binary_map}}
  \quad 
  \subfloat{\input{figs/hist_confs_svhn_binary_map}}
  \quad 
  \subfloat{\input{figs/hist_confs_cifar100_binary_map}}

  \vspace{-1em}
  \setcounter{subfigure}{0}

  \subfloat{\input{figs/hist_confs_mnist_binary_temp}}
  \quad 
  \subfloat{\input{figs/hist_confs_cifar10_binary_temp}}
  \quad 
  \subfloat{\input{figs/hist_confs_svhn_binary_temp}}
  \quad 
  \subfloat{\input{figs/hist_confs_cifar100_binary_temp}}

  \vspace{-1em}
  \setcounter{subfigure}{0}

  \subfloat[Bin.-MNIST]{\input{figs/hist_confs_mnist_binary_llla}}
  \quad 
  \subfloat[Bin.-CIFAR10]{\input{figs/hist_confs_cifar10_binary_llla}}
  \quad 
  \subfloat[Bin.-SVHN]{\input{figs/hist_confs_svhn_binary_llla}}
  \quad 
  \subfloat[Bin.-CIFAR100]{\input{figs/hist_confs_cifar100_binary_llla}}

  \caption{The histograms of MAP (top row), temperature scaling (middle row), and LLLA (bottom row) over the binary datasets. Each entry ``Out - FarAway'' refers to the OOD dataset obtained by scaling the corresponding in-distribution dataset with some $\delta > 0$.}
  \label{fig:histograms_binary}
\end{figure*}

\begin{figure*}[t!]
  \centering

  \subfloat{\input{figs/hist_confs_mnist_map}}
  \quad 
  \subfloat{\input{figs/hist_confs_cifar10_map}}
  \quad 
  \subfloat{\input{figs/hist_confs_svhn_map}}
  \quad 
  \subfloat{\input{figs/hist_confs_cifar100_binary_map}}

  \vspace{-1em}
  \setcounter{subfigure}{0}

  \subfloat{\input{figs/hist_confs_mnist_temp}}
  \quad 
  \subfloat{\input{figs/hist_confs_cifar10_temp}}
  \quad 
  \subfloat{\input{figs/hist_confs_svhn_temp}}
  \quad 
  \subfloat{\input{figs/hist_confs_cifar100_temp}}

  \vspace{-1em}
  \setcounter{subfigure}{0}

  \subfloat[MNIST]{\input{figs/hist_confs_mnist_llla}}
  \quad 
  \subfloat[CIFAR10]{\input{figs/hist_confs_cifar10_llla}}
  \quad 
  \subfloat[SVHN]{\input{figs/hist_confs_svhn_llla}}
  \quad 
  \subfloat[CIFAR100]{\input{figs/hist_confs_cifar100_llla}}

  \caption{The histograms of MAP (top row), temperature scaling (middle row), and LLLA (bottom row) over the multi-class datasets. Each entry ``Out - FarAway'' refers to the OOD dataset obtained by scaling the corresponding in-distribution dataset with some $\delta > 0$.}
  \label{fig:histograms_multiclass}
\end{figure*}

\begin{table*}[t]
  \caption{Adversarial OOD detection results.}
  \label{tab:adversarial}

  \vspace{1em}

  \centering
  \scriptsize
  \renewcommand{\tabcolsep}{6pt}

  \begin{tabular}{lrrrr|rrrrrr}
    \toprule

      & \multicolumn{2}{c}{\bf MAP} & \multicolumn{2}{c|}{\bf +Temp.} & \multicolumn{2}{c}{\bf +LLLA} & \multicolumn{2}{c}{\bf +DLA} & \multicolumn{2}{c}{\bf +KFLA} \\
    \cmidrule(r){2-3} \cmidrule(r){4-5} \cmidrule(l){6-7} \cmidrule(l){8-9} \cmidrule(l){10-11}
      & MMC & AUR & MMC & AUR & MMC & AUR & MMC & AUR & MMC & AUR \\

    \midrule

    MNIST - Adversarial & 100.0$\pm$0.0 & 0.3$\pm$0.0 & 100.0$\pm$0.0 & 6.8$\pm$4.1 & 100.0$\pm$0.0 & 5.3$\pm$0.1 & 99.6$\pm$0.2 & 2.0$\pm$0.9 & \textbf{91.3$\pm$1.2} & \textbf{69.2$\pm$3.5} \\
    MNIST - FarAwayAdv & 100.0$\pm$0.0 & 0.1$\pm$0.0 & 100.0$\pm$0.0 & 6.8$\pm$4.1 & 99.9$\pm$0.0 & 9.3$\pm$0.6 & 85.3$\pm$1.4 & 53.0$\pm$3.8 & \textbf{55.6$\pm$2.0} & \textbf{97.4$\pm$0.3} \\

    \midrule

    CIFAR10 - Adversarial & 100.0$\pm$0.0 & 0.0$\pm$0.0 & 100.0$\pm$0.0 & 0.0$\pm$0.0 & 99.7$\pm$0.0 & \textbf{9.1$\pm$0.1} & 99.3$\pm$0.1 & 9.0$\pm$1.0 & \textbf{99.2$\pm$0.0} & 5.8$\pm$0.4 \\
    CIFAR10 - FarAwayAdv & 99.5$\pm$0.0 & 8.8$\pm$0.0 & 99.2$\pm$0.0 & 7.9$\pm$0.1 & \textbf{17.4$\pm$0.1} & \textbf{100.0$\pm$0.0} & 61.3$\pm$2.4 & 89.4$\pm$1.0 & 61.2$\pm$1.3 & 87.8$\pm$0.8 \\

    \midrule

    SVHN - Adversarial & 100.0$\pm$0.0 & 0.0$\pm$0.0 & 100.0$\pm$0.0 & 0.0$\pm$0.0 & \textbf{97.6$\pm$0.0} & \textbf{32.5$\pm$0.3} & 98.6$\pm$0.0 & 6.8$\pm$0.3 & 98.6$\pm$0.1 & 9.6$\pm$0.4 \\
    SVHN - FarAwayAdv & 99.7$\pm$0.0 & 7.7$\pm$0.0 & 99.5$\pm$0.0 & 6.9$\pm$0.1 & \textbf{27.5$\pm$0.1} & \textbf{99.6$\pm$0.0} & 61.7$\pm$1.4 & 92.4$\pm$0.9 & 61.0$\pm$1.2 & 94.4$\pm$0.3 \\

    \midrule

    CIFAR100 - Adversarial & \textbf{100.0$\pm$0.0} & 0.0$\pm$0.0 & \textbf{100.0$\pm$0.0} & 0.0$\pm$0.0 & \textbf{100.0$\pm$0.0} & \textbf{0.2$\pm$0.0} & \textbf{100.0$\pm$0.0} & 0.1$\pm$0.0 & \textbf{100.0$\pm$0.0} & 0.0$\pm$0.0 \\
    CIFAR100 - FarAwayAdv & 100.0$\pm$0.0 & 1.3$\pm$0.0 & 99.9$\pm$0.0 & 1.2$\pm$0.0 & \textbf{5.9$\pm$0.0} & \textbf{99.9$\pm$0.0} & 42.0$\pm$1.5 & 83.9$\pm$0.9 & 42.3$\pm$1.8 & 80.8$\pm$1.2 \\

    \bottomrule
  \end{tabular}
\end{table*}

\begin{table*}[t]
  \caption{OOD detection results when applying post-hoc Bayesian methods on top of models trained with ACET \citep{hein2019relu}.}
  \label{tab:ood_results_acet}

  \vspace{1em}

  \centering
  \scriptsize
  \renewcommand{\tabcolsep}{4pt}

  \begin{tabular}{lrrrr|rrrrrr}
    \toprule

      & \multicolumn{2}{c}{\bf MAP} & \multicolumn{2}{c|}{\bf +Temp.} & \multicolumn{2}{c}{\bf +LLLA} & \multicolumn{2}{c}{\bf +DLA} & \multicolumn{2}{c}{\bf +KFLA} \\
    \cmidrule(r){2-3} \cmidrule(r){4-5} \cmidrule(l){6-7} \cmidrule(l){8-9} \cmidrule(l){10-11}
      & MMC & AUR & MMC & AUR & MMC & AUR & MMC & AUR & MMC & AUR \\

    \midrule

    MNIST - MNIST & 98.9$\pm$0.0 & - & 99.5$\pm$0.0 & - & 98.9$\pm$0.0 & - & 98.9$\pm$0.0 & - & 98.9$\pm$0.0 & - \\
    MNIST - EMNIST & 59.1$\pm$0.0 & \textbf{96.9$\pm$0.0} & 70.9$\pm$1.8 & 96.5$\pm$0.1 & \textbf{59.0$\pm$0.0} & \textbf{96.9$\pm$0.0} & \textbf{59.0$\pm$0.0} & \textbf{96.9$\pm$0.0} & 59.1$\pm$0.0 & \textbf{96.9$\pm$0.0} \\
    MNIST - FMNIST & \textbf{10.2$\pm$0.0} & \textbf{100.0$\pm$0.0} & 10.3$\pm$0.0 & \textbf{100.0$\pm$0.0} & \textbf{10.2$\pm$0.0} & \textbf{100.0$\pm$0.0} & \textbf{10.2$\pm$0.0} & \textbf{100.0$\pm$0.0} & \textbf{10.2$\pm$0.0} & \textbf{100.0$\pm$0.0} \\
    MNIST - Noise ($\delta = 2000$) & 100.0$\pm$0.0 & 0.0$\pm$0.0 & 100.0$\pm$0.0 & \textbf{21.9$\pm$9.2} & 100.0$\pm$0.0 & 0.3$\pm$0.2 & \textbf{99.9$\pm$0.0} & 0.3$\pm$0.2 & 100.0$\pm$0.0 & 0.2$\pm$0.1 \\
    MNIST - Adversarial & \textbf{10.0$\pm$0.0} & \textbf{100.0$\pm$0.0} & \textbf{10.0$\pm$0.0} & \textbf{100.0$\pm$0.0} & 10.1$\pm$0.0 & \textbf{100.0$\pm$0.0} & \textbf{10.0$\pm$0.0} & \textbf{100.0$\pm$0.0} & \textbf{10.0$\pm$0.0} & \textbf{100.0$\pm$0.0} \\
    MNIST - FarAwayAdv & \textbf{100.0$\pm$0.0} & 0.0$\pm$0.0 & \textbf{100.0$\pm$0.0} & \textbf{21.9$\pm$9.2} & \textbf{100.0$\pm$0.0} & 0.1$\pm$0.0 & \textbf{100.0$\pm$0.0} & 0.2$\pm$0.0 & \textbf{100.0$\pm$0.0} & 0.1$\pm$0.0 \\

    \midrule

    CIFAR10 - CIFAR10 & 97.3$\pm$0.0 & - & 95.2$\pm$0.2 & - & 96.7$\pm$0.1 & - & 94.7$\pm$0.0 & - & 94.9$\pm$0.0 & - \\
    CIFAR10 - SVHN & 62.8$\pm$0.0 & 96.1$\pm$0.0 & \textbf{52.9$\pm$0.7} & \textbf{96.5$\pm$0.0} & 59.5$\pm$0.5 & 96.1$\pm$0.1 & 53.1$\pm$0.1 & 96.2$\pm$0.1 & 53.7$\pm$0.1 & 96.2$\pm$0.0 \\
    CIFAR10 - LSUN & 72.1$\pm$0.0 & 92.8$\pm$0.0 & 62.6$\pm$0.7 & 93.2$\pm$0.1 & 68.9$\pm$0.6 & 92.8$\pm$0.1 & \textbf{59.9$\pm$0.6} & \textbf{93.8$\pm$0.2} & 60.4$\pm$0.2 & 93.7$\pm$0.1 \\
    CIFAR10 - Noise ($\delta = 2000$) & 100.0$\pm$0.0 & 0.0$\pm$0.0 & 100.0$\pm$0.0 & 0.0$\pm$0.0 & \textbf{16.0$\pm$0.0} & \textbf{100.0$\pm$0.0} & 71.7$\pm$1.7 & 92.3$\pm$0.7 & 65.8$\pm$1.8 & 94.4$\pm$0.5 \\
    CIFAR10 - Adversarial & 78.1$\pm$0.0 & 83.1$\pm$0.1 & 71.1$\pm$0.5 & 84.1$\pm$0.1 & 76.8$\pm$0.0 & 83.2$\pm$0.1 & 67.9$\pm$0.4 & 88.5$\pm$0.2 & \textbf{67.7$\pm$0.3} & \textbf{88.8$\pm$0.2} \\
    CIFAR10 - FarAwayAdv & 100.0$\pm$0.0 & 0.0$\pm$0.0 & 100.0$\pm$0.0 & 0.0$\pm$0.0 & \textbf{16.0$\pm$0.0} & \textbf{100.0$\pm$0.0} & 72.5$\pm$2.4 & 92.1$\pm$0.7 & 70.7$\pm$1.9 & 93.1$\pm$0.5 \\

    \midrule

    SVHN - SVHN & 98.5$\pm$0.0 & - & 97.3$\pm$0.2 & - & 98.3$\pm$0.0 & - & 96.7$\pm$0.0 & - & 96.2$\pm$0.0 & - \\
    SVHN - CIFAR10 & 65.9$\pm$0.0 & 95.6$\pm$0.0 & 58.5$\pm$0.8 & 95.7$\pm$0.0 & 64.0$\pm$0.3 & 95.7$\pm$0.0 & 49.8$\pm$0.1 & \textbf{97.5$\pm$0.0} & \textbf{48.3$\pm$0.1} & 97.4$\pm$0.0 \\
    SVHN - LSUN & 28.0$\pm$0.0 & 99.3$\pm$0.0 & 24.6$\pm$0.3 & 99.4$\pm$0.0 & 27.8$\pm$0.1 & 99.3$\pm$0.0 & 22.8$\pm$0.6 & \textbf{99.6$\pm$0.0} & \textbf{21.7$\pm$0.6} & \textbf{99.6$\pm$0.0} \\
    SVHN - Noise ($\delta = 2000$) & 17.9$\pm$0.2 & \textbf{100.0$\pm$0.0} & 16.0$\pm$0.3 & \textbf{100.0$\pm$0.0} & \textbf{15.0$\pm$0.0} & \textbf{100.0$\pm$0.0} & 45.1$\pm$2.1 & 99.0$\pm$0.2 & 41.6$\pm$1.3 & 99.1$\pm$0.1 \\
    SVHN - Adversarial & 10.4$\pm$0.0 & \textbf{100.0$\pm$0.0} & \textbf{10.3$\pm$0.0} & \textbf{100.0$\pm$0.0} & 10.8$\pm$0.0 & \textbf{100.0$\pm$0.0} & 10.4$\pm$0.0 & \textbf{100.0$\pm$0.0} & 10.4$\pm$0.0 & \textbf{100.0$\pm$0.0} \\
    SVHN - FarAwayAdv & 17.6$\pm$0.0 & \textbf{100.0$\pm$0.0} & 15.7$\pm$0.2 & \textbf{100.0$\pm$0.0} & \textbf{15.0$\pm$0.0} & \textbf{100.0$\pm$0.0} & 44.9$\pm$2.6 & 99.0$\pm$0.2 & 44.8$\pm$1.5 & 98.8$\pm$0.1 \\

    \midrule

    CIFAR100 - CIFAR100 & 82.0$\pm$0.1 & - & 78.1$\pm$0.5 & - & 79.6$\pm$0.1 & - & 78.7$\pm$0.1 & - & 76.3$\pm$0.1 & - \\
    CIFAR100 - SVHN & 57.1$\pm$0.0 & 77.8$\pm$0.1 & \textbf{49.5$\pm$0.8} & \textbf{78.7$\pm$0.1} & 52.7$\pm$0.1 & 78.4$\pm$0.1 & 52.4$\pm$0.0 & 77.7$\pm$0.1 & \textbf{49.5$\pm$0.0} & 77.4$\pm$0.2 \\
    CIFAR100 - LSUN & 55.1$\pm$0.0 & 78.8$\pm$0.1 & 48.3$\pm$0.7 & 79.0$\pm$0.1 & 50.6$\pm$0.1 & \textbf{79.5$\pm$0.1} & 49.8$\pm$0.1 & 79.3$\pm$0.1 & \textbf{46.8$\pm$0.2} & 79.2$\pm$0.2 \\
    CIFAR100 - Noise ($\delta = 2000$) & 99.3$\pm$0.1 & 4.2$\pm$0.2 & 99.2$\pm$0.1 & 3.8$\pm$0.2 & \textbf{5.4$\pm$0.0} & \textbf{100.0$\pm$0.0} & 58.5$\pm$1.3 & 76.1$\pm$0.9 & 51.8$\pm$1.1 & 77.6$\pm$0.9 \\
    CIFAR100 - Adversarial & 1.5$\pm$0.0 & \textbf{100.0$\pm$0.0} & \textbf{1.4$\pm$0.0} & \textbf{100.0$\pm$0.0} & 1.5$\pm$0.0 & \textbf{100.0$\pm$0.0} & \textbf{1.4$\pm$0.0} & \textbf{100.0$\pm$0.0} & \textbf{1.4$\pm$0.0} & \textbf{100.0$\pm$0.0} \\
    CIFAR100 - FarAwayAdv & 99.7$\pm$0.0 & 3.4$\pm$0.0 & 99.6$\pm$0.0 & 3.1$\pm$0.0 & \textbf{5.4$\pm$0.0} & \textbf{100.0$\pm$0.0} & 57.7$\pm$1.5 & 76.6$\pm$1.0 & 57.9$\pm$1.4 & 73.4$\pm$1.0 \\

    \bottomrule
  \end{tabular}
\end{table*}

\begin{table*}[t]
  \caption{OOD detection results when applying post-hoc Bayesian methods on top of models trained with outlier exposure (OE) \citep{hendrycks2018deep}.}
  \label{tab:ood_results_oe}

  \vspace{1em}

  \centering
  \scriptsize
  \renewcommand{\tabcolsep}{4pt}

  \begin{tabular}{lrrrr|rrrrrr}
    \toprule

      & \multicolumn{2}{c}{\bf MAP} & \multicolumn{2}{c|}{\bf +Temp.} & \multicolumn{2}{c}{\bf +LLLA} & \multicolumn{2}{c}{\bf +DLA} & \multicolumn{2}{c}{\bf +KFLA} \\
    \cmidrule(r){2-3} \cmidrule(r){4-5} \cmidrule(l){6-7} \cmidrule(l){8-9} \cmidrule(l){10-11}
      & MMC & AUR & MMC & AUR & MMC & AUR & MMC & AUR & MMC & AUR \\

    \midrule

    MNIST - MNIST & 99.6$\pm$0.0 & - & 99.4$\pm$0.1 & - & 97.8$\pm$0.8 & - & 99.4$\pm$0.0 & - & 99.4$\pm$0.0 & - \\
    MNIST - EMNIST & 84.2$\pm$0.0 & 96.0$\pm$0.1 & 77.1$\pm$2.6 & \textbf{96.3$\pm$0.1} & \textbf{67.3$\pm$3.1} & 94.3$\pm$0.7 & 79.6$\pm$0.0 & 95.6$\pm$0.1 & 79.1$\pm$0.0 & 95.9$\pm$0.1 \\
    MNIST - FMNIST & 27.9$\pm$0.0 & \textbf{99.9$\pm$0.0} & \textbf{22.8$\pm$1.5} & \textbf{99.9$\pm$0.0} & 25.6$\pm$1.6 & \textbf{99.9$\pm$0.0} & 27.5$\pm$0.1 & \textbf{99.9$\pm$0.0} & 27.3$\pm$0.0 & \textbf{99.9$\pm$0.0} \\
    MNIST - Noise ($\delta = 2000$) & 99.9$\pm$0.0 & 26.4$\pm$0.2 & 99.9$\pm$0.0 & 5.0$\pm$2.4 & 66.0$\pm$0.6 & 95.9$\pm$0.4 & 58.4$\pm$0.3 & 97.6$\pm$0.3 & \textbf{49.8$\pm$0.3} & \textbf{99.3$\pm$0.1} \\
    MNIST - Adversarial & 40.5$\pm$0.0 & 98.8$\pm$0.0 & \textbf{35.2$\pm$1.1} & 99.1$\pm$0.0 & 38.7$\pm$0.0 & 98.1$\pm$0.0 & 38.1$\pm$0.0 & 98.7$\pm$0.0 & 35.8$\pm$0.1 & \textbf{99.2$\pm$0.0} \\
    MNIST - FarAwayAdv & 100.0$\pm$0.0 & 25.5$\pm$0.2 & 100.0$\pm$0.0 & 3.6$\pm$2.4 & 66.6$\pm$0.1 & 95.7$\pm$0.1 & 59.2$\pm$0.3 & 97.2$\pm$0.2 & \textbf{50.5$\pm$0.3} & \textbf{99.3$\pm$0.1} \\

    \midrule

    CIFAR10 - CIFAR10 & 89.4$\pm$0.1 & - & 92.5$\pm$0.4 & - & 89.2$\pm$0.1 & - & 89.3$\pm$0.1 & - & 89.3$\pm$0.1 & - \\
    CIFAR10 - SVHN & \textbf{10.8$\pm$0.0} & \textbf{98.8$\pm$0.0} & 11.2$\pm$0.1 & \textbf{98.8$\pm$0.0} & 10.9$\pm$0.0 & 98.7$\pm$0.0 & \textbf{10.8$\pm$0.0} & \textbf{98.8$\pm$0.0} & \textbf{10.8$\pm$0.0} & \textbf{98.8$\pm$0.0} \\
    CIFAR10 - LSUN & \textbf{10.4$\pm$0.0} & \textbf{98.6$\pm$0.0} & 10.7$\pm$0.1 & \textbf{98.6$\pm$0.0} & 10.6$\pm$0.0 & 98.5$\pm$0.1 & \textbf{10.4$\pm$0.0} & \textbf{98.6$\pm$0.0} & \textbf{10.4$\pm$0.0} & \textbf{98.6$\pm$0.0} \\
    CIFAR10 - Noise ($\delta = 2000$) & 99.1$\pm$0.1 & 6.5$\pm$0.6 & 99.4$\pm$0.1 & 7.6$\pm$0.7 & \textbf{25.0$\pm$0.1} & \textbf{93.6$\pm$0.1} & 77.9$\pm$1.0 & 79.5$\pm$2.0 & 72.7$\pm$1.5 & 84.6$\pm$1.2 \\
    CIFAR10 - Adversarial & \textbf{98.5$\pm$0.0} & 2.4$\pm$0.0 & 98.8$\pm$0.0 & \textbf{2.6$\pm$0.2} & \textbf{98.5$\pm$0.0} & 2.4$\pm$0.0 & \textbf{98.5$\pm$0.0} & 2.4$\pm$0.0 & \textbf{98.5$\pm$0.0} & 2.4$\pm$0.0 \\
    CIFAR10 - FarAwayAdv & 99.5$\pm$0.0 & 5.2$\pm$0.0 & 99.8$\pm$0.0 & 6.2$\pm$0.3 & \textbf{25.1$\pm$0.1} & \textbf{93.6$\pm$0.1} & 79.4$\pm$1.1 & 78.4$\pm$1.9 & 78.1$\pm$1.4 & 79.6$\pm$1.7 \\

    \midrule

    SVHN - SVHN & 97.4$\pm$0.0 & - & 95.8$\pm$0.3 & - & 95.7$\pm$0.2 & - & 92.5$\pm$0.0 & - & 93.5$\pm$0.0 & - \\
    SVHN - CIFAR10 & 10.2$\pm$0.0 & \textbf{100.0$\pm$0.0} & \textbf{10.1$\pm$0.0} & \textbf{100.0$\pm$0.0} & 14.3$\pm$0.6 & 99.9$\pm$0.0 & 10.8$\pm$0.0 & \textbf{100.0$\pm$0.0} & 10.8$\pm$0.0 & \textbf{100.0$\pm$0.0} \\
    SVHN - LSUN & \textbf{10.1$\pm$0.0} & \textbf{100.0$\pm$0.0} & \textbf{10.1$\pm$0.0} & \textbf{100.0$\pm$0.0} & 14.2$\pm$0.6 & 99.9$\pm$0.0 & 10.8$\pm$0.0 & \textbf{100.0$\pm$0.0} & 10.9$\pm$0.1 & \textbf{100.0$\pm$0.0} \\
    SVHN - Noise ($\delta = 2000$) & 99.7$\pm$0.0 & 3.0$\pm$0.2 & 99.6$\pm$0.1 & 2.7$\pm$0.2 & \textbf{16.2$\pm$0.0} & \textbf{99.7$\pm$0.0} & 31.5$\pm$1.4 & 98.4$\pm$0.2 & 33.0$\pm$1.3 & 98.4$\pm$0.2 \\
    SVHN - Adversarial & 44.9$\pm$0.0 & 98.2$\pm$0.0 & 34.4$\pm$0.7 & 98.5$\pm$0.0 & 34.2$\pm$0.0 & 98.5$\pm$0.0 & \textbf{17.9$\pm$0.2} & 99.5$\pm$0.0 & 18.2$\pm$0.2 & \textbf{99.6$\pm$0.0} \\
    SVHN - FarAwayAdv & 99.9$\pm$0.0 & 2.4$\pm$0.0 & 99.8$\pm$0.0 & 2.2$\pm$0.0 & \textbf{16.3$\pm$0.0} & \textbf{99.7$\pm$0.0} & 32.1$\pm$1.2 & 98.3$\pm$0.2 & 31.7$\pm$1.6 & 98.6$\pm$0.2 \\

    \midrule

    CIFAR100 - CIFAR100 & 59.6$\pm$0.2 & - & 71.8$\pm$0.5 & - & 54.9$\pm$0.2 & - & 51.5$\pm$0.2 & - & 52.0$\pm$0.2 & - \\
    CIFAR100 - SVHN & \textbf{3.6$\pm$0.0} & \textbf{93.5$\pm$0.1} & 7.2$\pm$0.2 & 93.4$\pm$0.1 & \textbf{3.6$\pm$0.2} & 92.9$\pm$0.5 & \textbf{3.6$\pm$0.0} & 93.2$\pm$0.1 & \textbf{3.6$\pm$0.0} & 93.4$\pm$0.1 \\
    CIFAR100 - LSUN & 2.6$\pm$0.0 & 95.4$\pm$0.1 & 5.0$\pm$0.1 & 95.3$\pm$0.1 & 2.9$\pm$0.1 & 94.6$\pm$0.2 & \textbf{2.5$\pm$0.1} & \textbf{95.9$\pm$0.2} & \textbf{2.5$\pm$0.1} & \textbf{95.9$\pm$0.2} \\
    CIFAR100 - Noise ($\delta = 2000$) & 100.0$\pm$0.0 & 1.3$\pm$0.0 & 100.0$\pm$0.0 & 7.3$\pm$0.7 & \textbf{25.3$\pm$0.1} & \textbf{64.5$\pm$0.2} & 89.7$\pm$1.9 & 25.0$\pm$2.7 & 82.4$\pm$1.4 & 33.5$\pm$1.3 \\
    CIFAR100 - Adversarial & 95.6$\pm$0.0 & 21.7$\pm$0.1 & 96.7$\pm$0.0 & 24.6$\pm$0.4 & \textbf{67.3$\pm$0.1} & \textbf{40.2$\pm$0.2} & 89.8$\pm$0.3 & 23.9$\pm$0.3 & 89.8$\pm$0.2 & 24.9$\pm$0.2 \\
    CIFAR100 - FarAwayAdv & 100.0$\pm$0.0 & 1.3$\pm$0.0 & 100.0$\pm$0.0 & 7.3$\pm$0.7 & \textbf{25.3$\pm$0.1} & \textbf{64.5$\pm$0.2} & 89.4$\pm$1.6 & 25.6$\pm$2.2 & 89.1$\pm$1.9 & 27.2$\pm$2.2 \\

    \bottomrule
  \end{tabular}
\end{table*}

\end{appendices}

\end{document}